\documentclass[runningheads]{llncs}

\usepackage{eccv}

\usepackage{eccvabbrv}

\usepackage{graphicx}
\usepackage{booktabs}
\usepackage{amssymb}
\usepackage{mdframed}
\usepackage{algorithm}
\usepackage{algorithmic}
\usepackage{amsmath}
\usepackage{tabularx}
\usepackage{multirow}
\usepackage{array}
\usepackage[table]{xcolor}
\definecolor{groupbg}{HTML}{E0E0E0}
\definecolor{hilite}{HTML}{D6F1FF}
\definecolor{stronghl}{RGB}{255, 245, 200}
\usepackage{wrapfig}

\usepackage{pifont}
\newcommand{\cmark}{\ding{51}}
\newcommand{\xmark}{\ding{55}}
\usepackage[accsupp]{axessibility}  

\usepackage{hyperref}

\usepackage{orcidlink}

\begin{document}

\title{Mixed-Modality Dual Face–Hair Retrieval} 

\titlerunning{Mixed-Modality Dual Face--Hair Retrieval}
\author{Quoc-Anh Bui-Huynh\inst{1,2} \and
Mai-Tuyen Lam\inst{1,2} \and
Dai-Anh-Tuan Nguyen\inst{1,2} \and
Thanh Duc Ngo\inst{1,2}\thanks{Corresponding author.}}
\authorrunning{A.~Bui et al.}
\institute{Vietnam National University, Ho Chi Minh City, Vietnam \and
University of Information Technology, VNU-HCM, Ho Chi Minh City, Vietnam\\
\email{\{22520038, 22521629, 21522753\}@gm.uit.edu.vn, thanhnd@uit.edu.vn}}



\maketitle

\begin{abstract}

We introduce Dual Face–Hair Retrieval (DFHR), a new mixed-modality dual-reference task in image retrieval where a query consists of a face image specifying identity and a hairstyle reference expressed as either an image or text. Unlike prior retrieval settings, DFHR requires cross-component reasoning between two semantically independent attributes—identity and hairstyle—originating from heterogeneous modalities. This formulation demands localized feature disentanglement, cross-modal semantic alignment, and mixed-modality composition within a unified embedding space. We construct DFHR-Bench, the first benchmark for mixed-modality face–hair retrieval, comprising over 180K annotated triplets across dual-image and image–text settings, built via a multi-stage annotation protocol ensuring semantic and identity integrity. We further propose MFHC (Multimodal Face–Hair Combiner), a unified framework that fuses disentangled identity and hairstyle embeddings through token injection and multi-view supervision. DFHR and DFHR-Bench together establish a new paradigm for identity-aware, attribute-controllable visual retrieval across modalities.
\end{abstract}  
\section{Introduction}
\label{sec:intro}

Mixed-modality has emerged as a new paradigm for image retrieval, unifying multiple query modalities within a single retrieval framework~\cite{huang2024dynamic}. Unlike conventional single-modality retrieval \cite{datta2008image}, where the query is expressed solely as an image or text \cite{rao2022does, qu2020context}, mixed-modality retrieval allows users to express intent through multiple complementary modalities, enabling more precise and expressive search. This capability becomes particularly crucial in fine-grained domains such as facial analysis, where visual appearance depends on intricate, localized structures that are often difficult to describe verbally. Designing a system that can interpret and integrate these diverse modalities coherently remains a key challenge for next-generation retrieval.

We introduce \textbf{Dual Face–Hair Retrieval (DFHR)}, a representative instance of mixed-modality image retrieval that captures the interplay between identity and style in facial images. In DFHR, the query consists of a face image, defining the subject’s identity, and a hair cue that specifies the desired hairstyle. Crucially, this hair cue can be provided in either visual form (an image) or linguistic form (a text description), reflecting the mixed nature of the problem. Allowing interchangeable modalities for the hair cue significantly broadens the expressiveness of user intent: an image can convey fine structural details, while text can specify abstract semantic attributes or styles visually unavailable. The retrieval system must therefore retrieve target images that preserve the same face identity while adopting the specified hairstyle, regardless of whether that hairstyle is defined visually or linguistically. This dual-constraint formulation offers a realistic and underexplored retrieval setting, one that unifies multimodal reasoning, identity preservation, and controllable appearance variation within a single framework.


\begin{figure}[t]
  \centering
  \includegraphics[width=0.95\textwidth]{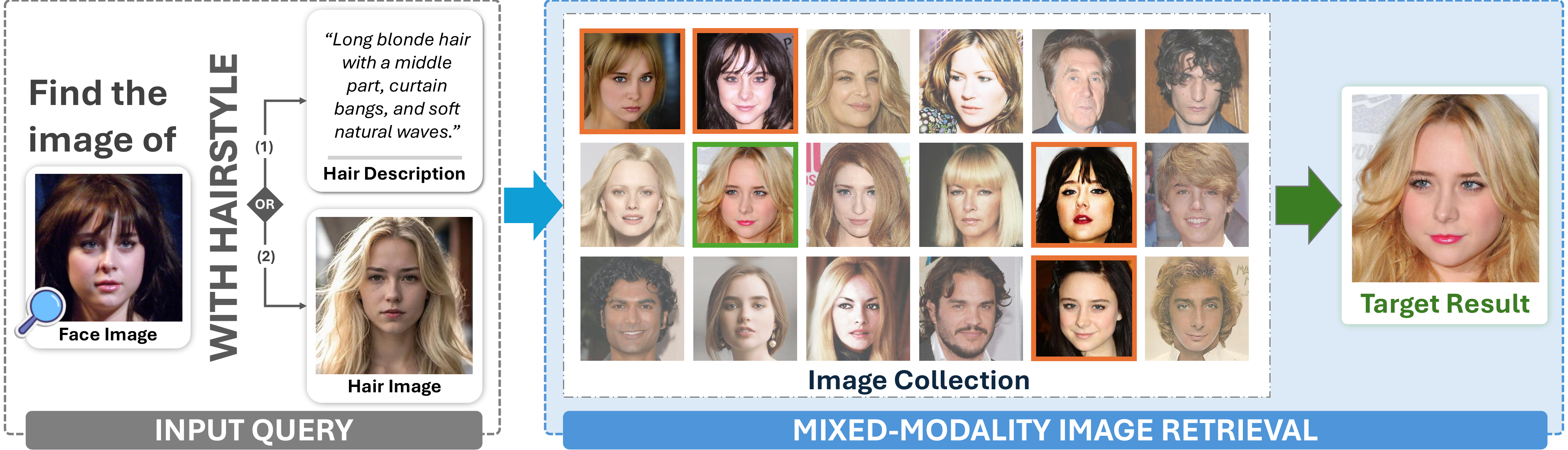}
  \caption{\textbf{Dual Face-Hair Retrieval.} 
  Given a face image and a hair cue expressed as either a textual description or a hair image, DFHR retrieves the image \textcolor{ForestGreen}{(green box)} that preserves the queried identity while adopting the target hairstyle. The retrieval is performed over a gallery exhibiting compound complexity from dual constraints: (1) multiple hairstyle variations of the same identity \textcolor{orange}{(orange box)}, and (2) numerous hairstyle look-alikes across different identities. Models that rely solely on a single attribute fail to capture this challenge.}
  \label{fig:teaser}
    \vspace{-15pt}
\end{figure}

Although conceptually related to prior retrieval paradigms, DFHR differs fundamentally from both \textit{Composed Image Retrieval (CIR)}~\cite{pic2word, isearle, lin2024fine, liu2024bi, wen2024simple} and \textit{Multiple Queries Image Retrieval (MQIR)} \cite{fernando2013mining, ji2025visual}. CIR operates on image reference-text modifier pairs, where the objective is to retrieve an image that resembles the reference but reflects a specified textual modification. MQIR aggregates multiple queries via images or texts to capture richer intent representations. DFHR departs from both in three key aspects. First, the query structure is a paired composition of a face image and a hair cue, rather than a reference-plus-modification format (CIR) or an unstructured multi-query aggregation (MQIR). Second, the task enforces dual semantic constraints, requiring simultaneous satisfaction of identity preservation and hairstyle conformity. Third, DFHR uniquely embraces cross-modal flexibility, as the hair cue may exist in either image or text form, bridging linguistic and visual domains within a unified retrieval process. This mixed representation capability is central to DFHR’s novelty, enabling a single retrieval framework to handle both visual and textual specifications of the same semantic factor.

As illustrated in Figure~\ref{fig:teaser}, DFHR presents distinct technical challenges rooted in \textit{localized feature disentanglement}. The model must isolate identity-specific facial features while suppressing hairstyle-related signals from the input face, and conversely, extract hairstyle attributes without leaking identity cues from the reference. This difficulty is compounded by the requirement for a \textit{unified mixed-modality embedding space}, in which visual and textual hairstyle cues must coexist coherently. Although Vision–Language Models (VLMs) \cite{clip} exhibit strong compositional reasoning in general image–text settings, extending them to fine-grained identity–hairstyle composition often leads to semantic drift, causing the fused representation to deviate from the intended combination. Ultimately, achieving robust cross-modal semantic alignment is essential, ensuring that a hairstyle expressed in text form conveys the same retrieval intent as one expressed in an image while maintaining fidelity to the queried identity.

As DFHR defines a new retrieval paradigm, no existing benchmark is capable of supporting its systematic study or fair evaluation. To fill this critical gap, we introduce DFHR-Bench, the first benchmark explicitly designed for mixed-modality face–hair retrieval, constructed at a scale surpassing existing datasets in the current CIR literature (Table \ref{tab:comp_benchmarks}). Under the extended subset, DFHR-Bench comprises up to 116K and 68K triplets for the Dual-Image and Image–Text settings, respectively, featuring rich identity, hairstyle, and modality combinations. Constructing such a dataset is non-trivial: unlike conventional visual attributes, hairstyle semantics are fine-grained, subjective, and entangled with identity, demanding careful disentanglement during annotation. To overcome these challenges, we devise a multi-stage annotation pipeline that combines expert-guided reasoning with procedural safeguards to ensure semantic consistency, identity integrity, and annotation quality at scale. This benchmark not only establishes the first standardized foundation for evaluating DFHR systems but also enables broader exploration of identity–style disentanglement, cross-modal alignment, and controllable visual retrieval.

\setlength{\tabcolsep}{5pt} 
\begin{table*}[t]
\centering
\caption{Comparison of representative composition-based image retrieval benchmarks.}
\label{tab:comp_benchmarks}
\resizebox{\textwidth}{!}{%
\begin{tabular}{@{} l | c | c | c c c | c c | l @{}}
\toprule
\textbf{Benchmark} & 
\textbf{\#Triplets} & 
\textbf{Pool Size} &
\textbf{ID-Aware} &
\textbf{Multi-GT} &
\textbf{Hard-Neg.} &
\textbf{Img--Img} & 
\textbf{Img--Text} & 
\textbf{Data Sources} \\
\midrule
CIRR (ICCV 2021) \cite{liu2021_cirr} & 4{,}148 & 2{,}316 &
\textcolor{red}{\xmark} & \textcolor{red}{\xmark} & \textcolor{red}{\xmark} &
\textcolor{red}{\xmark} & \textcolor{ForestGreen}{\cmark} &
NLVR2 \\
FashionIQ (CVPR 2021) \cite{wu2020_fashioniq} & 2{,}005 & 5{,}179 &
\textcolor{red}{\xmark} & \textcolor{red}{\xmark} & \textcolor{red}{\xmark} &
\textcolor{red}{\xmark} & \textcolor{ForestGreen}{\cmark} &
Shopping sites \\
CIRCO (ICCV 2023) \cite{baldrati2023zero} & 800 & 123{,}403 &
\textcolor{red}{\xmark} & \textcolor{ForestGreen}{\cmark} & \textcolor{red}{\xmark} &
\textcolor{red}{\xmark} & \textcolor{ForestGreen}{\cmark} &
MS COCO \\
\midrule
\rowcolor{hilite}
\textbf{DFHR-Bench (Ours)} & \textbf{68{,}760} & \textbf{116{,}283} &
\textcolor{ForestGreen}{\cmark} & \textcolor{ForestGreen}{\cmark} & \textcolor{ForestGreen}{\cmark} &
\textcolor{ForestGreen}{\cmark} & \textcolor{ForestGreen}{\cmark} &
\textbf{CelebA-HQ, FFHQ, Web Sources} \\
\bottomrule
\end{tabular}%
}
\end{table*}

\section{Related Work}
\label{sec:related-works}

\subsection{Multi-Modal Composed Representation.}
Pre-trained vision–language models (VLMs) have demonstrated strong cross-modal alignment through large-scale image–text supervision, such as CLIP \cite{clip} and ALIGN \cite{li2021align}. These models capture implicit visual–semantic correspondences, enabling strong zero-shot transfer in both recognition \cite{saha2024improved, mirza2024meta, yousaf2025enhancing} and retrieval \cite{xiao2025flair, koley2024you}. The recent \textit{textual inversion} \cite{cohen2022my, gal2022image, zhou2022learning} paradigm encodes a visual exemplar into pseudo-word tokens compatible with the textual embedding space of pretrained VLMs, supporting personalized and compositional representations for generation \cite{wei2023elite, gal2022image, zhang2023inversion} and retrieval \cite{pic2word, baldrati2023zero, isearle}. \textit{Composed Image Retrieval (CIR)} extends such multimodal reasoning by formulating the query as a combination of a reference image and a textual modifier \cite{vo2018composingtextimageimage, song2025comprehensive}. Early works rely on feature-fusion strategies \cite{liu2024bi, wen2024simple, huang2024dynamic, delmas2022artemis, chen2024fashionern}, while recent textual-inversion-based approaches map reference images into pseudo-word tokens for zero-shot composed image retrieval (ZS-CIR) \cite{pic2word, isearle, lin2024fine, suo2024knowledge}. Despite these advances, identity-aware composition from visual inputs remains largely unexplored. Existing CIR benchmarks focus on object- \cite{liu2021_cirr, baldrati2023zero} or fashion-level \cite{wu2020_fashioniq} attributes, overlooking scenarios involving a fixed human identity with a changing attribute. We bridge this gap by introducing mixed-modality composition learning with multi-view supervision for identity-aware retrieval within a unified framework.
We present the first face-centric benchmark that extends the query formulation to include visual guidance components for fine-grained dual-reference retrieval and evaluation.

\subsection{Generation-based Image Retrieval (GBIR).}
Advances in generative modeling, particularly GAN- and diffusion-based architectures, have motivated generative-based image retrieval (GBIR) \cite{wang2025generative}, where models generate samples approximating the retrieval target. In the context of DFHR, the intended identity–hairstyle composition could, in principle, be handled through generative synthesis. However, generative paradigms reveal inherent limitations. Hairstyle-transfer frameworks \cite{hairfast, stablehair, hairclipv2, styleyourhair, chung2025preserve} modify a source image to apply a new hairstyle but often distort facial geometry and lose subtle identity cues due to encoder bottlenecks and spatial remapping. ID-consistent generation \cite{papantoniou2024arc2face, han2024face, instantid} synthesize identity-aware images guided by text, yet their global conditioning entangles identity and hairstyle, hindering localized control. Although visually plausible, these methods remain unsuitable for this task.

\subsection{Facial Domain Datasets}
Large-scale facial datasets~\cite{karras2019_ffhq, wang2025_facebench, vggface2, LFWTech, celeba, webface260M} provide collections of faces with rich annotation schemes, while hairstyle-oriented datasets such as K-Hairstyle~\cite{kim2021_k_hairstyle} deliver fine-grained hairstyle labels for classification or generation. As no existing dataset targets identity-aware composed retrieval, we construct ours using established facial datasets. Specifically, we build upon CelebA-HQ~\cite{karras2018_celeba_hq} and FFHQ~\cite{karras2019_ffhq}, chosen for their high-resolution imagery, diversity, and well-defined identity metadata. COCO-PFS~\cite{messina2025_coco_pfs} similarly targets identity-aware retrieval, focusing on cross-modal name-based captions rather than our compositional identity–hairstyle setting, and also utilizes deepfake-based augmentation for robustness.

\section{Method}
\label{sec:dual-facehair}
\begin{figure*}
    \centering
    \includegraphics[width=1\linewidth]{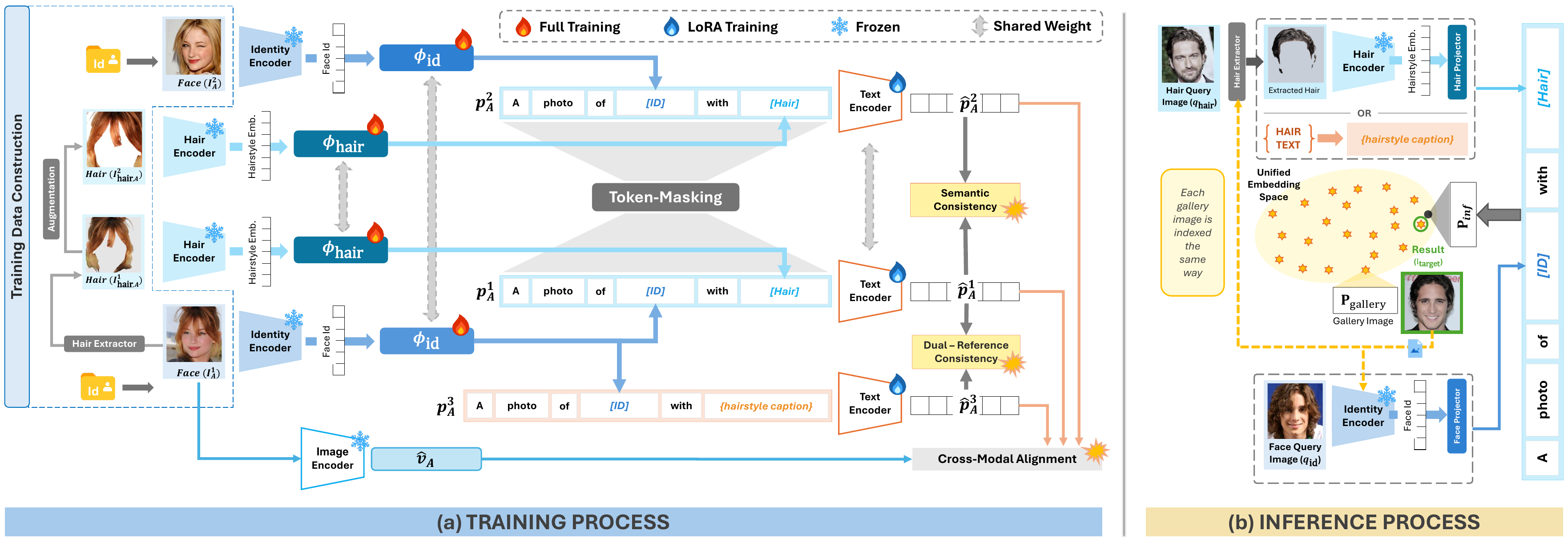}
    \caption{
    \textbf{Overview of MFHC framework.}
    (a) \textit{Training process.} Each person provides two facial images and their corresponding hairstyle views obtained via augmentation, along with a synthesized hairstyle caption (e.g. \texttt{\{hairstyle caption\}} = “Loose copper updo with wispy curtain bangs”).
    (b) \textit{Inference process.} Given a query consisting of a face image and a hairstyle reference—either an image or a text description, MFHC encodes both into a unified text-embedding space for retrieval by cosine similarity.
    }
    \label{fig:method-pipeline}
    \vspace{-10pt}
\end{figure*}

In this section, we present the Multimodal Face–Hair Combiner (MFHC), a unified framework designed for our DFHR task. We first detail the overall architecture in \S\ref{subsec:framework_overview}, then our multi-view supervision strategy (\S\ref{subsec:mixed_modal_supervision}), and finally the inference process in \S\ref{subsec:inference} (additional details in the Appendix~\ref{supp:method}).

\subsection{Framework Overview}
\label{subsec:framework_overview}

\noindent\textbf{Problem Formulation}.
We formulate DFHR as a mixed-modality dual-reference retrieval task.  
Given a gallery of face images $\mathcal{X} = \{x_i\}_{i=1}^{N}$, a query is defined as a pair $(q_{\text{id}}, q_{\text{hair}})$, where $q_{\text{id}}$ specifies the facial identity and $q_{\text{hair}}$ specifies the target hairstyle.  
The identity reference is always an image, while the hairstyle reference may appear in either visual or textual form, $q_{\text{hair}} \in \{q^{\text{img}}_{\text{hair}}, q^{\text{txt}}_{\text{hair}}\}$, reflecting the mixed-modality nature of the task.  
The objective is to retrieve an image $x^{*} \in \mathcal{X}$ that simultaneously matches the identity in $q_{\text{id}}$ and the hairstyle specified by $q_{\text{hair}}$, by maximizing the similarity between the composed query embedding $f(q_{\text{id}}, q_{\text{hair}})$ and the gallery representation $g(x)$.

\noindent\textbf{Proposed Approach.}
MFHC models the composed embedding $f(q_{\text{Id}}, q_{\text{Hair}})$ within a unified compositional space, trained on paired identity-hairstyle signals that express the same retrieval intent across modalities. Using a multi-view face dataset~\cite{karras2018_celeba_hq}, we sample two images of the same identity $(I^1_A, I^2_A)$ and extract a hairstyle crop from $I^1_A$ via segmentation~\cite{deng2020retinaface}. This crop is augmented with stochastic transformations~\cite{simclr} to produce two correlated hairstyle views $(I^1_{\text{hair},A}, I^2_{\text{hair},A})$. The pairs $(I^1_A, I^1_{\text{hair},A})$ and $(I^2_A, I^2_{\text{hair},A})$ therefore encode equivalent DFHR intents. Additionally, a textual hairstyle description $C_{\text{hair},A}$, synthesized using~\cite{bai2023qwen}, serves as a complementary cross-modal view.

\noindent\textbf{Model Architecture.} MFHC employs a dual-branch token mapping network parameterized by two lightweight MLPs \(\phi_{\text{id}}\) and \(\phi_{\text{hair}}\) to encode identity and hairstyle separately.
Given \(I_{\text{id}}\) and \(I_{\text{hair}}\), we extract embeddings \(V_{\text{id}}\) and \(V_{\text{hair}}\) derived from frozen pretrained Arcface \cite{arcface} and CLIP image encoder models \cite{clip} for identity and hairstyle encoder, respectively.
The mappers produce pseudo-token sequences
\(T_{\text{id}}=\phi_{\text{id}}(V_{\text{id}})\) and
\(T_{\text{hair}}=\phi_{\text{hair}}(V_{\text{hair}})\),
where \(T_{\text{id}}, T_{\text{hair}} \in \mathbb{R}^{n \times d_{\text{text}}}\) are spans in CLIP’s text-token space \footnote{We employ ViT-B/32 published by OpenAI.}, with \(n\) injected tokens and token dimension \(d_{\text{text}}\). 

\noindent\textbf{Multi-View Textual Embeddings.}
We form three textual compositions for each person \(A\) based on two prompt templates:
\(P_{\text{vis}}=\texttt{a photo of [id] with [hair]}\)
and
\(P_{\text{txt}}=\texttt{a photo of [id] with }\{caption\}\).
A token injection operator \(\mathcal{I}(\cdot)\) is used to form the composed token sequence by 
replacing the indicator tokens \texttt{[id]} and \texttt{[hair]} in the template with the corresponding token spans \(T_{\text{id}}\) and \(T_{\text{hair}}\).
For the text-driven case, \(\mathcal{I}\) appends the caption \(C_{\text{hair},A}\) directly in place of \(\{caption\}\).
Formally:
\[
\begin{aligned}
p_A^{1} & = \mathcal{I}(P_{\text{vis}};\,T_{\text{id}}^{1},T_{\text{hair}}^{1}),\\
p_A^{2} & = \mathcal{I}(P_{\text{vis}};\,T_{\text{id}}^{2},T_{\text{hair}}^{2}),\\
p_A^{3} & = \mathcal{I}(P_{\text{txt}};\,T_{\text{id}}^{1},C_{\text{hair},A}).
\end{aligned}
\]
Each token sequence is encoded by the LoRA-adapted CLIP text encoder \cite{peft}. to yield \(\hat{p}_A^{k} = \mathrm{CLIP}_{\text{text}}(p_A^{k})\),
forming a set of multi-view textual embeddings that capture both visual and linguistic hairstyle semantics for person \(A\).

\subsection{Mixed-Modality Multi-View Supervision}
\label{subsec:mixed_modal_supervision}

To address mixed-modality representation, MFHC adopts a \emph{multi-view supervision} strategy that jointly aligns and regularizes the learned representations. Using the multi-view embeddings $\{\hat{p}^1_A, \hat{p}^2_A, \hat{p}^3_A\}$ together with the image feature $\hat{v}_A$, MFHC enforces comprehensive cross-modal agreement and factor disentanglement. The overall formulation combines three complementary objectives: (1) \emph{cross-modal alignment} to anchor multimodal signals in a shared space, (2) \emph{semantic consistency} to maintain factor-wise coherence, and (3) \emph{hard negative compositionality} to encourage discriminative separation.

\noindent\textbf{Cross-Modal Alignment.}
To unify representations across modalities, we adopt a symmetric contrastive objective that aligns each text embedding \(\hat{p}^{k}_A\) (\(k=1,2,3\)) with its paired image embedding \(\hat{v}_A\).
For a batch \(\mathcal{B}\) of size \(|\mathcal{B}|\), the directional objectives are defined as:
\begin{equation}
\mathcal{L}_{\text{t2i}}
= -\frac{1}{3|\mathcal{B}|}
\sum_{k=1}^{3}
\sum_{i\in\mathcal{B}}
\log
\frac{\exp(\tau\,\hat{p}_{i}^{k\top}\hat{v}_i)}
{\sum_{j\in\mathcal{B}}\exp(\tau\,\hat{p}_{i}^{k\top}\hat{v}_j)},
\end{equation}
\begin{equation}
\mathcal{L}_{\text{i2t}}
= -\frac{1}{3|\mathcal{B}|}
\sum_{k=1}^{3}
\sum_{i\in\mathcal{B}}
\log
\frac{\exp(\tau\,\hat{v}_{i}^{\top}\hat{p}_{i}^{k})}
{\sum_{j\in\mathcal{B}}\exp(\tau\,\hat{v}_{i}^{\top}\hat{p}_{j}^{k})},
\end{equation}
where \(\tau\) is a temperature parameter.
The final symmetric alignment loss is:
\begin{equation}
\mathcal{L}_{\text{align}}
= \tfrac{1}{2}(\mathcal{L}_{\text{t2i}} + \mathcal{L}_{\text{i2t}}).
\end{equation}
In this way, we jointly align all three compositional textual views of visual counterparts, promoting consistent mixed-modal embedding learning.

\noindent\textbf{Semantic Consistency.}
While the self-attention mechanism enables cross-token interaction between the injected \texttt{[id]} and \texttt{[hair]} embeddings to form a unified composition, such interaction may also induce \emph{semantic drift}, causing the identity and hairstyle representations to become entangled and lose factor-specific fidelity.
To regularize this interaction, we introduce a factorized regularization technique applied via \emph{token masking}. This approach allows us to selectively expose different textual spans and independently enforce their consistency. Specifically, we mask the text template to produce three variants that isolate distinct semantic spans (i) identity \(M_{\text{id}}(p)\): ``\texttt{a photo of [id]}'', (ii) hairstyle \(M_{\text{hair}}(p)\): ``\texttt{a photo of person with [hair]}'', and (iii) full composition \(M_{\text{full}}(p)\): ``\texttt{a photo of [id] with [hair]}''.
Each masked sequence is encoded by the CLIP text encoder \(E(\cdot)=\mathrm{CLIP}_{\text{text}}(\cdot)\), yielding \(\hat{p}=E(M_{\{\cdot\}}(p))\).
We enforce consistency across masked views and add a Dual-Reference Consistency (DRC) term that aligns the visual composition with its caption-based counterpart:
\[
\left\{
\begin{aligned}
\mathcal{L}_{\text{id}}   &= 1 - \mathrm{sim}\!\big(E(M_{\text{id}}(p_A^{1})),\, E(M_{\text{id}}(p_A^{2}))\big),\\
\mathcal{L}_{\text{hair}} &= 1 - \mathrm{sim}\!\big(E(M_{\text{hair}}(p_A^{1})),\, E(M_{\text{hair}}(p_A^{2}))\big),\\
\mathcal{L}_{\text{comp}} &= 1 - \mathrm{sim}\!\big(E(M_{\text{full}}(p_A^{1})),\, E(M_{\text{full}}(p_A^{2}))\big),\\
\mathcal{L}_{\text{DRC}}  &= 1 - \mathrm{sim}\!\big(E(p_A^{1}),\, E(p_A^{3})\big).
\end{aligned}
\right.
\]
Here, \(\mathrm{sim}\) is cosine similarity. The overall semantic objective is:
\begin{equation}
\mathcal{L}_{\text{sem}} =
\lambda_{\text{id}}\mathcal{L}_{\text{id}} +
\lambda_{\text{hair}}\mathcal{L}_{\text{hair}} +
\lambda_{\text{comp}}\mathcal{L}_{\text{comp}} +
\lambda_{\text{drc}}\mathcal{L}_{\text{DRC}}.
\end{equation}
This factorized regularization yields view-invariant, factor-aware representations while maintaining compositional coherence in the shared token space.

\noindent\textbf{Hard Negative Compositionality.}
To further strengthen factor disentanglement, we incorporate a hard-negative compositional loss that penalizes invalid identity–hairstyle pairings
. Within each batch, we construct negatives by shuffling one factor while keeping the other fixed, producing
\((T_{\text{id}}^{\text{neg}}, T_{\text{hair}})\) and
\((T_{\text{id}}, T_{\text{hair}}^{\text{neg}})\).
These yield negative text embeddings
\(\hat{p}_{A}^{k-}\),
contrasted against the positive composition
\(\hat{p}_{A}^{k+}\) paired with its image embedding \(\hat{v}_A\).
The objective enforces higher similarity for valid compositions than for mismatched ones:
\begin{equation}
\label{eq:hardneg_fixed}
\mathcal{L}_{\text{hardneg}}
= \frac{1}{3|\mathcal{B}|}
\sum_{A\in\mathcal{B}}
\sum_{k=1}^{3}
\alpha_{k}\,
\max\!\Big(0,\,
m_{k}
+ \cos\!\big(\hat{p}_{A}^{k-}, \hat{v}_A\big)
- \cos\!\big(\hat{p}_{A}^{k+}, \hat{v}_A\big)
\Big).
\end{equation}

where \(m_k\) is the margin and \(\alpha_k\) balances the three compositional views.
This loss encourages MFHC to prefer valid identity–hairstyle pairs over mismatched ones, reinforcing factor disentanglement in the shared embedding space.

\subsection{Inference with MFHC}
\label{subsec:inference}

At inference (as illustrated in Figure \ref{fig:method-pipeline}), MFHC operates entirely within the unified text embedding space. Given a query $(q_{\text{id}}, q_{\text{hair}})$, where $q_{\text{hair}}$ can be $q^\text{img}_{\text{hair}}$ or $q^\text{txt}_{\text{hair}}$. the model encodes both factors into their respective token spans: $T_{\text{id}} = \phi_{\text{id}}(f_{\text{id}}(q_{\text{id}}))$ and $T_{\text{hair}} = \phi_{\text{hair}}(f_{\text{hair}}(q_{\text{hair}}))$.  
These tokens are injected into the fixed textual template "\texttt{a photo of [id] with [hair]}" and processed by the LoRA-adapted CLIP text encoder to produce the query embedding $\hat{p}_q = \mathrm{CLIP}_{\text{text}}(\mathcal{I}(P_{\text{vis}};\,T_{\text{id}},T_{\text{hair}}))$.  
Each gallery image $x_i \in \mathcal{X}$ is embedded through the same pathway, yielding $\hat{p}_{x_i} = \mathrm{CLIP}_{\text{text}}(\mathcal{I}(P_{\text{vis}};\,T_{\text{id},i},T_{\text{hair},i}))$. We rank the candidate images according to cosine similarity $\mathrm{sim}(\hat{p}_q, \hat{p}_{x_i})$.

\section{DFHR-Bench Dataset}
\label{sec:benchmark}

\begin{figure*}[t]
    \centering
    \includegraphics[width=1\linewidth]{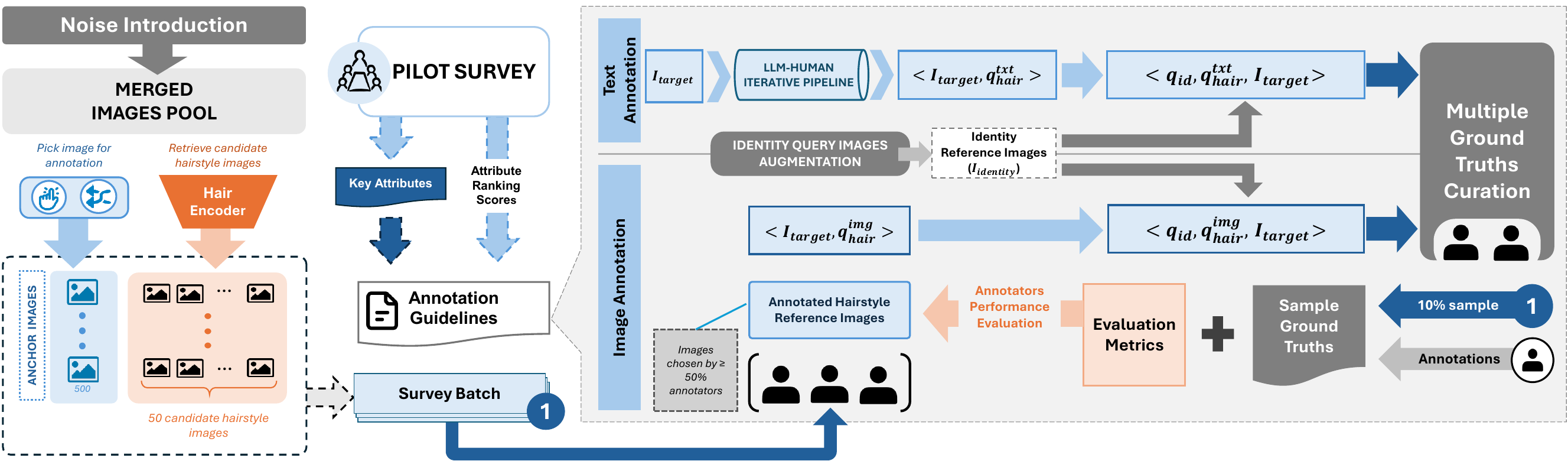}
    \caption{\textbf{Benchmark Construction Protocol.} DFHR-Bench follows a three-stage protocol with dual annotation streams. \textit{The Pre-Annotation Stage} conducts a pilot study and designs survey batches. \textit{The Annotation Stage} runs parallel text- and image-based pipelines combining LLM generation, human verification, and quality assessment. \textit{The Post-Annotation Stage} finalizes triplets through identity augmentation, multi-target curation, and synthesis-based noise injection.}
    \label{fig:overall-pipeline}
\end{figure*}

\begin{figure}[tb]
\centering
\resizebox{0.9\linewidth}{!}{%
\begin{minipage}{\linewidth}
  \centering
  \begin{subfigure}[t]{0.48\linewidth}
    \centering
    \includegraphics[width=\linewidth]{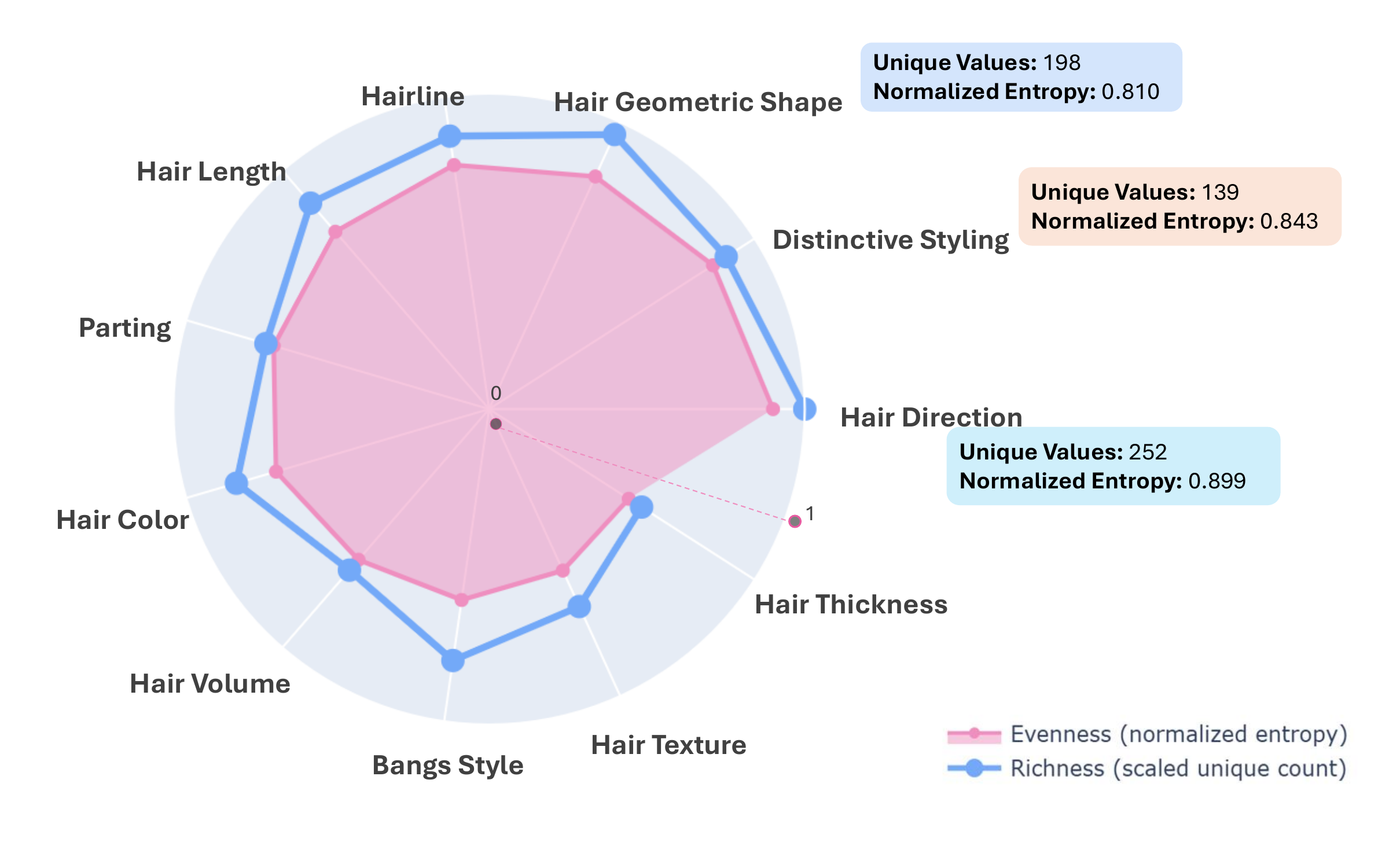}
    \caption{Semantic diversity visualization of hairstyle attributes across all annotated identities.}
    \label{fig:benchmark_diversity}
  \end{subfigure}
  \hfill
  \begin{subfigure}[t]{0.48\linewidth}
    \centering
    \includegraphics[width=\linewidth]{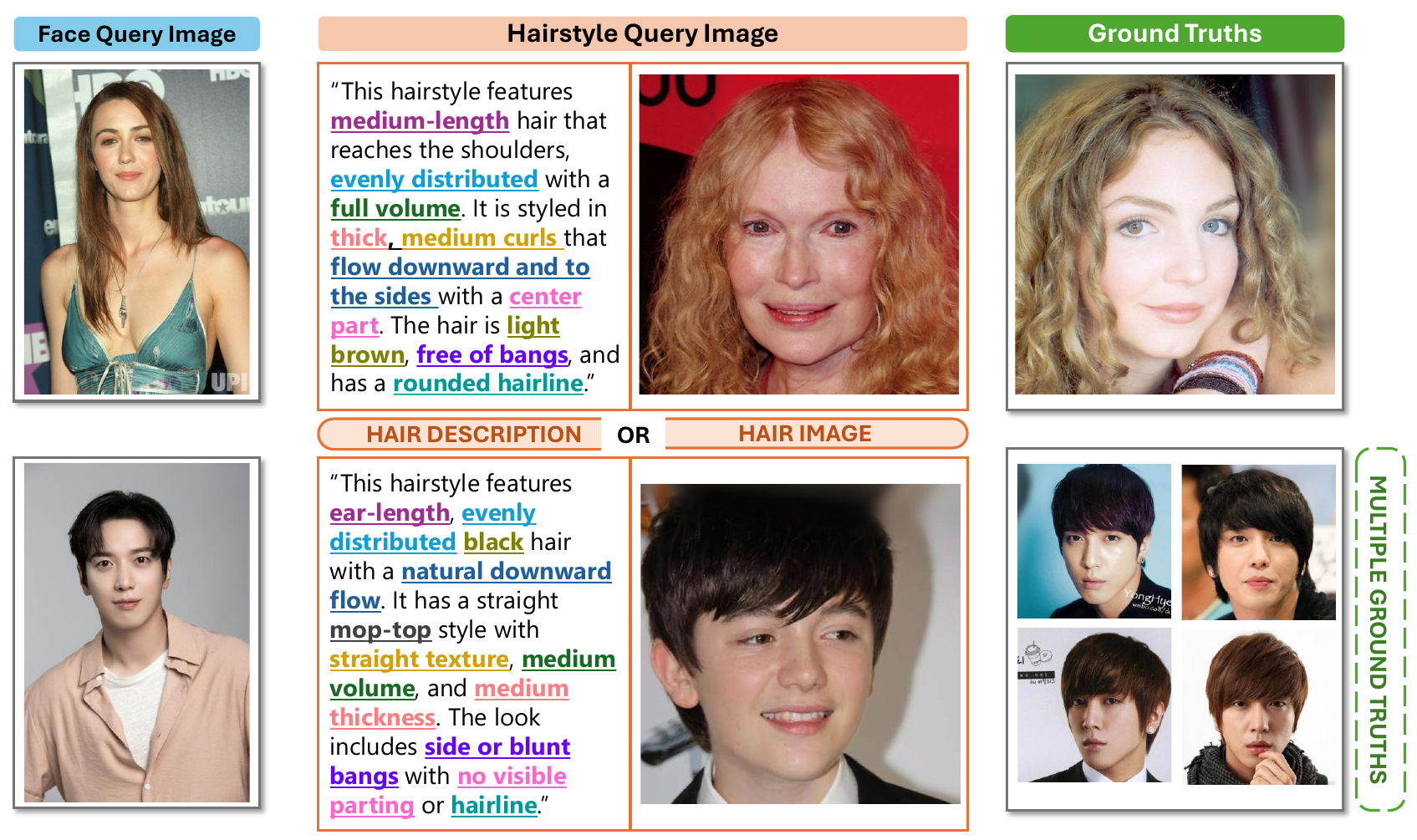}
    \caption{Some triplet representatives in both settings from our benchmark dataset.}
    \label{fig:benchmark_examples}
  \end{subfigure}
\end{minipage}%
}
\caption{\textbf{Benchmark dataset analysis.} Semantic diversity (left) and qualitative examples (right).}
\label{fig:benchmark_overview}
\end{figure}

We organize DFHR-Bench into triplets with multiple ground truths, $\langle q_{\text{id}}, q_{\text{hair}}, \mathcal{G} \rangle$, following the CIRCO formulation~\cite{baldrati2023zero}, where $\mathcal{G}$ denotes the set of valid target images.
We construct three annotated subsets:
(i) \textit{Image–Image}, with a hair image as the hairstyle reference;
(ii) \textit{Image–Text}, with hairstyle provided as a textual description; and
(iii) \textit{Bimodal Alignment}, which extends the formulation into a quadruplet containing both hair conditions for the same targets.

\subsection{Predefined Criteria}
\label{sec:benchmark-criteria}
DFHR-Bench is designed according to four principles:
\textbf{(i) \textit{Conceptual Alignment}} ensures a consistent understanding of hairstyle concepts among annotators to address hairstyle intrinsic subjectivity.
\textbf{(ii) \textit{Diversity}} ensures rich variation across visual appearances and descriptive expressions, fostering heterogeneous conditions for evaluating identity–hairstyle composition.
\textbf{(iii) \textit{Annotation Precision}} emphasizes high-quality, bias-mitigated labeling to produce reliable ground truths, minimizing artifacts introduced by fatigue, gender imbalance, or task complexity \cite{bias_control, IRR_assessment, bias_mitigation, research_bias}.
\textbf{(iv) \textit{Multi-Modality}} integrates both visual and textual inputs, reflecting the dual-condition query form $\langle q_\text{id}, q_\text{hair} \rangle$ used in our benchmark. 
Detailed discussions of each criterion, including challenges and mitigation strategies, are provided in the Appendix~\ref{benchmark-quality}.

\begin{figure}[t] 
\centering

\begin{minipage}[b]{0.52\textwidth}
\centering

\resizebox{\linewidth}{!}{
\begin{tabular}{lcc}
\toprule
\multicolumn{3}{c}{\textbf{General Statistics}} \\
\midrule
\# Identities & \multicolumn{2}{l}{500} \\
\quad \(\llcorner\)\textcolor{darkgray}{Male} 
& \multicolumn{2}{l}{\textcolor{darkgray}{200 \textsc{(40\%)}}} \\
\quad \(\llcorner\)\textcolor{darkgray}{Female} 
& \multicolumn{2}{l}{\textcolor{darkgray}{300 \textsc{(60\%)}}} \\
\# Images in Retrieval Pool & \multicolumn{2}{l}{105,998} \\
\quad \(\llcorner\)\textcolor{darkgray}{CelebA-HQ}~\cite{karras2018_celeba_hq} 
& \multicolumn{2}{l}{\textcolor{darkgray}{69,059 \textsc{(65.15\%)}}} \\
\quad \(\llcorner\)\textcolor{darkgray}{FFHQ}~\cite{karras2019_ffhq} 
& \multicolumn{2}{l}{\textcolor{darkgray}{27,659 \textsc{(26.09\%)}}} \\
\quad \(\llcorner\)\textcolor{darkgray}{Deepfake}  
& \multicolumn{2}{l}{\textcolor{darkgray}{9,280 \textsc{(8.75\%)}}} \\
\# Face Images (\textcolor{RoyalBlue}{off.}/ext.) 
& \multicolumn{2}{l}{\textcolor{RoyalBlue}{1,889} / 13,752} \\
\# Hair Images (\textcolor{RoyalBlue}{off.}/ext.) 
& \multicolumn{2}{l}{\textcolor{RoyalBlue}{3,040} / 3,762} \\
\# Text Descriptions (\textcolor{RoyalBlue}{off.}/ext.) 
& \multicolumn{2}{l}{\textcolor{RoyalBlue}{2,310} / 2,475} \\
Avg. Ground Truths per Identity 
& \multicolumn{2}{l}{2.66 images} \\
Bimodal Alignment Subset 
& \multicolumn{2}{l}{1,279 triplets} \\
\midrule
\multicolumn{3}{c}{\textbf{Retrieval Settings}} \\
\midrule
\textbf{Statistic} & \textbf{Image – Image} & \textbf{Image – Text} \\
\midrule
\# Triplets (\textcolor{RoyalBlue}{off.}/ext.) 
& \textcolor{RoyalBlue}{6,246} / 116,283 
& \textcolor{RoyalBlue}{4,162} / 68,760 \\
\# Identities (\textcolor{RoyalBlue}{off.}/ext.) 
& \textcolor{RoyalBlue}{477} / 492 
& \textcolor{RoyalBlue}{482} / 495 \\
Avg. \# Face cues/ID (\textcolor{RoyalBlue}{off.}/ext.) 
& \textcolor{RoyalBlue}{3.78} / 27.82 
& \textcolor{RoyalBlue}{3.67} / 27.78 \\
Avg. \# Hair cues/ID (\textcolor{RoyalBlue}{off.}/ext.) 
& \textcolor{RoyalBlue}{6.81} / 8.36 
& \textcolor{RoyalBlue}{4.79} / 5.00 \\
\bottomrule
\end{tabular}
}
\captionof{table}{
\textbf{Benchmark Dataset Statistics.} The upper section reports overall dataset statistics, while the lower section summarizes the two retrieval configurations. The dataset includes a refined \textcolor{RoyalBlue}{\textit{official}} subset for standardized evaluation and an \textit{extended} version with broader, less constrained samples.
}
\label{tab:benchmark_stats}

\end{minipage}
\hfill
\begin{minipage}[b]{0.45\textwidth}
\centering
\includegraphics[width=\linewidth]{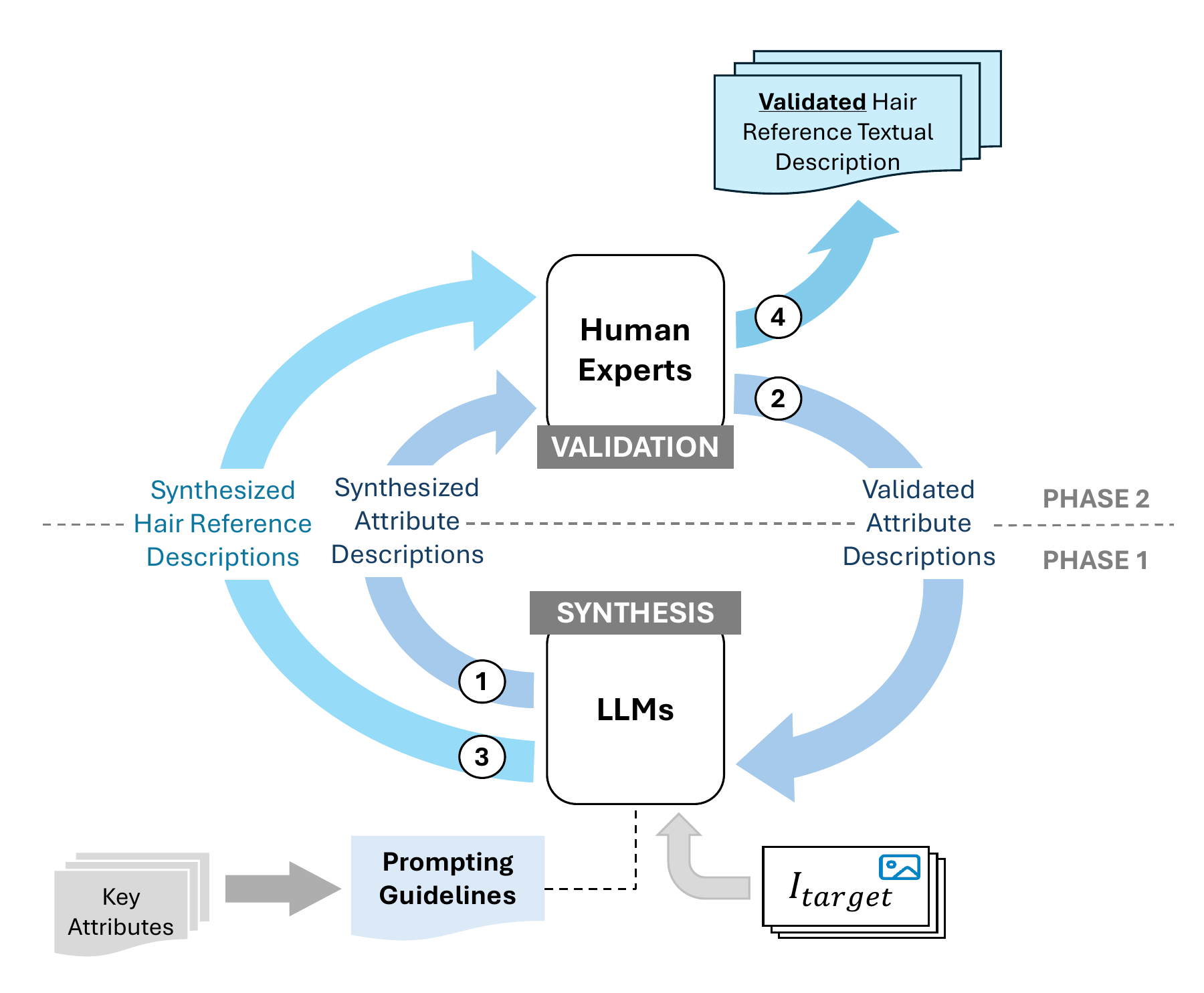}
\captionof{figure}{
\textbf{Text Reference Pipeline.}
A two-phase iterative pipeline for generating validated hair-reference descriptions via LLM synthesis and human validation. The first loop (Steps 1–2) produces attribute descriptions, while the second loop (Steps 3–4) composes them into complete textual hair cues.
}
\label{fig:text-pipeline}
\end{minipage}
\end{figure}

\subsection{Dataset Construction Methodology}
\label{sec:benchmark-construction-methodology}
We establish a structured annotation protocol that leverages human visual reasoning for hairstyle discrimination, complemented by verification and refinement strategies to minimize labor-induced errors. 
Figure~\ref{fig:overall-pipeline} illustrates the benchmark construction workflow covering the \textit{Pre-Annotation}, dual \textit{Annotation Pipelines}, and \textit{Post-Annotation} stages, detailed in the following subsections. Additional implementation and procedural details are provided in Appendix~\ref{sec:benchmark-construction}.

\noindent\textbf{Pre-Annotation Stage.}
To explore how people perceive hairstyle similarity, we conducted a pilot study with ten human experts who evaluated hairstyle pairs and explained their reasoning. From their responses, we distilled key hairstyle attributes and formalized them into a set of core annotation guidelines for later stages (see Appendix~\ref{sec:attribute-clarification}).
We further built a 100K-image pool by merging CelebA-HQ \cite{karras2018_celeba_hq} (for identity clustering) and FFHQ \cite{karras2019_ffhq} (for hairstyle diversity). From this pool, 500 anchor images representing distinct CelebA-HQ \cite{karras2018_celeba_hq} identities were selected as $I_{\text{target}}$ and used to initialize the subsequent annotation stage. 

\noindent\textbf{Image Annotation Pipeline.}
The image pipeline aims to identify visually similar hairstyles for each input anchor image, denoted as pairs of $\langle q^\text{img}_{\text{hair}}, I_{\text{target}} \rangle$. We employ a dedicated hair retriever to retrieve the top 50 candidate images that exhibit similar hairstyle features to those of the anchor. The retriever leverages a learned hair encoder, detailed in Appendix~\ref{sec:hair-encoder-implementation}.
For each query, an average of 3.4 annotators are instructed to select the valid hairstyle references from the candidate pool following predefined guidelines. Their reliability is assessed via qualification tests on 10\% of the queries. The final reference $q^\text{img}_{\text{hair}}$ is determined via majority voting, requiring consensus from over 50\% of annotators. The resulting annotations yield an AC1~\cite{ac1} of 0.66 and 0.78 agreement, reflecting substantial consistency.

\noindent\textbf{Text Annotation Pipeline.}
Complementing the image annotation pipeline, the text annotation pipeline generates fine-grained hairstyle descriptions for the same anchor set. As illustrated in Figure~\ref{fig:text-pipeline}, we adopt a hybrid human–LLM \footnote{We use GPT-4o for this work} annotation strategy operating in an iterative two-phase cycle. Large language models (LLMs) are employed to produce initial textual candidates, which are then refined and validated by human annotators to ensure accuracy and consistency. The pipeline aims at constructing paired samples \(\langle q^\text{txt}_{\text{hair}}, I_{\text{target}} \rangle\) that capture hairstyle semantics.

\noindent\textbf{Post-Annotation Stage.} 
After collecting hairstyle references, we refine and finalize the benchmark triplets through a series of post-processing steps.
Each anchor’s identity reference $q_{\text{id}}$ is expanded by retrieving additional images of the same individual from public sources\footnote{Images crawled from Google Image for research use.}. These images undergo face detection and identity verification using \textit{InsightFace}~\cite{arcface} to ensure precise alignment with CelebA-HQ identities. Low-resolution samples are automatically filtered, followed by manual inspection to ensure visual appropriateness.
We then organize all components into triplets, curate multiple valid ground truths by adding visually consistent hairstyle images from the same identity, and apply controlled image synthesis to introduce distractors and noise, which enhances benchmark robustness.


\subsection{DFHR-Bench Statistics}
\label{sec:benchmark-analysis}
DFHR-Bench comprises 500 identities, with an average of 2.66 ground-truth images per identity.
We define Image-Image and Image-Text protocols with 6.2K and 4.2K triplets in the official set, scaling to 116K and 69K in the extended one. We also support analysis on cross-modality with a 1.2K-triplet bimodal subset.
Table~\ref{tab:benchmark_stats} summarizes the general statistics of the dataset, while Figure~\ref{fig:benchmark_diversity} illustrates its semantic diversity, highlighting balanced coverage and rich variability across hairstyle attributes. Representative examples are shown in Figure~\ref{fig:benchmark_examples}, with additional samples provided in Appendix~\ref{sec:qualitative_examples}.


\begin{table}[t]
\centering
\caption{
\textbf{Evaluation on the Bimodal Alignment Subset}. We evaluate the modality gap between visual and textual hairstyle references. We also compare representative methods for each task, which do not support mixed-modality queries. The best and second-best results are highlighted in \textbf{bold} and \underline{underlined}, respectively.
}
\label{tab:dual_retrieval}
\resizebox{\textwidth}{!}{%
\begin{tabular}{@{} l c *{7}{c} | *{7}{c} @{}}
\toprule
\multirow{2}{*}{Method} & \multirow{2}{*}{Venue} &
  \multicolumn{7}{c|}{Image + Image Query (\%)} &
  \multicolumn{7}{c}{Image + Text Query (\%)} \\
\cmidrule(lr){3-9} \cmidrule(lr){10-16}
 &  & R@1 & R@3 & R@5 & P@1 & P@3 & P@5 & mAP
   & R@1 & R@3 & R@5 & P@1 & P@3 & P@5 & mAP \\
   
\midrule
 
InstantID \cite{instantid} & arXiv'24 & - & - & - & - & - & - & -
& 7.58 & 15.64 & 24.86 & 7.58 & 6.93 & 7.04 & 8.43 \\
withAnyone \cite{withanyone} & arXiv'24 & - & - & - & - & - & - & -
& 3.34 & 9.30 & 15.09 & 3.34 & 3.83 & 4.19 & 4.49 \\
ConsistentID \cite{consistentid} & arXiv'24 & - & - & - & - & - & - & -
& 9.23 & 19.94 & \underline{28.07} & 9.23 & 8.68 & 8.19 & 10.36 \\
Face-IP-Adapter \cite{han2024face} & ECCV'24 & - & - & - & - & - & - & -
& 10.63 & 20.17 & 26.82 & 10.63 & \underline{9.51} & \underline{8.71} & \underline{10.76} \\

HairCLIPv2 \cite{hairclipv2} & ICCV'23 & 4.38 & 10.56 & 20.02 & 4.38 & 4.25 & 5.22 & 5.48
& - & - & - & - & - & - & - \\
HairFast \cite{hairfast} & NeurIPS'24 & 6.33 & 14.00 & 26.19 & 6.33 & 5.79 & 6.94 & 7.49
& - & - & - & - & - & - & - \\
StableHair \cite{stablehair} & AAAI'25 & 5.71 & 13.29 & 22.99 & 5.71 & 5.58 & 6.10 & 6.67
& - & - & - & - & - & - & - \\

ZS-CIR \cite{ZS-CIR} & BMVC'23 & - & - & - & - & - & - & -
& 2.42 & 6.18 & 8.37 & 2.42 & 2.32 & 2.14 & 2.19 \\
Pic2Word \cite{pic2word} & CVPR'23 & - & - & - & - & - & - & -
& 2.81 & 4.30 & 5.79 & 2.81 & 1.85 & 1.64 & 1.87 \\
\midrule

 Face-only & - & 7.51 & 14.07 & 18.76 & 7.51 & 6.59 & 5.69 & 7.13
            & 7.51 & 14.07 & 18.76 & 7.51 & 6.59 & 5.69 & 7.13 \\
 Hair-only & - & 0.24 & 0.79  & 2.05  & 0.24 & 0.34 & 0.46 & 0.39
            & 0.00 & 0.00  & 0.00  & 0.00 & 0.00 & 0.00 & 0.00 \\
 Early Fusion & - & 9.14 & 20.09 & 34.99 & 9.14 & 8.43 & \underline{9.49} & 10.37
             & 7.97 & 16.89 & 27.29 & 7.97 & 7.22 & 7.63 & 8.42 \\
 Late Fusion & - & \underline{15.05} & \underline{29.55} & \underline{35.15} & \underline{15.05} & \underline{11.24} & 8.86 & \underline{12.57}
             & \underline{11.42} & \underline{21.42} & 27.68 & \underline{11.42} & 8.99 & 7.66 & 9.82 \\

\midrule
\rowcolor{hilite}

 \textbf{Ours (MFHC)} & - &  \textbf{24.82} & \textbf{42.63} & \textbf{50.99} & \textbf{24.82} & \textbf{18.70} & \textbf{15.29} & \textbf{22.63} 
 & \textbf{14.70} & \textbf{26.51} & \textbf{35.26} & \textbf{14.70} & \textbf{12.12} & \textbf{10.70} & \textbf{13.31} \\
\bottomrule
\end{tabular}%
}

\end{table}

\begin{table}[t]
\centering
\setlength{\tabcolsep}{2pt}
\renewcommand{\arraystretch}{0.95}

\begin{minipage}[t]{0.49\linewidth}
\centering
\captionsetup{type=table}
\caption{
\textbf{Evaluation on the \textit{Image+Image} official set.}
The best and second-best results are shown in \textbf{bold} and \underline{underlined}.
}
\label{tab:dual_retrieval_imgimg}
\resizebox{\linewidth}{!}{%
\begin{tabular}{@{} l *{3}{c} | *{3}{c} | c @{}}
\toprule
Method & R@1 & R@3 & R@5 & P@1 & P@3 & P@5 & mAP \\
\midrule
\multicolumn{8}{@{}l}{\textbf{Fusion-Based}} \\
Face-only     & 6.52 & 12.38 & 16.87 & 6.52 & 5.71 & 4.91 & 5.89 \\
Hair-only     & 0.43 & 0.98 & 1.42 & 0.43 & 0.35 & 0.30 & 0.32 \\
Early Fusion  & 7.91 & 20.03 & \underline{35.51} & 7.91 & 8.07 & \underline{9.39} & 9.33 \\
Late Fusion   & \underline{13.46} & \underline{28.96} & 34.52 & \underline{13.46} & \underline{11.48} & 9.05 & \underline{11.90} \\
\midrule
\multicolumn{8}{@{}l}{\textbf{Generation-Based}} \\
HairCLIPv2 & 4.02 & 10.26 & 19.26 & 4.02 & 3.98 & 4.85 & 4.88 \\
HairFast & 5.28 & 13.35 & 24.72 & 5.28 & 5.19 & 6.46 & 6.59 \\
StableHair & 4.77 & 11.56 & 21.20 & 4.77 & 4.70 & 5.74 & 5.98 \\
\midrule
\rowcolor{hilite}
Ours & \textbf{24.72} & \textbf{42.75} & \textbf{52.43} & 
\textbf{24.72} & \textbf{18.50} & \textbf{15.17} & \textbf{22.77} \\
\bottomrule
\end{tabular}}
\end{minipage}
\hfill
\begin{minipage}[t]{0.49\linewidth}
\centering
\captionsetup{type=table}
\caption{
\textbf{Evaluation on the \textit{Image+Text} official set.}
The best and second-best results are shown in \textbf{bold} and \underline{underlined}.
}
\label{tab:dual_retrieval_imgtxt}
\resizebox{\linewidth}{!}{%
\begin{tabular}{@{} l *{3}{c} | *{3}{c} | c @{}}
\toprule
Method & R@1 & R@3 & R@5 & P@1 & P@3 & P@5 & mAP \\
\midrule
\multicolumn{8}{@{}l}{\textbf{Fusion-Based}} \\
Face-only & 7.76 & 14.73 & 20.01 & 7.76 & 6.87 & 6.14 & 7.23 \\
Hair-only & 0.00 & 0.07 & 0.07 & 0.00 & 0.02 & 0.02 & 0.01 \\
Early Fusion & 8.58 & 18.40 & 29.26 & 8.58 & 7.86 & 8.28 & 8.74 \\
Late Fusion & \underline{12.37} & \underline{22.56} & 28.59 & 
\underline{12.37} & 9.57 & 8.09 & 10.26 \\
\midrule
\multicolumn{8}{@{}l}{\textbf{Generation-Based}} \\
InstantID & 6.17 & 15.02 & 23.71 & 6.17 & 6.46 & 6.69 & 7.51 \\
withAnyone & 3.72 & 9.15 & 16.24 & 3.72 & 3.91 & 4.50 & 4.64 \\
ConsistentID & 10.19 & 22.03 & \underline{30.54} & 10.19 & 9.51 & 8.89 & 10.99 \\
Face-IP-Adapter & 11.46 & 20.95 & 28.28 & 11.46 & \underline{9.79} & \underline{9.14} & \underline{11.34} \\
\midrule
\multicolumn{8}{@{}l}{\textbf{CIR}} \\
ZS-CIR & 2.79 & 6.63 & 8.82 & 2.79 & 2.61 & 2.29 & 2.55 \\
Pic2Word & 2.96 & 5.24 & 6.61 & 2.96 & 2.07 & 1.73 & 1.93 \\
\midrule
\rowcolor{hilite}
Ours & \textbf{14.70} & \textbf{26.58} & \textbf{35.18} &
\textbf{14.70} & \textbf{12.09} & \textbf{10.70} & \textbf{13.29} \\
\bottomrule
\end{tabular}}
\end{minipage}
\vspace{-5pt}
\end{table}

\section{Experiments}
\label{sec:experiment}

\subsection{Experiments Setup}

\noindent\textbf{Adaptation of Existing Methods.}  
We evaluate both CIR and generation-based paradigms under the DFHR setting. CIR methods \cite{pic2word, ZS-CIR} are re-trained on our dataset using the same backbone configuration as our model to ensure a fair comparison. Generation-based methods \cite{hairclipv2, stablehair, hairfast, instantid, han2024face, withanyone, consistentid} are adapted for retrieval following a consistent protocol. Detailed adaptation procedures for each method are provided in Appendix~\ref{supp:baselines}.


\noindent\textbf{Benchmark Configuration.}
We train all models on a private training split and evaluate them on the official \textit{DFHR-Bench} benchmark, which contains strictly unseen identities to ensure fair generalization.
Three evaluation protocols are defined:
(1) \textit{Image–Image}, using a visual hairstyle reference;
(2) \textit{Image–Text}, using a textual hairstyle description; and
(3) the \textit{Bimodal Alignment Subset}, designed to assess cross-modal semantic consistency between visual and textual references.

\noindent\textbf{Metrics}. We report standard retrieval metrics: Recall@$k$ (R@$k$), Precision@$k$ (P@$k$), and mean Average Precision (mAP) with results shown at $k\!\in\!\{1,3,5\}$ across all evaluation protocols.

\noindent\textbf{Implementation Details.} Detailed in Appendix~\ref{supp:mfhc_implement}.

\subsection{Experimental Results}
\label{subsec:experiment_results}

Although both visual and textual hairstyle references describe the same target intent, the visual reference provides a more precise and discriminative cue, as shown in the Bimodal Alignment Subset (Table~\ref{tab:dual_retrieval}). We further evaluate all methods on the official Image–Image (Table~\ref{tab:dual_retrieval_imgimg}) and Image–Text (Table~\ref{tab:dual_retrieval_imgtxt}) sets. Both \textit{Face-only} and \textit{Hair-only} variants yield low performance, confirming that single-factor cues are insufficient for DFHR. 
These results underscore DFHR's non-trivial nature: even advanced CIR and image-generation methods struggle to bridge the localized feature disentanglement required for identity–hairstyle composition. CIR-based methods \cite{ZS-CIR, pic2word} perform poorly despite retraining with the same identity encoder, as their objectives align composed representations in CLIP’s holistic image space where identity and appearance often remain entangled. This highlights the need for a more robust embedding space to balance visual and textual guidance while enabling effective disentanglement. Similarly, the high complexity and diversity of hairstyle information pose challenges for existing generation-based methods such as hairstyle-transfer \cite{stablehair, hairclipv2, hairfast} and ID-consistent generation models \cite{han2024face, instantid, withanyone, consistentid}, which often struggle to fully maintain joint face-hair fidelity. MFHC leverages correlated multi-view signals to robustly align hairstyle semantics while maintaining identity consistency. See Appendix~\ref{supp:extended_exp} for further analysis and visual examples.



\subsection{Ablation Study}
\label{sec:ablation}

\begin{wrapfigure}{r}{0.55\textwidth} 
  \centering
  \vspace{-15pt} 
  
  \begin{subfigure}[b]{0.48\linewidth} 
    \centering
    \includegraphics[width=\linewidth]{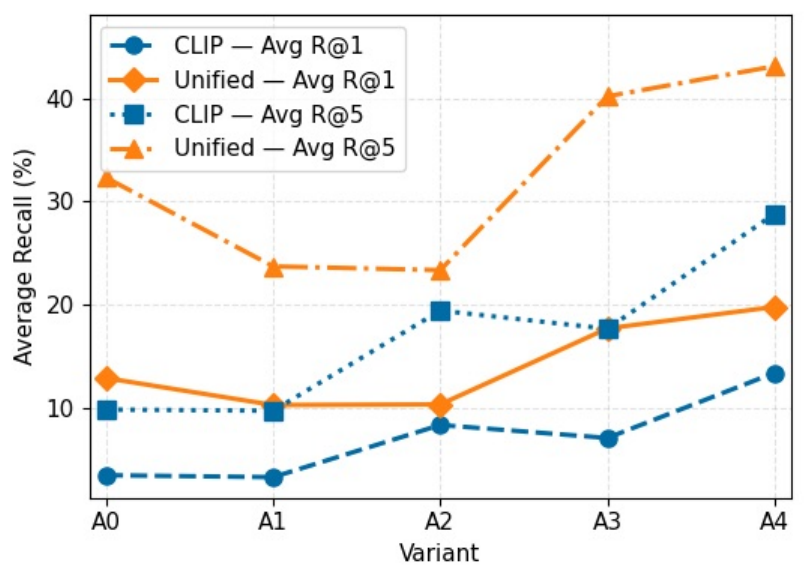}
    \vspace{-15pt} 
    \caption{}
    \label{fig:abl_variant}
  \end{subfigure}
  \hfill 
  \begin{subfigure}[b]{0.48\linewidth} 
    \centering
    \includegraphics[width=\linewidth]{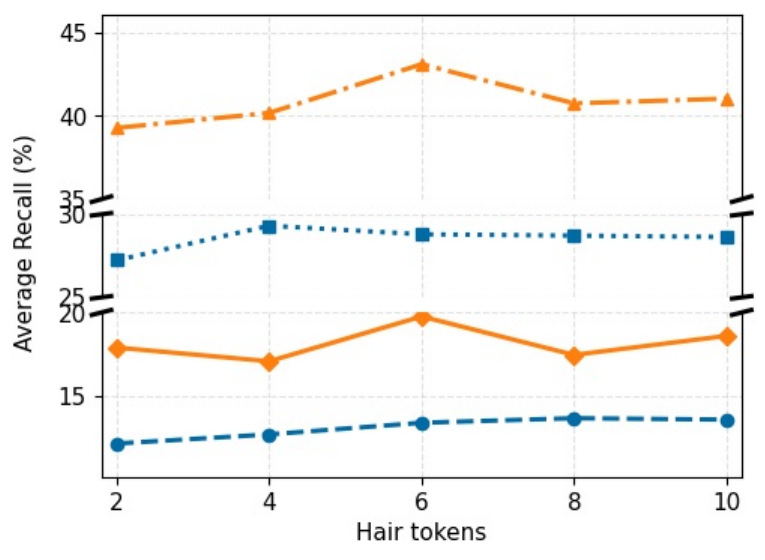}
    \vspace{-15pt} 
    \caption{}
    \label{fig:abl_token}
  \end{subfigure}
  
  \vspace{-5pt} 
  \caption{Ablation analysis: (a) variants A0–A4; (b) hair token-span sensitivity.}
  \label{fig:ablation_ab}
  \vspace{-15pt} 
\end{wrapfigure}

\noindent\textbf{Component Analysis and Capacity–Regularization Synergy}
We conduct ablations on the \textit{Bimodal Alignment Subset} to assess the contribution of each component in MFHC for both visual and textual hairstyle references. We report Avg R@1 / Avg R@5, defined as the mean of Image–Image and Image–Text scores  across the two embedding spaces (Figure~\ref{fig:abl_variant}, Table~\ref{tab:ablation}). The progression reveals a distinct \textit{capacity–regularization dynamic}.
\textit{single-view vs. multi-view contrastive (A0–A1)} shows limited gain, indicating that simply adding more alignment views without adaptation offers marginal benefit.
\textit{A2} introduces \textit{LoRA-based text adaptation and multi-token spans}, expanding representational capacity and yielding moderate improvement, though the unified space benefits less than CLIP as the added capacity amplifies identity–hairstyle entanglement (Figure~\ref{fig:abl_variant}).
\textit{Semantic Consistency (A3)} acts as a turning point: span-masked and dual-reference regularization substantially enhance factor disentanglement, producing the largest single-step gain in the unified space and elevating cross-modality alignment. Finally, hard negative composition further strengthens both CLIP and unified spaces by explicitly enhancing hairstyle discrimination and sharpening identity boundaries. Together, MFHC improves most when increased capacity is paired with disentangling regularizers, revealing a clear capacity–regularization synergy

\begin{table}[t]
\centering
\caption{Ablation on MFHC. Each row cumulatively adds components.}
\label{tab:ablation}
\resizebox{\textwidth}{!}{%
\begin{tabular}{@{} l  c  c  c  c  c  c | c  c  c | c  c  c @{}}
\toprule
\multirow{2}{*}{Variant (descriptive)} &
\multicolumn{5}{c}{Components Enabled} & \multirow{2}{*}{Gallery} &
\multicolumn{3}{c|}{Image -- Image (\%)} &
\multicolumn{3}{c}{Image -- Text (\%)} \\
\cmidrule(lr){2-6} \cmidrule(lr){8-10} \cmidrule(lr){11-13}
& Multi-View Align & LoRA & Multi-Token & $\mathcal{L}_{\text{sem}}$ & $\mathcal{L}_{\text{hardneg}}$
&  & R@5 & P@5 & mAP & R@5 & P@5 & mAP \\
\midrule
\textbf{A0}: Single-view contrastive    & \xmark & \xmark & \xmark & \xmark & \xmark & Image &  14.03 &  3.77 &  4.09 &  5.55 &  1.41 &  1.50 \\
\textbf{A1}: A0 + Multi-view alignment ($\mathcal{L}_{\text{align}}$)   & \cmark & \xmark & \xmark & \xmark & \xmark & Image &  15.84 &  3.89 &  5.27 &  3.52 &  0.80 &  0.89 \\
\textbf{A2}: A1 + LoRA text adaptation \& multi-token  & \cmark & \cmark & \cmark & \xmark & \xmark & Image &  24.35 &  6.67 &  9.16 &  14.46 &  4.07 &  4.37 \\
\textbf{A3}: A2 + Semantic Consistency Loss ($\mathcal{L}_{\text{sem}}$)     & \cmark & \cmark & \cmark & \cmark & \xmark & Image & 21.43 & 5.97  & 8.73  &  13.84 & 3.63 & 4.48 \\
\textbf{A4}: A3 + Hard Negative Mining ($\mathcal{L}_{\text{hardneg}}$)      & \cmark & \cmark & \cmark & \cmark & \cmark & Image & 35.62  & 10.15 & 14.35 & 21.97 & 6.60  & 8.07 \\
\midrule
\rowcolor{hilite}
Full (unified text-space indexing)  & \cmark & \cmark & \cmark & \cmark & \cmark & Unified  &  \textbf{50.99} &  \textbf{15.29} &  \textbf{22.63} &  \textbf{35.26} &  \textbf{10.70} &  \textbf{13.31} \\
\bottomrule
\end{tabular}%
}
\vspace{-5pt}
\end{table}

\begin{table}[t]
  \centering
  \caption{\textbf{Encoder Capacity Ablation.} The impact of scaling down the hair encoder and CLIP version on overall retrieval accuracy, using a frozen identity backbone.}
  \label{tab:enc_ablation}
  \vspace{-5pt} 
  \resizebox{0.75\linewidth}{!}{ 
    \setlength{\tabcolsep}{3pt}
    \renewcommand{\arraystretch}{0.75}
    \begin{tabular}{cc|ccc|ccc|cc}
    \toprule
    \multicolumn{2}{c|}{\textbf{Encoders}} 
    & \multicolumn{3}{c|}{\textbf{Img + Img}} 
    & \multicolumn{3}{c|}{\textbf{Img + Txt}} 
    & \multicolumn{2}{c}{\textbf{Average}} \\
    \textbf{Hair} & \textbf{CLIP Version}
    & R@1 & R@5 & mAP 
    & R@1 & R@5 & mAP 
    & @1 & $\Delta$ \\
    \midrule
    ViT-B & ViT-B & 24.82 & 50.99 & 22.63 & 14.70 & 35.26 & 13.31 & 19.76 & -- \\
    ViT-B & RN50  & 17.10 & 45.71 & 16.97 & 12.43 & 36.04 & 12.22 & 14.77 & \textcolor{red}{-4.99} \\
    ViT-S & ViT-B & 23.80 & 49.80 & 21.00 & 11.81 & 30.57 & 10.78 & 17.81 & \textcolor{red}{-1.95} \\
    ViT-S & RN50  & 17.10 & 46.89 & 17.81 & 13.14 & 32.99 & 11.81 & 15.12 & \textcolor{red}{-4.64} \\
    RN18  & ViT-B & 21.28 & 51.06 & 20.37 & 10.63 & 30.81 & 10.06 & 15.96 & \textcolor{red}{-3.80} \\
    RN18  & RN50  & 17.73 & 47.83 & 18.38 & 14.00 & 38.00 & 13.95 & 15.87 & \textcolor{red}{-3.89} \\
    \bottomrule
    \end{tabular}
  }
  \vspace{-10pt} 
\end{table}

\begin{figure}[t]
  \centering
  \begin{subfigure}[t]{0.49\linewidth}
    \centering
    \includegraphics[width=\linewidth]{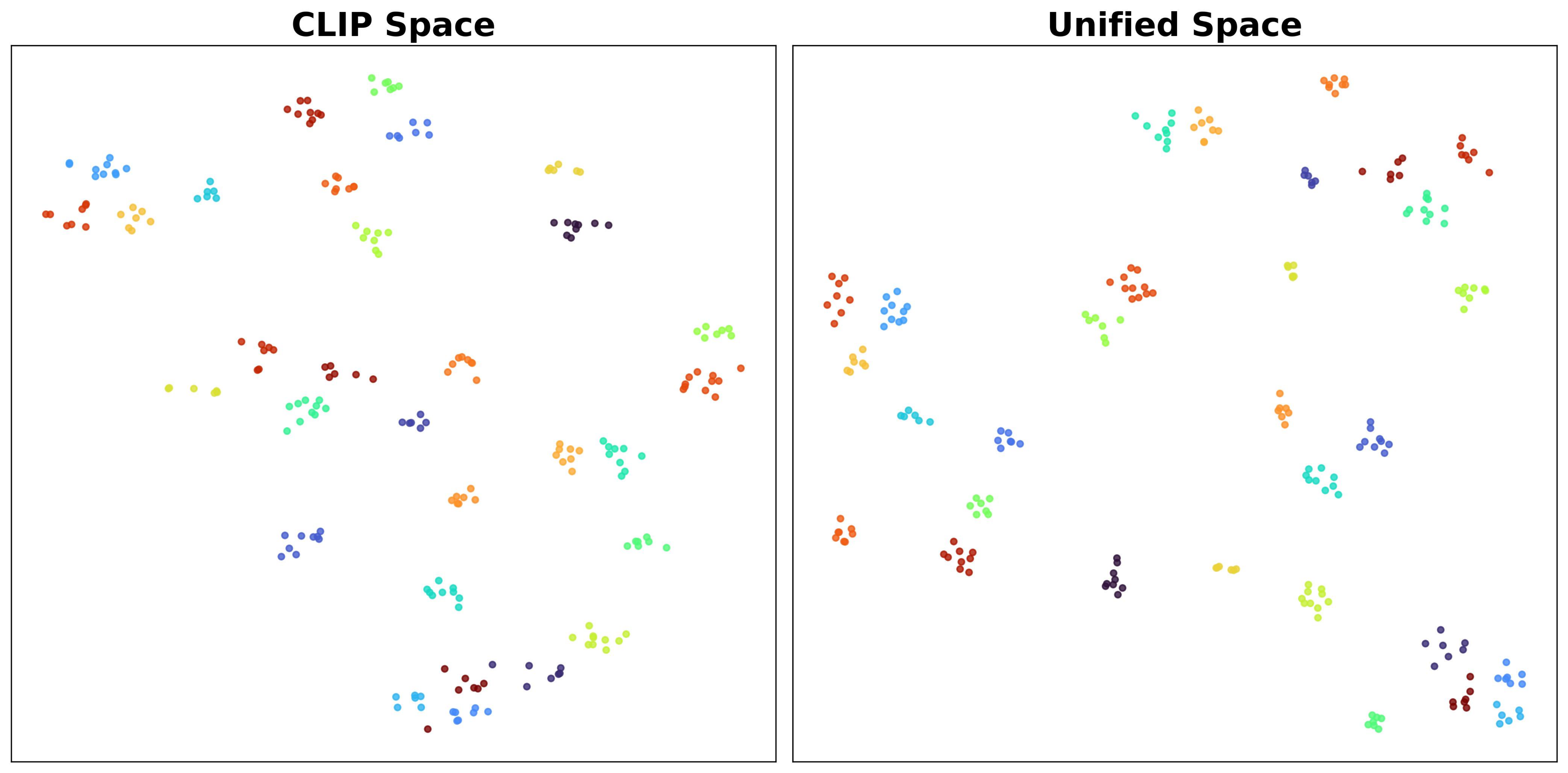}
    \caption{}
    \label{fig:abl_third}
  \end{subfigure}\hfill
  \begin{subfigure}[t]{0.49\linewidth}
    \centering
    \includegraphics[width=\linewidth]{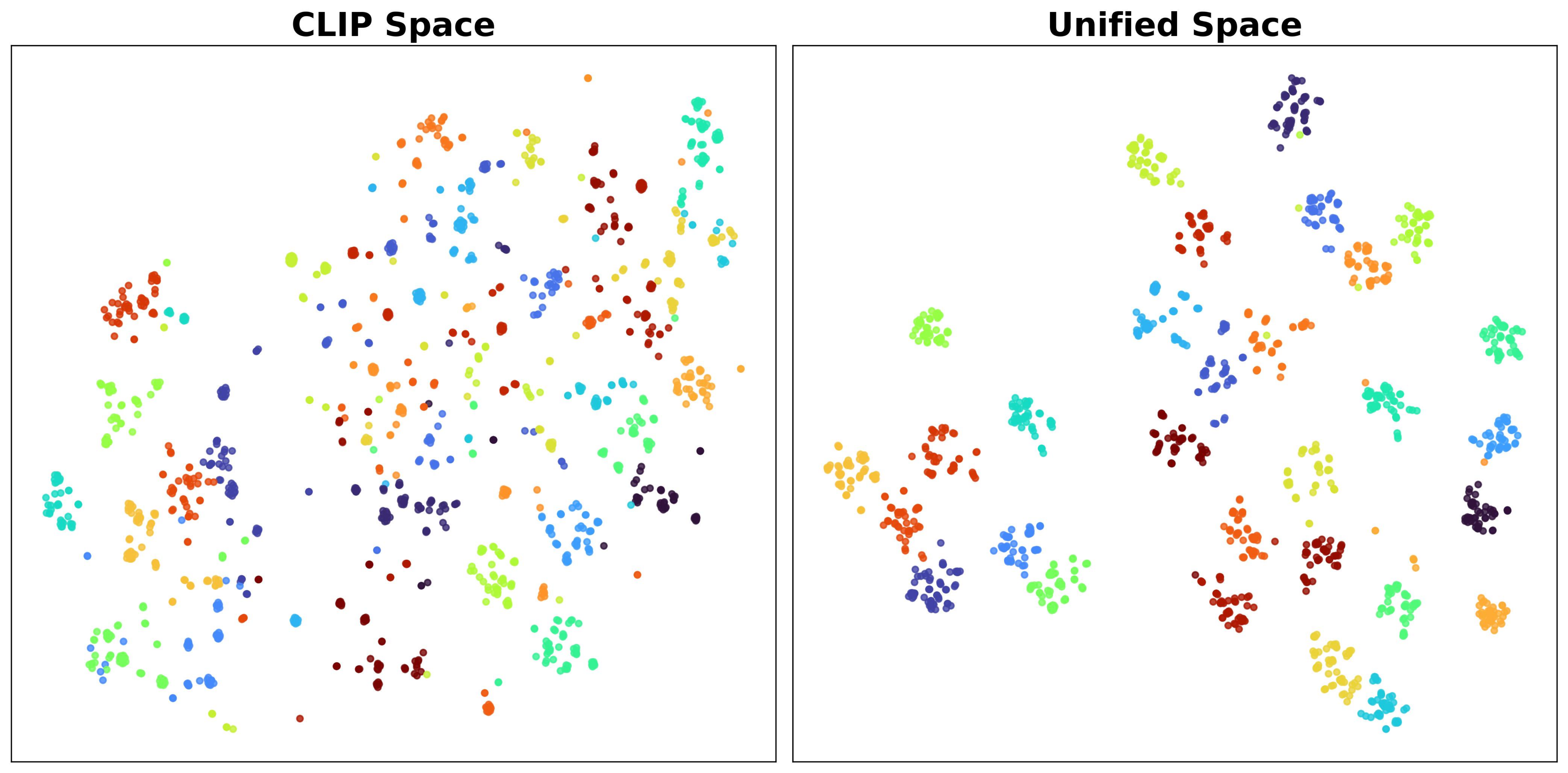}
    \caption{}
    \label{fig:abl_fourth}
  \end{subfigure}
  \caption{
  Embedding visualization:
  (a) \textit{No hairstyle variation}. Both CLIP and Unified spaces form tight, well-separated identity clusters.
  (b) \textit{With hairstyle variation}. CLIP embeddings become more dispersed and entangled, while the Unified space maintains compact, well-separated clusters.
  }
  \label{fig:ablation_cd}
  \vspace{-10pt}
\end{figure}


\noindent\textbf{Effect of Token Span Length}
We study the effect of hair-token length on retrieval (Figure~\ref{fig:abl_token}). Performance increases with span size and peaks at six tokens before slightly declining, indicating an effective granularity for hairstyle semantics. Short spans underfit fine details, whereas longer spans introduce redundancy and semantic drift in the text encoder. The unified space benefits more than CLIP, highlighting the role of token-level compositionality in cross-modal alignment. A similar sensitivity is observed for identity tokens, where a span of six yields the most robust performance in both Image--Image and Image--Text settings (see Appendix~\ref{supp:extended_exp}).

\noindent\textbf{Toward identity-consistent image retrieval:}
DFHR requires strict preservation of the subject’s identity. We therefore evaluate MFHC under a face-recognition setting. For each query, we check whether the retrieved images correspond to the same identity as the anchor and report Recall@5. MFHC achieves \textbf{92.5 R@5}, confirming that the unified representation preserves identity fidelity while composing identity and hairstyle.

\noindent\textbf{Identity Clustering Under Hairstyle Variation.}
Figure~\ref{fig:ablation_cd} shows t-SNE visualizations of identity embeddings for 30 randomly selected identities in CLIP (left) and our Unified space (right). Without hairstyle variation (Figure~\ref{fig:abl_third}), both spaces form tight, well-separated clusters, confirming that CLIP naturally preserves identity when appearance is uniform. When multiple hairstyles are introduced (Figure~\ref{fig:abl_fourth}), CLIP clusters degrade: embeddings drift across hairstyles, and similar-haired identities sometimes overlap, revealing strong entanglement between identity and hair. In contrast, the Unified space maintains compact, distinct clusters, effectively disentangling identity from hairstyle and ensuring identity-consistent representations under appearance changes.

\noindent\textbf{Encoder Robustness.}
We evaluate MFHC across different encoder capacities to assess its sensitivity to architectural choices. As shown in Table~\ref{tab:enc_ablation}, we fix the identity encoder and vary both the hair encoder and the CLIP backbone to analyze how performance changes under different capacity settings. Despite reducing model capacity, MFHC exhibits consistent performance trends with only moderate degradation in retrieval accuracy. These results suggest that the framework is not overly dependent on a specific backbone configuration. We attribute this stability to using the CLIP text embedding space for composition, which provides semantic regularization for mixed image–text conditioning.

\section{Conclusion}
\label{sec:conclusion}

In this work, we introduce Dual Face–Hair Retrieval, a mixed-modality retrieval task that studies the composition of facial identity and hairstyle across image and text inputs. To support this problem, we present DFHR-Bench, a benchmark with carefully designed annotations for evaluating identity-aware hairstyle retrieval. We further evaluate representative state-of-the-art methods from related fields and find that existing approaches struggle to address this setting. We also propose the MFHC, which composes disentangled identity and hairstyle representations through token injection and multi-view supervision. We believe this work provides a concrete foundation and a reliable testbed for future research on identity-preserving, attribute-controlled retrieval.

\bibliographystyle{splncs04}
\bibliography{main}

\clearpage
\begin{center}
{\LARGE\bfseries Supplementary Material}\\[0.5em]
{\Large Mixed-Modality Dual Face--Hair Retrieval}
\end{center}
\section*{Table Of Content}
\appendix
\setcounter{page}{1}

{\color{blue}

\noindent A~~\hyperref[sec:ethics]{Ethics Statement} \dotfill \pageref{sec:ethics}

\noindent B~~\hyperref[sec:scope]{DFHR Scope} \dotfill \pageref{sec:scope}

\noindent C~~\hyperref[benchmark-quality]{Benchmark Qualities Clarification} \dotfill \pageref{benchmark-quality}

\noindent D~~\hyperref[sec:extended-version]{Official Set and Extended Set} \dotfill \pageref{sec:extended-version}

\noindent E~~\hyperref[supp:details_guideline]{Detailed Annotation Guidelines} \dotfill \pageref{supp:details_guideline}

\hspace{1em} E.1~~\hyperref[sec:attribute-clarification]{Pilot Study on Hairstyle Semantics} \dotfill \pageref{sec:attribute-clarification}

\hspace{1em} E.2~~\hyperref[sec:shared-guidelines]{Annotation Guidelines} \dotfill \pageref{sec:shared-guidelines}

\hspace{1em} E.3~~\hyperref[sec:annotation_platform]{Annotation Platforms} \dotfill \pageref{sec:annotation_platform}

\noindent F~~\hyperref[sec:benchmark-construction]{Benchmark Dataset Construction} \dotfill \pageref{sec:benchmark-construction}

\hspace{1em} F.1~~\hyperref[sec:target-selection]{Ground Truths Selection} \dotfill \pageref{sec:target-selection}

\hspace{1em} F.2~~\hyperref[sec:hair-encoder-implementation]{Hair Encoder Implementation} \dotfill \pageref{sec:hair-encoder-implementation}

\hspace{1em} F.3~~\hyperref[sec:image-pipeline]{Image Annotation Pipeline} \dotfill \pageref{sec:image-pipeline}

\hspace{1em} F.4~~\hyperref[sec:text-pipeline]{Text Annotation Pipeline} \dotfill \pageref{sec:text-pipeline}

\hspace{1em} F.5~~\hyperref[sec:synthesis-algorithm]{Image Synthesis Algorithms} \dotfill \pageref{sec:synthesis-algorithm}

\noindent G~~\hyperref[supp:addiotnal_dataset_analysis]{Additional Dataset Analysis} \dotfill \pageref{supp:addiotnal_dataset_analysis}

\hspace{1em} G.1~~\hyperref[supp:bench_extended]{Inter-Annotator Agreement} \dotfill \pageref{supp:bench_extended}

\hspace{1em} G.2~~\hyperref[sec:qualitative_examples]{Qualitative Examples} \dotfill \pageref{sec:qualitative_examples}

\hspace{1em} G.3~~\hyperref[cross-dataset-analysis]{Benchmark Comparison and Positioning} \dotfill \pageref{cross-dataset-analysis}

\noindent H~~\hyperref[supp:mfhc_implement]{Implementation Details} \dotfill \pageref{supp:mfhc_implement}

\noindent I~~\hyperref[supp:method]{Method Details} \dotfill \pageref{supp:method}

\noindent J~~\hyperref[supp:extended_exp]{Additional Analysis} \dotfill \pageref{supp:extended_exp}

\hspace{1em} J.1~~\hyperref[supp:qualitative-comparison]{Qualitative Comparison} \dotfill \pageref{supp:qualitative-comparison}

\noindent K~~\hyperref[supp:baselines]{Adaptation of Existing Methods} \dotfill \pageref{supp:baselines}

\hspace{1em} K.1~~\hyperref[sec:fusion-baseline]{Fusion-Based Methods} \dotfill \pageref{sec:fusion-baseline}

\hspace{1em} K.2~~\hyperref[sec:generation-baseline]{Generation-Based Baselines} \dotfill \pageref{sec:generation-baseline}

\hspace{1em} K.3~~\hyperref[sec:composed-baseline]{Composed Image Retrieval Baselines} \dotfill \pageref{sec:composed-baseline}

\noindent L~~\hyperref[sec:limitation]{Limitations} \dotfill \pageref{sec:limitation}
\vspace{15pt}
}

We release all training resources, including scripts, checkpoints, and inference code.
The accompanying annotation platform includes dedicated modules for image annotation, text annotation, and validation. Publicly Available at \texttt{github/xyz} (repository anonymized for double-blind review).
\section{Ethics Statement}
\label{sec:ethics}
\textbf{Our data collection procedures have been approved by XYZ Institutional Review Board (IRB) for ethical compliance} (anonymized for double-blind review).
All datasets used in this work were obtained solely from official repositories or publicly available web sources to ensure data integrity, authenticity, and respect for the original intent of the data creators.
We have carefully reviewed and adhered to the license agreements of all included datasets and models to guarantee lawful secondary use for academic research.
All images were manually screened to exclude any offensive, sexual, violent, or other sensitive content.
Annotations were conducted under institutional authorization, following fair compensation and ethical employment practices.
No personally identifiable or sensitive personal metadata is released.
Synthetic images used for hairstyle augmentation are explicitly labeled as synthetic and contain no deceptive or harmful material.
The dataset and all accompanying resources are released under the \textit{Creative Commons Attribution–NonCommercial 4.0 (CC BY-NC 4.0)} license for non-commercial academic research.
\section{DFHR Scope}
\label{sec:scope}

DFHR is designed as a \emph{stress-test} for studying \emph{dual-constraint reasoning} in identity-aware retrieval systems, targeting a failure regime that arises in real-world settings but is not isolated by existing benchmarks.
Rather than introducing multiple attributes, DFHR focuses on a single attribute to expose this difficulty in a controlled manner, as existing models already struggle under single-attribute identity-preserving conditioning.
Within this setting, attribute reference choice is critical.

Analysis of CelebA\cite{celeba} and MAAD-Face\cite{maad-face} (Table \ref{tab:attribute_analysis}) shows that many commonly suggested facial attributes (e.g., facial hair, makeup, accessories) appear in fewer than \texttt{50\%} of images and exhibit strong demographic skew: facial hair is predominantly male (\texttt{$\sim$40\%} male vs.\ \texttt{$<$0.1\%} female), while makeup is predominantly female (\texttt{$\sim$65-72\%} female vs.\ \texttt{$<$0.3\%} male).
Such skew undermines their suitability as category-agnostic diagnostic constraints.
Head hairstyle is therefore selected for its broad coverage (\texttt{$>$80\%}), comparatively balanced gender representation, and strong identity entanglement with meaningful intra-identity variation.
The face--hair formulation is not intended to fix the task boundary to hairstyle, but to establish a reproducible testbed for a broader goal in \emph{identity-aware facial analysis}: enforcing identity consistency under attribute conditioning, extendable to other localized facial attributes.

\begin{table}[t]
\centering
\caption{\textbf{Face Attribute Statistics and Suitability Analysis.}
Gender Balance indicates whether the male–female proportion difference is below a predefined threshold.
High Proportion indicates whether the attribute covers a sufficient portion of the dataset.}
\resizebox{\linewidth}{!}{
\begin{tabular}{l|l|ccc|cc}
\toprule
\textbf{Dataset} & \textbf{Attribute} 
& \textbf{P(\text{Attr}|\text{Male}) (\%) }
& \textbf{P(\text{Attr}|\text{Female}) (\%)} 
& \textbf{Total (\%)} 
& \textbf{Gender Balance} 
& \textbf{High Prop.} \\
\midrule

\multirow{4}{*}{CelebA \cite{celeba}}
 & Beard / Facial Hair & 39.4 & 0.12 & 16.5 & \textcolor{red}{\xmark} & \textcolor{red}{\xmark} \\
 & Makeup              & 0.3 & 64.8 & 38.7 & \textcolor{red}{\xmark} & \textcolor{red}{\xmark} \\
 & Accessory           & 35.1 & 46.0 & 41.4 & \textcolor{ForestGreen}{\cmark} & \textcolor{red}{\xmark} \\
 
 & \cellcolor{hilite} \textbf{Head Hair} 
 & \cellcolor{hilite} \textbf{72.7} 
 & \cellcolor{hilite} \textbf{89.0} 
 & \cellcolor{hilite} \textbf{82.2} 
 & \cellcolor{hilite} \textcolor{ForestGreen}{\cmark} 
 & \cellcolor{hilite} \textcolor{ForestGreen}{\cmark} \\
\midrule

\multirow{4}{*}{MAAD-Face \cite{maad-face}}
 & Beard / Facial Hair & 38.0 & 0.02 & 18.1 & \textcolor{red}{\xmark} & \textcolor{red}{\xmark} \\
 & Makeup              & 0.01 & 72.8 & 29.7 & \textcolor{red}{\xmark} & \textcolor{red}{\xmark}  \\
 & Accessory           & 34.8 & 77.9 & 52.4 & \textcolor{red}{\xmark} & \textcolor{red}{\xmark}  \\
 
 & \cellcolor{hilite} \textbf{Head Hair} 
 & \cellcolor{hilite} \textbf{76.7} 
 & \cellcolor{hilite} \textbf{91.0} 
 & \cellcolor{hilite} \textbf{82.5} 
 & \cellcolor{hilite} \textcolor{ForestGreen}{\cmark} 
 & \cellcolor{hilite} \textcolor{ForestGreen}{\cmark} \\

\bottomrule
\end{tabular}}
\label{tab:attribute_analysis}
\end{table}
\section{Benchmark Qualities Clarification}
\label{benchmark-quality}
This section outlines the key principles guiding the construction of our benchmark, ensuring it is conceptually coherent, diverse, precisely annotated, and multi-modal.

\noindent\textbf{Conceptual Alignment.}
Conceptual alignment refers to the extent to which annotators share a consistent and well-defined understanding of hairstyle concepts within the benchmark. Achieving this alignment is essential for maintaining annotation integrity, as perceptual subjectivity can introduce inconsistencies across annotators \cite{subjectivity, disagreementsurvey, subjectivity1}. In practice, hairstyle perception varies along multiple dimensions—shape, texture, color, boundary definition, and fine-grained styling attributes—making it challenging to ensure coherent interpretation across individuals.
We conducted a pilot survey designed to identify and quantify the key criteria underlying hairstyle similarity judgments. Responses from this survey were analyzed to derive a set of core annotation principles, which serve as a unifying conceptual framework for subsequent labeling. Based on this analysis, we distilled representative hairstyle attributes and categorized them into three groups according to their relative importance in similarity evaluation. These guidelines were formalized into the annotation protocol, enabling annotators to calibrate their judgments and promoting consistent, conceptually aligned labeling across the benchmark.

\noindent\textbf{Diversity.} 
As DFHR targets fine-grained compositional retrieval, it requires diversity manifested through rich intra-identity and intra-hairstyle variation across poses, contexts, viewpoints, and reference modalities.
Achieving such diversity is challenging. The retrieval gallery must be sufficiently large and varied to include numerous near-duplicates and plausible distractors; however, real-world data scarcity limits the number of available images of the same individual across multiple hairstyles. Privacy restrictions and the limited hairstyle variation present in standard face datasets further amplify this difficulty.
To mitigate these constraints, we leverage the inherent diversity of CelebA-HQ~\cite{karras2018_celeba_hq} and FFHQ~\cite{karras2019_ffhq} as the foundation of our image pool. We manually curate approximately 500 distinct identities to ensure breadth across demographic and appearance attributes. Because existing datasets rarely provide multiple hairstyle instances per person, we expand the pool by collecting additional publicly available images from the Internet and validating them with facial verification. Finally, we incorporate controlled hairstyle synthesis via deepfake-based augmentation to enrich intra-identity hairstyle diversity while preserving photorealism and identity fidelity.

\noindent\textbf{High Precision.} Building on conceptual alignment, the benchmark requires meticulous annotations to achieve reliable ground truths. Annotation quality directly determines the benchmark’s value for evaluation and comparison. However, this quality can be degraded by various sources of bias, including fatigue, gender differences, task complexity, demographic stereotypes, recency, and the halo effect~\cite{bias_mitigation, research_bias, bias_llm, cui2025biasinbiasout}. To mitigate these factors, we scale the number of annotators relative to labeling time to reduce individual workload and cognitive fatigue, and aggregate their labels using \textbf{majority voting} to ensure precision and robustness against outliers. Regular annotation breaks are introduced to maintain focus and consistency over time. We also monitor gender distribution to counteract stylistic bias frequently observed in appearance-related perception tasks. Additionally, we raise annotator awareness of common cognitive biases such as recency and the halo effect through oral briefings and explicit annotation guidelines.

\noindent\textbf{Multi-Modality.} To fully capture dual-condition query format $\langle q_\text{id}, q_\text{hair} \rangle$, the dataset should incorporate multi-modal components for flexible retrieval scenarios. These include image-based and text-based representations for hairstyle. This multi-modality ensures the benchmark reflects the comprehensive, real-world nature of Dual Face-Hair Retrieval, enabling evaluations across varied input combinations.

These criteria collectively position our benchmark as a comprehensive tool for advancing research in Dual Face-Hair Retrieval, facilitating fair comparisons and driving innovations in multi-modal, fine-grained image retrieval.
\section{Official Set and Extended Set}
\label{sec:extended-version}
To support different use cases, DFHR-Bench provides two scales of triplets: an official set and a substantially larger extended set. The official split is curated under stricter conditions, including filtering identity-reference images that closely resemble the ground-truth targets. This prevents models from exploiting trivial face-similarity shortcuts, ensuring that retrieval performance truly reflects the identity–hairstyle composition rather than simple visual matching. In contrast, the extended set preserves the whole pool of valid triplets, offering a much larger collection (up to 180K) that is well suited for model fine-tuning, ablation, or further methodological development. Because the extended set contains triplets that overlap with the official split, it should not be used for training when the official set is used for evaluation, to avoid data leakage and overfitting.

\section{Detailed Annotation Guidelines}
\label{supp:details_guideline}
Our methodology to construct the dataset is designed to minimize potential sources of disagreement among annotators. Four common sources of disagreement are identified: subject ambiguity, subjectivity, unclear annotation schemes, and annotator errors \cite{disagreementsurvey}. To reduce subject ambiguity, we first filter out low-quality or ambiguous images. To mitigate subjectivity, we conduct pilot annotations to analyze different perspectives and align annotators toward a shared understanding. Clear annotation guidelines are then established to provide a consistent scheme and reduce interpretation gaps. 
\subsection{Pilot Study on Hairstyle Semantics}
\label{sec:attribute-clarification}
We conducted a pilot study to elicit human notions of hairstyle similarity. The format of our pilot survey is listed as follows:

\begin{mdframed}[linewidth=0.5pt, roundcorner=8pt, backgroundcolor=gray!3]
\vspace{6pt}
\noindent\textbf{Questionnaire — Page \textit{i}} (\textit{i}=1\,..\,30)

\noindent\textbf{Anchor:} \texttt{ref\_\{i\}.jpg}

\noindent\textbf{Gallery:} 50 images (select all perceived as similar)

\vspace{6pt}\noindent\textbf{List of Supporting Attributes:}\\
\textit{Write key hairstyle attributes that make the selected images appear similar (e.g., color, length, texture, bangs, volume).}

\vspace{6pt}
\noindent\textbf{Sampled Candidates (up to 3):}
\begin{enumerate}
  \item \texttt{cand\_\{i\}\_s1.jpg} \\
        Similarity (6–10): \rule{2.2cm}{0.15mm} \\
        Key differences: \rule{3.5cm}{0.15mm}
  \item \texttt{cand\_\{i\}\_s2.jpg} \\
        Similarity (6–10): \rule{2.2cm}{0.15mm} \\
        Key differences: \rule{3.5cm}{0.15mm}
  \item \texttt{cand\_\{i\}\_s3.jpg} \\
        Similarity (6–10): \rule{2.2cm}{0.15mm} \\
        Key differences: \rule{3.5cm}{0.15mm}
\end{enumerate}

\vspace{4pt}
\noindent\textbf{Final Page — Attribute Ranking.}\\
List and rank hairstyle attributes by importance when judging similarity.

\vspace{6pt}
\label{fig:pilot_survey}
\end{mdframed}

Each participant completed 30 queries. For each query, the annotator was shown a single \textit{anchor} (reference) image along with a gallery of 50 candidate images. Their primary task was to select all candidates they judged to exhibit a similar hairstyle to the anchor. From these selections, they then chose up to three examples to justify in greater detail, specifying the shared hairstyle attributes that supported their choices. For every sampled candidate, annotators additionally provided a similarity rating on a 6–10 scale\footnote{The scale is intentionally compressed toward the upper range to reduce sampling bias \cite{selection-bias} and ensure more consistent interpretation across annotators.}, accompanied by a brief description of the key visual differences relative to the anchor. After completing all 30 queries, each annotator concluded the session by ranking hairstyle attributes by importance.

\begin{algorithm}[H]
\caption{Computation of Global Attribute Importance}
\begin{algorithmic}[1]
\REQUIRE Queries $\{q\}$, supporting attribute sets $\{\mathcal{S}_q\}$, sampled candidates $\{(s_{qj},\, \mathcal{D}_{qj})\}_{j=1}^{k_q}$
\ENSURE Normalized attribute weights $\{w(a)\}_{a\in\mathcal{A}}$

\STATE Initialize $w(a) \leftarrow 0$ for all $a \in \mathcal{A}$; \quad $c \leftarrow 0$
\FOR{each query $q$}
    \FOR{$j = 1$ \TO $k_q$}
        \STATE $\tilde{s}_{qj} \leftarrow s_{qj} - 5$ \hfill // rescale to $[1,5]$
        \FOR{each $a \in \mathcal{S}_q$}
            \STATE $w(a) \leftarrow w(a) + \tilde{s}_{qj}$
        \ENDFOR
        \FOR{each $a \in \mathcal{D}_{qj}$}
            \STATE $w(a) \leftarrow w(a) + (5 - \tilde{s}_{qj})$
        \ENDFOR
        \STATE $c \leftarrow c + 1$
    \ENDFOR
\ENDFOR
\FOR{each $a \in \mathcal{A}$}
    \STATE $w(a) \leftarrow w(a) / c$ \hfill // normalize aggregated evidence
\ENDFOR
\RETURN $\{w(a)\}_{a\in\mathcal{A}}$
\end{algorithmic}
\label{algo:attribute-importance-score}
\end{algorithm}
\begin{wrapfigure}{r}{0.5\textwidth} 
\centering
\vspace{-25pt}
\resizebox{0.5\columnwidth}{!}{%
\begin{tabular}{@{} l c c c @{}}
\toprule
\textbf{Hair Attribute} & \textbf{Score} & \textbf{Freq. Count} & \textbf{Freq. Rate} \\
\midrule
\multicolumn{4}{l}{\textit{High-impact attributes}} \\
Hair Length      & 2.97 & 655 & 81.98\% \\
\rowcolor{stronghl}
Hairstyle        & 2.58 & 547 & 72.59\% \\
Hair Color       & 2.49 & 580 & 68.46\% \\
\midrule
\multicolumn{4}{l}{\textit{Mid-level attributes}} \\
\rowcolor{stronghl}
Gender           & 1.45 & 319 & 40.80\% \\
Hair Texture     & 1.39 & 326 & 39.92\% \\
Parting Style    & 1.20 & 260 & 32.54\% \\
Bangs Style      & 0.92 & 218 & 27.28\% \\
\midrule
\multicolumn{4}{l}{\textit{Low-frequency attributes}} \\
Hair Thickness   & 0.50 & 147 & 18.40\% \\
Hair Volume      & 0.42 & 129 & 16.15\% \\
Hairline Shape   & 0.23 & 69  & 8.64\% \\
\bottomrule
\end{tabular}
}
\captionof{table}{Global importance scores and occurrence statistics for annotated hairstyle attributes.
Highlighted rows denote attributes that display greater semantic variability across annotators, particularly \textit{Hairstyle} (complex structure) and \textit{Gender} (sensitive bias cue), which require additional analysis.
}
\vspace{-20pt}
\label{tab:pilot-study-ranking}
\end{wrapfigure}
\noindent\textbf{Attribute Weight Extraction.} 
To quantify the relative importance of hairstyle attributes, we process annotators' free-text responses from the pilot study. All descriptions are first parsed by an LLM to extract attribute mentions; a randomly sampled 50\% subset is then manually verified by experts, who confirmed high consistency with minimal refinement. The extracted terms are mapped into ten canonical categories 
$\mathcal{A}=$\{\textit{Hairstyle}, \textit{Hair Color}, \textit{Gender}, \textit{Hair Length}, \textit{Hair Texture}, \textit{Hair Volume}, \textit{Hair Thickness}, \textit{Parting Style}, \textit{Bangs Style}, \textit{Hairline Shape}\}.

\begin{figure}
\includegraphics[width=1\linewidth]{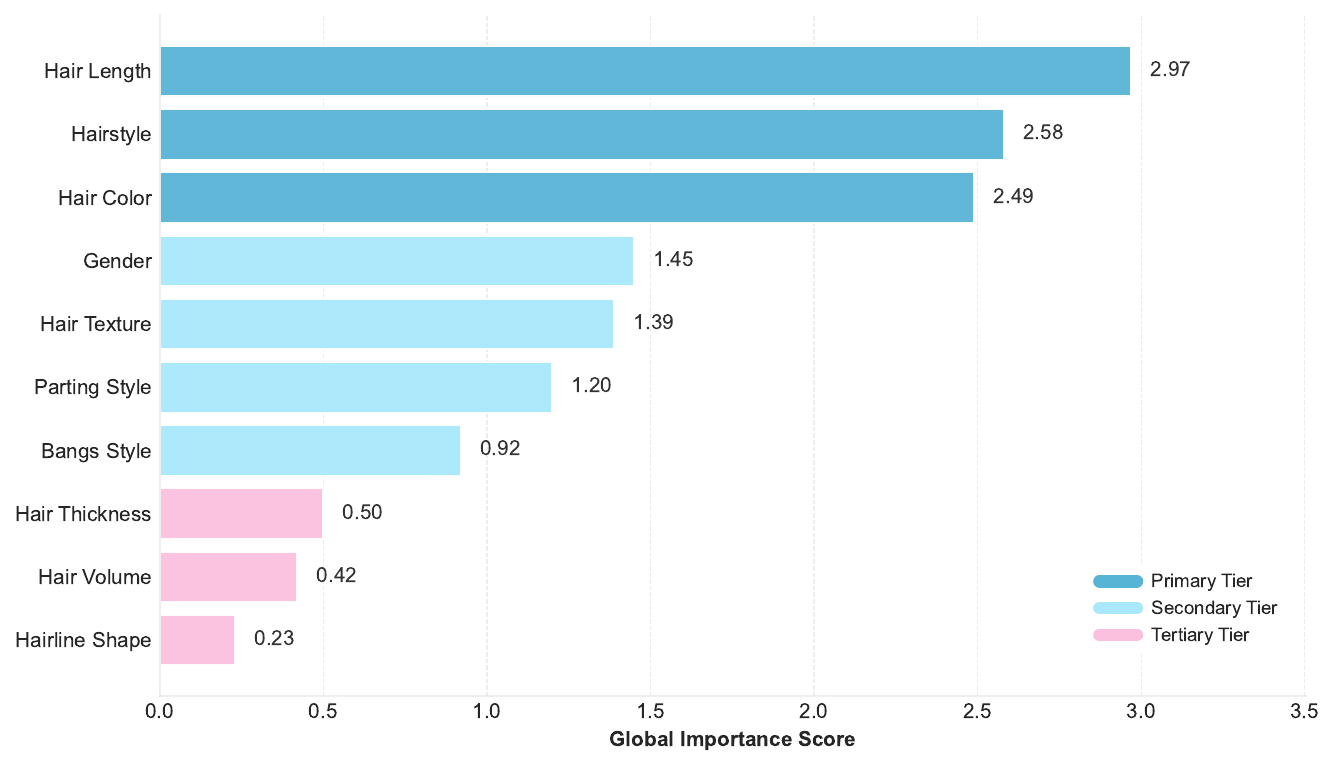}
\captionof{figure}{The bar chart illustrates the rankings of hair attributes by importance score resulting from the pilot study.}
\label{fig:pilot-study-ranking}
\end{figure}

As described in Algorithm~\ref{algo:attribute-importance-score}, for each query, annotators provide (1) a set of \emph{supporting attributes} 
$\mathcal{S}_q \subseteq \mathcal{A}$ representing shared properties among all gallery images they marked as similar, and (2) for each sampled candidate $j$, a similarity score $s_{qj}\in[6,10]$ and a set of \emph{conflicting attributes} 
$\mathcal{D}_{qj} \subseteq \mathcal{A}$ describing the key differences. 
We rescale the similarity to $\tilde{s}_{qj}=s_{qj}-5\in[1,5]$ so that higher values reward supporting attributes, while lower values emphasize conflicting ones. 
For each sampled candidate, we add $\tilde{s}_{qj}$ to the weight of all $a\in\mathcal{S}_q$ and $(5-\tilde{s}_{qj})$ to all $a\in\mathcal{D}_{qj}$, and increment a global pair counter $c$. 
After processing all responses ($c=799$), the final importance of each attribute is obtained by normalizing its accumulated weight by~$c$.
The results are described in Table \ref{tab:pilot-study-ranking} and Figure \ref{fig:pilot-study-ranking}.

\noindent\textbf{Discussion.}  
The normalized weights reveal that human hairstyle perception is dominated by three attributes \{\textit{Hair Length}, \textit{Hairstyle}, \textit{Hair Color}\} which consistently accumulate the strongest supporting and conflicting evidence across annotated pairs. Mid-level factors such as \{ \textit{Gender}, \textit{Texture}, \textit{Parting}, \textit{Bangs} \} contribute more modestly, typically refining judgments in cases where the primary cues are similar. Attributes like \{ \textit{Thickness}, \textit{Volume}, \textit{Hairline Shape} \} have minimal influence, suggesting that they are either less visually discriminative or less salient to human observers. The distribution highlights a clear hierarchy of cues governing human hairstyle similarity.

\noindent\textbf{Additional Hair Attribute Analysis.}
Among all attributes, \textit{Hairstyle} is the most abstract, as it encompasses several fine-grained and often hard-to-verbalize cues. Annotators frequently expressed these cues using vague or example-based phrases (e.g., “\texttt{combed upwards}”, “\texttt{tied at the back}”, “\texttt{Pixie cut}”, “\texttt{Slicked back}”). To obtain a more interpretable structure, we decompose \textit{Hairstyle} into three latent factors, including \textit{Geometric Shape}, \textit{Distinctive Styling}, and \textit{Combing Direction}. These sub-attributes are derived from recurring patterns in annotator reasoning and consolidated through consensus review.
Another related consideration is the occasional use of \textit{Gender} in annotator explanations. Although such references naturally arise when humans judge hairstyle similarity, gender is not a structural attribute of hair and may introduce unintended bias. To mitigate this, annotators were explicitly instructed \emph{not} to rely on gender presentation as a basis for similarity judgments. We still log these mentions to quantify how often gender implicitly influences human reasoning, but we intentionally exclude \textit{Gender} from our primary attribute set. This separation ensures that the analysis centers on intrinsic hairstyle characteristics rather than social cues, leading to a more principled and generalizable understanding of hairstyle similarity across identities and contexts.

\subsection{Annotation Guidelines}
\label{sec:shared-guidelines}

\noindent\textbf{Core Principles.}  
Guided by insights from our pilot study, we identified a set of hairstyle attributes that collectively define the structure and appearance of a hairstyle. Based on the empirical ranking of attribute importance, we group these attributes into three functional categories: 

\begin{itemize}
    \item \textbf{Main Attributes:} Attributes whose differences fundamentally alter hairstyle identity. When two hairstyles differ noticeably in any of these attributes, they are regarded as distinct. These attributes correspond to the top-ranked factors in Table~\ref{tab:pilot-study-ranking}.
    \item \textbf{Complementary Attributes:} Attributes that contribute secondary but meaningful cues to similarity. Hairstyles may still be considered similar when only a small subset of these attributes differ or when such differences are subtle. Typical examples include \textit{Hair Texture}, \textit{Parting Style}, and \textit{Bangs Style}.
    \item \textbf{Side Attributes:} Attributes of lesser importance that annotators may reference to refine their decisions, but are not decisive factors. Examples include \textit{Hair Thickness}, \textit{Hair Volume}, or \textit{Hairline Shape}.
\end{itemize}

\noindent\textbf{Attribute Flexibility.}  
Certain attributes inherently exhibit a wide range of valid variations across styles. These are flagged as \textit{flexible attributes} to remind annotators to apply a tolerant matching range during evaluation. Examples include \textit{Hair Color}, which may vary within similar tonal groups, and \textit{Bangs Style}, which can appear in diverse shapes but serve similar stylistic functions.

To promote consistent interpretation across annotators, we provide example sub-categories and corresponding visual illustrations for each attribute (Table~\ref{tab:text_attribute_properties}). These examples capture commonly observed variations but are not intended to be exhaustive.

\begin{table}[t]
\centering
\caption{Example categories for each hairstyle attribute. These lists, paired with visual references, assist annotators in interpreting attribute meanings; they may be refined or extended as needed.}
\label{tab:text_attribute_properties}
\renewcommand{\arraystretch}{1.25}
\begin{tabular}{p{0.36\linewidth} p{0.54\linewidth}}
\toprule
\textbf{Attribute} & \textbf{Sample Categories} \\
\midrule
\textit{Hair Length} & buzzed / ear-length / neck-length / shoulder-length / long \\

\textit{Geometric Shape} & even-length / layered / long-top--short-sides / mullet / asymmetric \\

\textit{Distinctive Styling} & ponytail / space buns / bun / half-up ponytail / cornrows / shaved patterns \\

\textit{Combing Direction} & downward / side-swept / center-outward (spiky) / upward / forward / backward \\

\textit{Hair Color} & black / dark brown / light brown / dyed / highlighted \\

\textit{Hair Texture} & straight / wavy / loose curls / tight curls \\

\textit{Parting Style} & no part / middle part / side part \\

\textit{Bangs Style} & no bangs / side bangs / curtain bangs / blunt bangs / see-through bangs / long bangs \\

\textit{Hair Thickness} & bald / thin / medium / thick \\

\textit{Hair Volume} & flat / medium / voluminous \\

\textit{Hairline Shape} & straight / M-shaped / rounded / receding \\

\bottomrule
\end{tabular}
\end{table}

\begin{figure}
    \centering
    \begin{subfigure}[b]{1\linewidth}
        \includegraphics[width=1\linewidth]{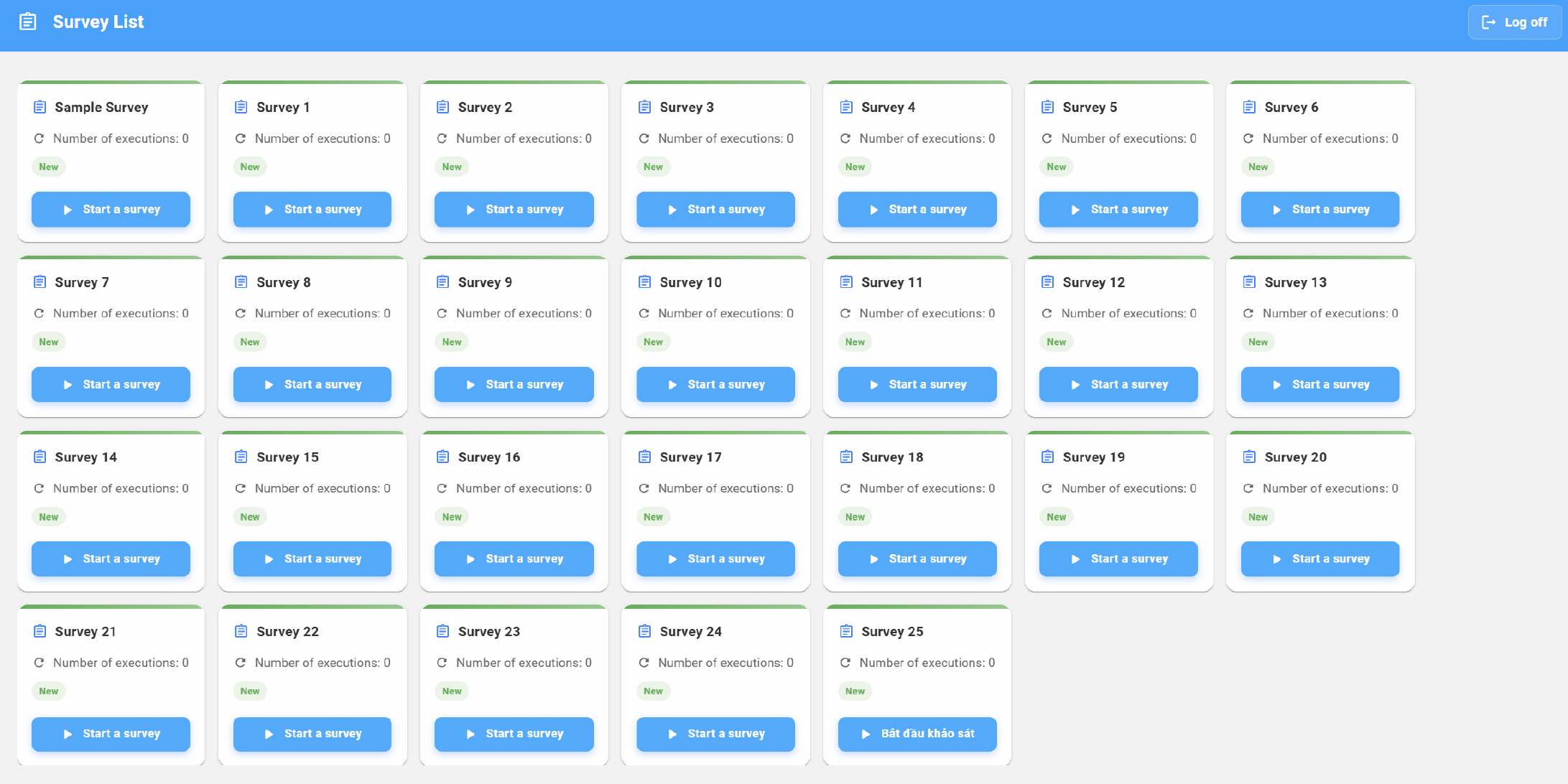}
        \caption{The list view of the annotation platform. There are 25 surveys, each of which contains 20 queries, representing 20 target images.}
        \label{fig:platform-hair-annotation-list-view}
    \end{subfigure}
    \hfill
    \begin{subfigure}[b]{1\linewidth}
        \includegraphics[width=1\linewidth]{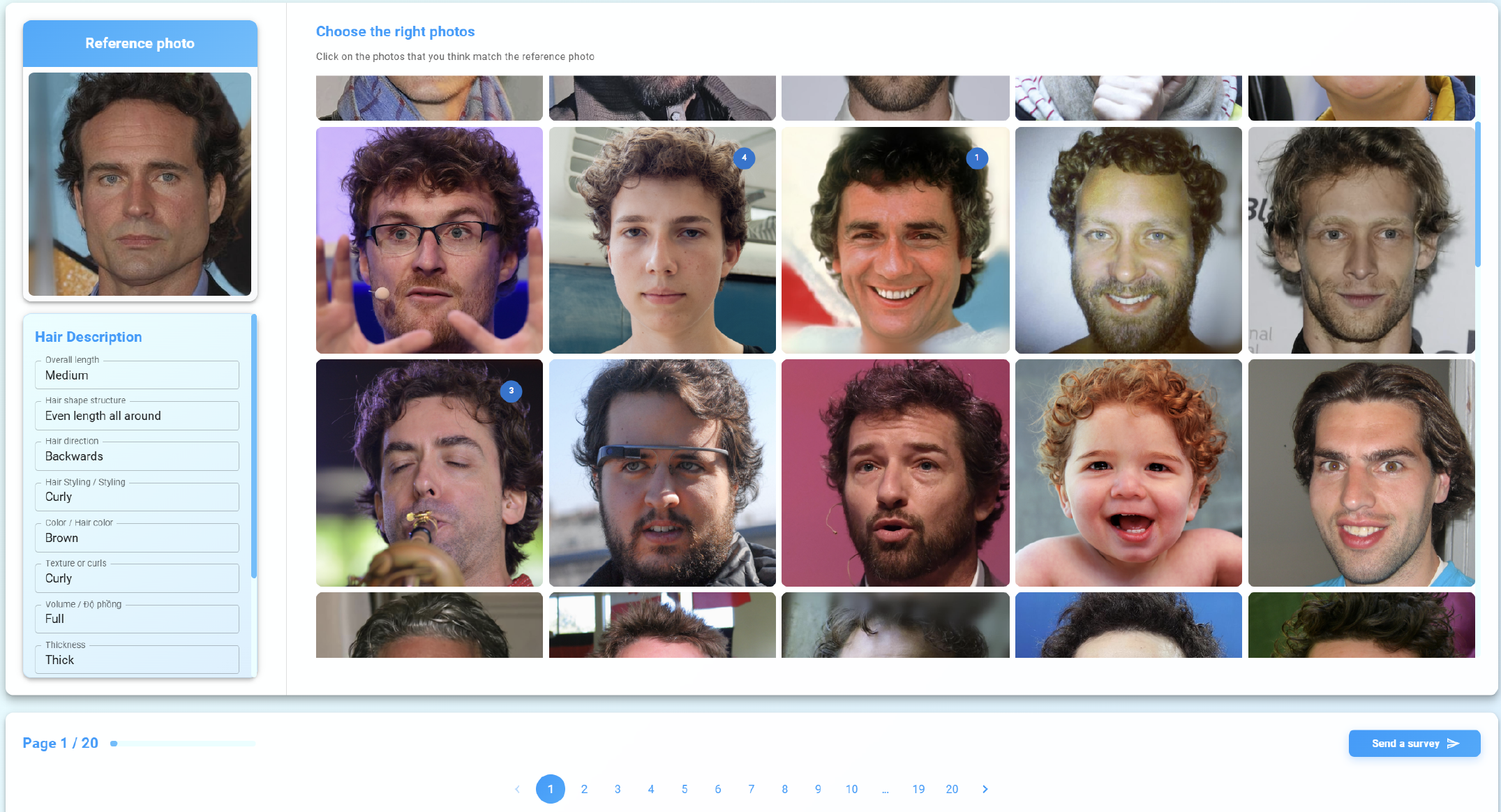}
        \caption{The interface of the platform used for hair reference image selection. Annotators choose images from the retriever's hairstyle candidate images.}
        \label{fig:platform-hair-annotation-main-view}
    \end{subfigure}
    \caption{ Annotation Platform User Interface}
    \label{fig:image-annotaion-platform}
\end{figure}

\subsection{Annotation Platforms}
\label{sec:annotation_platform}
\begin{figure}
    \centering
    \begin{subfigure}[b]{0.85\linewidth}
        \includegraphics[width=\linewidth]{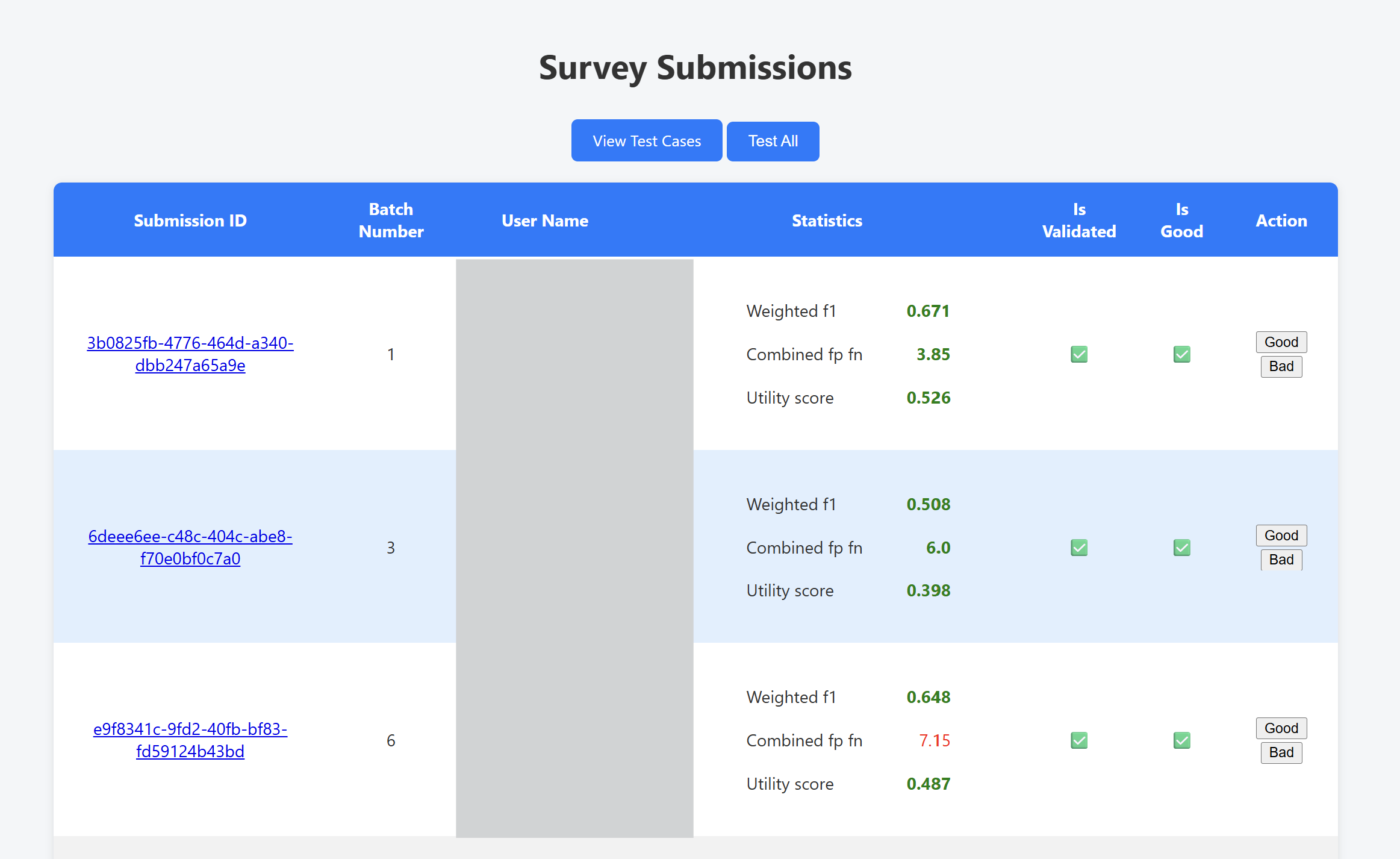}
        \caption{\textbf{Submission Overview.} The platform aggregates annotator submissions.}
    \end{subfigure}

    \begin{subfigure}[b]{0.85\linewidth}
        \includegraphics[width=\linewidth]{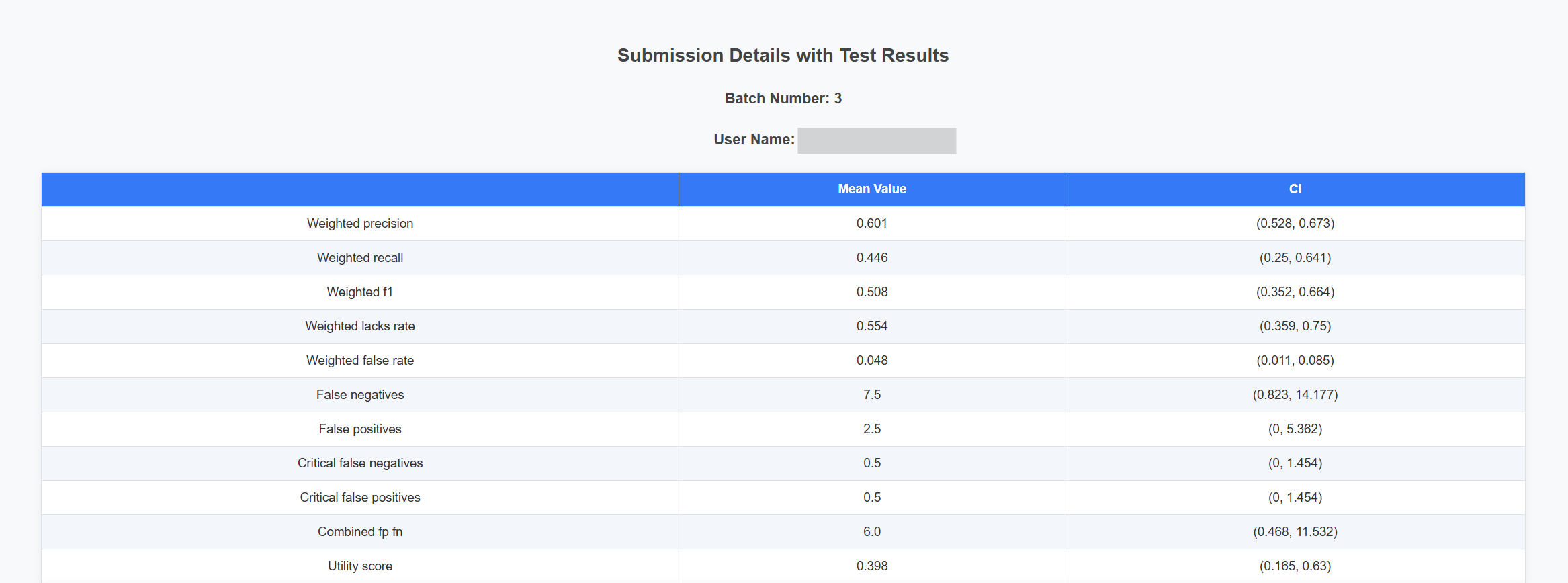}
        \caption{\textbf{Submission Detail.} For each submission, the system reports scores from the test cases and the mean score used for reliability assessment.}
    \end{subfigure}

    \begin{subfigure}[b]{0.85\linewidth}
        \includegraphics[width=\linewidth]{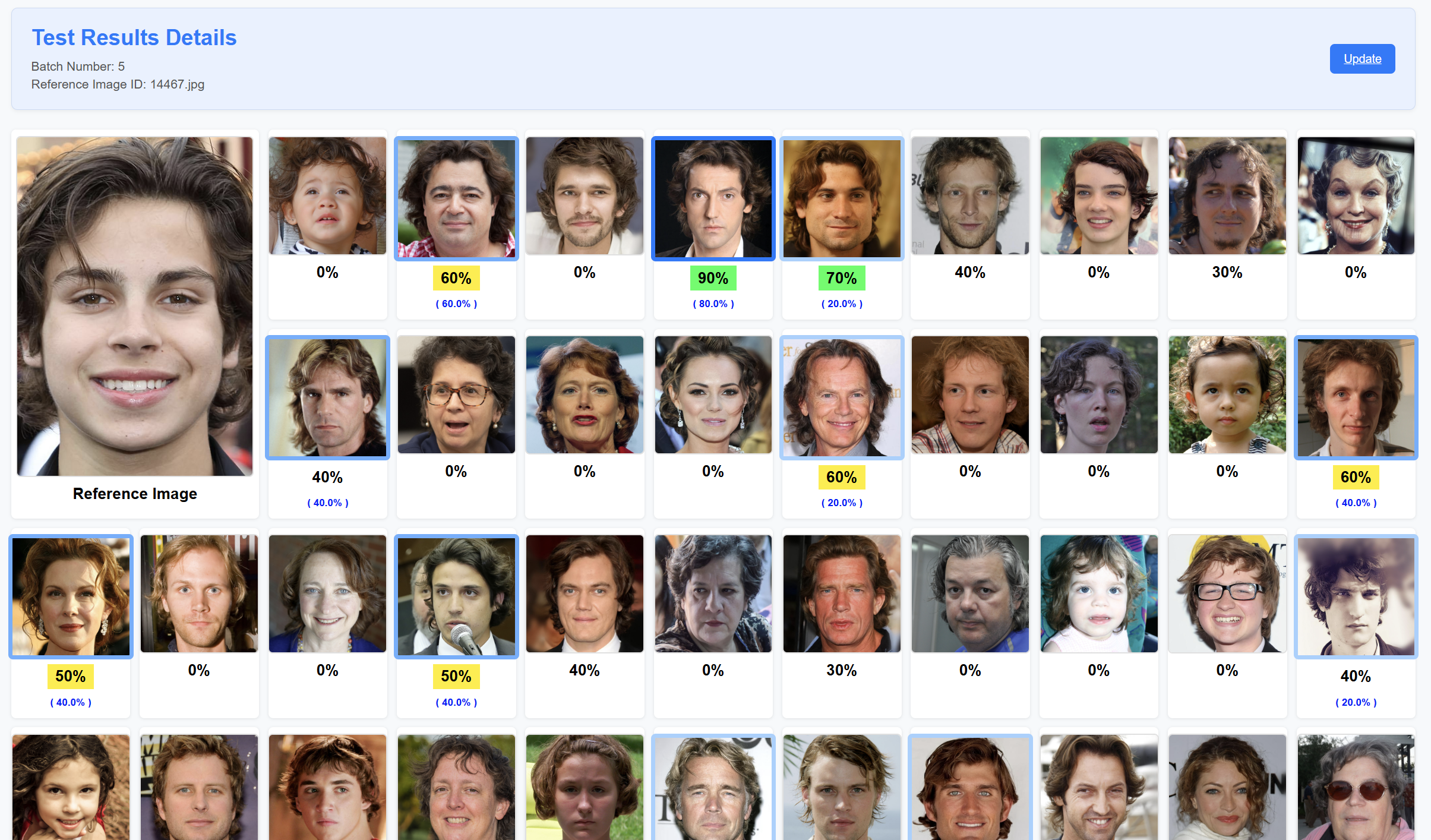}
        \caption{\textbf{Consensus Visualization.} The platform presents annotator agreement for each candidate image relative to the ground truths. Green indicates high-confidence matches, yellow denotes borderline cases, and the blue percentage reflects the consensus rate.}
    \end{subfigure}

    \caption{Annotator Evaluation Interface}
    \label{fig:image-evaluation-platform}
\end{figure}

We conduct annotation using a dedicated platform where annotators browse batches of queries and select hairstyle-consistent reference images from a retrieved pool. We additionally design a separate evaluation platform to construct a sampled 10\% ground-truth subset and measure annotator performance using several quantitative metrics. Figure~\ref{fig:image-annotaion-platform} presents the annotation interface, while Figure~\ref{fig:image-evaluation-platform} shows key views of the annotator evaluation system.

\section{Benchmark Dataset Construction}
\label{sec:benchmark-construction}
\subsection{Ground Truths Selection}
\label{sec:target-selection}
\begin{figure*}[t]
    \centering

    \begin{subfigure}[b]{0.49\linewidth}
        \centering
        \includegraphics[width=\linewidth]{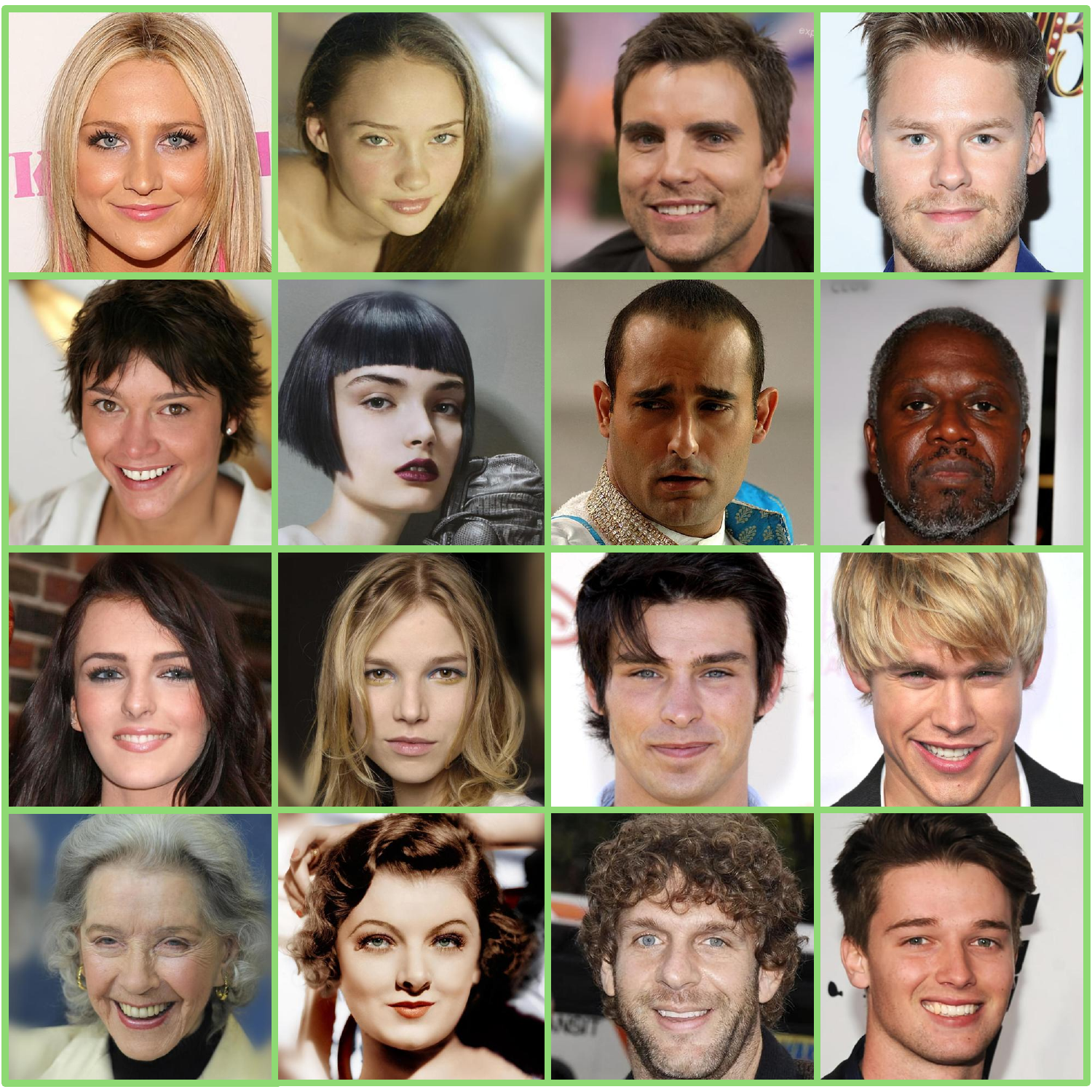}
        \caption{Accepted samples}
        \label{fig:is_gt}
    \end{subfigure}
    \hfill
    \begin{subfigure}[b]{0.49\linewidth}
        \centering
        \includegraphics[width=\linewidth]{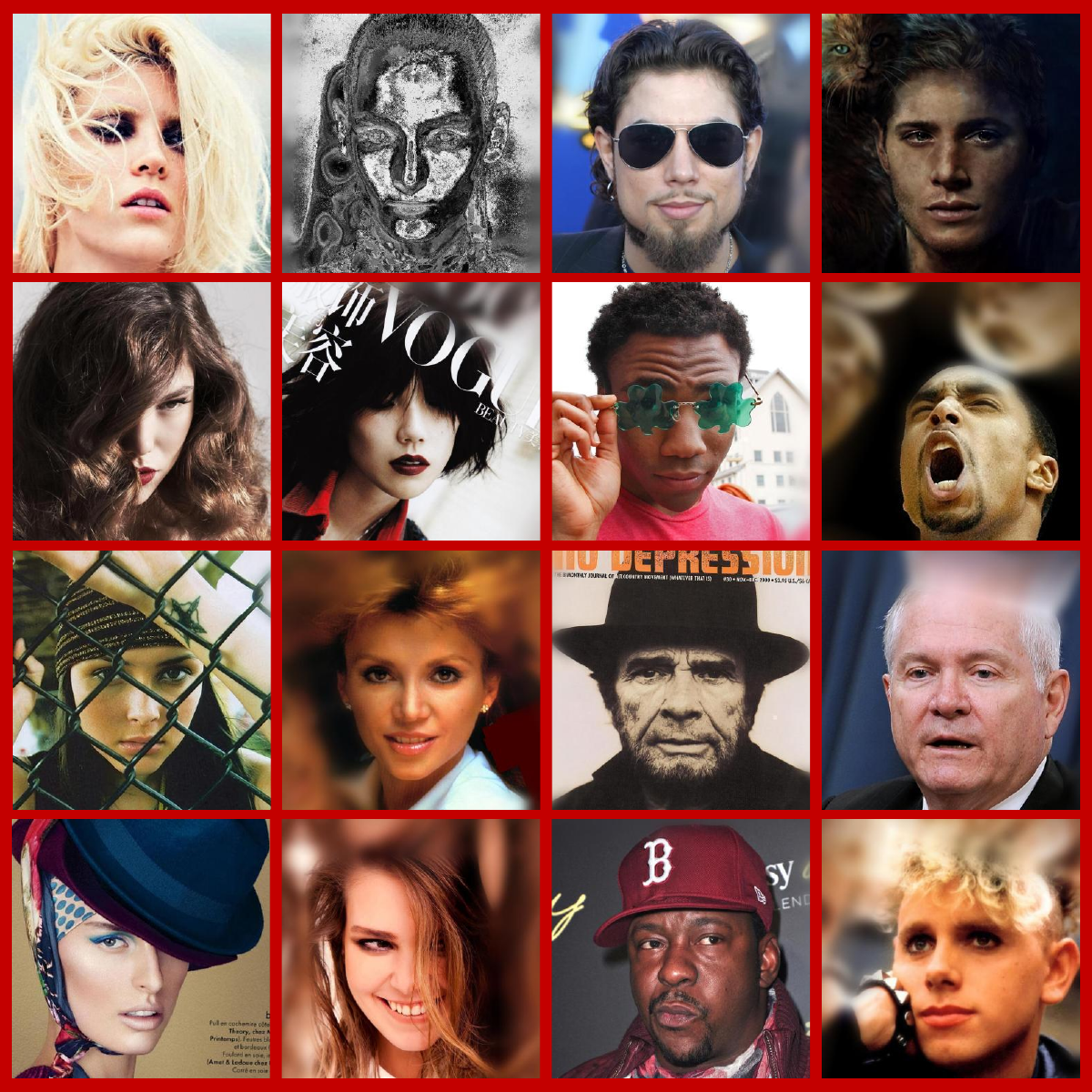}
        \caption{Rejected samples}
        \label{fig:not-gt}
    \end{subfigure}

    \caption{Examples from the 500 selected target images. (a) Accepted samples display clear hairstyle visibility and an acceptable pose. (b) Rejected samples illustrate common failure cases such as occlusion, low resolution, or ambiguous identity.}
    \label{fig:target-selection}
\end{figure*}

To construct a reliable set of ground-truth targets, we design a controlled sampling procedure that balances identity and hairstyle variation while maintaining high data quality. We select 500 target images, each representing a unique identity from CelebA-HQ \cite{karras2018_celeba_hq}. This scale provides sufficient diversity while remaining feasible for manual inspection. Candidates are restricted to identities with at least three images to ensure within-identity variability, after which one representative image per identity is randomly chosen. All selected samples are manually validated using four criteria: (1) clear visibility of the hairstyle without occlusion, (2) a suitable pose that preserves hair structure,(3) an approximately balanced gender distribution, and (4) a broad range of hairstyle lengths, textures, and styles. Figure~\ref{fig:is_gt} shows samples that meet our selection criteria, whereas Figure~\ref{fig:not-gt} presents examples that are excluded.
Many special cases such as bald or shaved heads are included in DFHR-Bench as a small, explicitly annotated subset (5\% of identities), allowing flexible inclusion or exclusion depending on experimental needs.

\subsection{Hair Encoder Implementation}
\label{sec:hair-encoder-implementation}

To enhance the quality and efficiency of annotation, we design a dedicated \textit{hair encoder}, enabling annotators to efficiently identify matching hairstyles without exhaustive manual searching. The encoder serves as the central module of our (Image--Image) retrieval system by ranking the top-50 hairstyle candidates for each query.

Hairstyle retrieval remains an underexplored problem, as most existing facial representation models primarily focus on identity rather than fine-grained hairstyle cues. In the absence of a hairstyle retrieval dataset, self-supervised learning methods—especially instance discrimination—are utilized, allowing the model to learn by attracting similar (positive) samples and repelling dissimilar (negative) ones. To address this gap, we develop a discriminative hairstyle encoder that explicitly models hairstyle-specific visual patterns while suppressing confounding factors such as facial identity and background.

\noindent\textbf{Hairstyle Representation Learning.} The encoder is built upon a ResNet-18 backbone and trained in a self-supervised contrastive learning manner. Given two augmented views $(x_i, x_i')$ of the same hairstyle image, the encoder $f(\cdot)$ produces corresponding embeddings $z_i = f(x_i)$ and $z_i' = f(x_i')$. The model is optimized with the contrastive objective:
\[
\mathcal{L} = - \log \frac{\exp(\text{sim}(z_i, z_i') / \tau)}{\sum_{k=1}^{N} \exp(\text{sim}(z_i, z_k) / \tau)},
\]
where $\text{sim}(\cdot)$ denotes cosine similarity and $\tau$ is a temperature parameter. This objective encourages positive pairs (two augmentations of the same hairstyle) to exhibit high similarity while pushing apart the embeddings of different hairstyles. 

\noindent\textbf{Training Data.} We leverage high-resolution facial datasets from FFHQ~\cite{karras2019_ffhq} and CelebA-HQ~\cite{karras2018_celeba_hq} to construct a diverse training corpus. Hair regions are extracted using segmentation masks to isolate hairstyle appearance from other facial components and contextual noise, ensuring that the encoder learns representations centered on hairstyle-relevant visual information.

Through this contrastive formulation and hairstyle-focused augmentations, the encoder learns to capture discriminative hairstyle attributes such as shape, texture, and volume, while remaining invariant to identity, background, and illumination variations. Although there is currently no established benchmark for evaluating hairstyle retrieval, the learned embeddings effectively narrow the candidate search space and substantially accelerate the annotation process. These embeddings form the foundation of our retrieval module, ensuring consistent and efficient hairstyle similarity estimation across annotators.

\subsection{Image Annotation Pipeline}
\label{sec:image-pipeline}

\begin{figure*}
    \centering
    \includegraphics[width=1\linewidth]{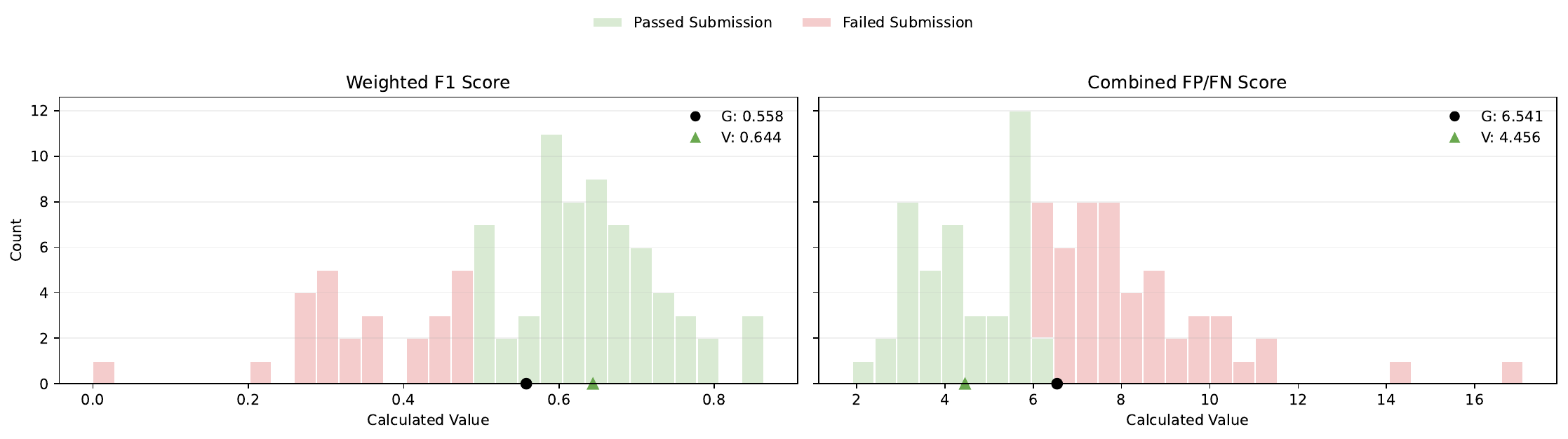}
    \caption{\textbf{Distribution of annotator performance.} Each histogram separates submissions that passed qualification (green) from those that failed (red). Markers indicate the mean scores of the full annotator population (G) and the subset that passed qualification (V).}
    \label{fig:evaluation-value}
\end{figure*}
\noindent\textbf{Annotator Recruitment and Training.}
We recruited fifty paid annotators and assigned them batches of 20 queries, each consisting of one \emph{anchor image} and 50 candidate images retrieved by our hair encoder. Annotators selected all candidates sharing a visually consistent hairstyle with the anchor. Before the main task, each annotator completed a five-query guided survey designed to illustrate the criteria and how annotation quality would be evaluated. Afterward, they received feedback using precision, recall, F1 score, and false-positive/false-negative counts computed against an internally curated ground truth. This stage served for familiarization and ensured that all annotators understood the expectations.

\noindent\textbf{Qualification and Performance Monitoring.}
During annotation, 10\% of all queries were embedded as hidden test cases with verified ground truths.  
Annotator reliability was assessed using two main metrics: a weighted F1 score and a combined FP/FN score.  
An annotator was considered \emph{qualified} if they satisfied 
$F_1 \ge 0.5$ or
$\text{FPFN} \le 6$.
Thresholds are determined through an empirical calibration study comparing highly reliable annotators with deliberately noisy ones. This design filters out random or careless submissions while accommodating natural variation in human judgment.
Figure~\ref{fig:evaluation-value} reveals that most annotators align well with the curated ground truth. As both metrics explain different aspects of annotators' performance, applying both metrics jointly filters true low-quality submissions while preserving normal annotator variation, preventing overfitting to our curated ground truths.

\medskip
\noindent\textbf{Ground-Truth Representation.}
For each query, let $\mathcal{I}$ denote the candidate image set and let
\[
w_i \in [0,1], \quad i \in \mathcal{I},
\]
be the ground-truth confidence assigned to image $i$.  
Images with $w_i \ge \tau$ (with threshold $\tau = 0.5$) are treated as \emph{valid} references:
\[
\mathcal{V} = \{ i \in \mathcal{I} \mid w_i \ge \tau \}.
\]
Given an annotator submission $S \subseteq \mathcal{I}$, the subset of correctly chosen images is
\[
C = S \cap \mathcal{V} = \{ i \in S \mid w_i \ge \tau \}.
\]

\medskip
\noindent\textbf{Weighted Precision, Recall, and F1.}
We define the weighted precision as
\begin{equation}
P = 
\frac{\displaystyle \sum_{i \in C} w_i^{2}}
     {\displaystyle \sum_{i \in S,\, w_i \ge \tau} w_i^{2}
      + \sum_{i \in S,\, w_i < \tau} (1 - w_i)^{2}},
\end{equation}
where correct selections are rewarded by $w_i^{2}$ and incorrect selections are penalized by $(1-w_i)^{2}$.  
The weighted recall is
\begin{equation}
R = 
\frac{\displaystyle \sum_{i \in C} w_i^{2}}
     {\displaystyle \sum_{i \in \mathcal{V}} w_i^{2}}.
\end{equation}
The weighted F1 score is then the harmonic mean of $P$ and $R$:
\begin{equation}
F_1 =
\begin{cases}
\displaystyle \frac{2PR}{P+R}, & \text{if } P+R > 0,\\[4pt]
0, & \text{otherwise}.
\end{cases}
\end{equation}

\medskip
\noindent\textbf{Combined FP/FN Score.}
We count false positives and false negatives as
\begin{align}
\text{FP} &= |\{ i \in S \mid w_i < \tau \}|, \\
\text{FN} &= |\{ i \notin S \mid w_i \ge \tau \}|.
\end{align}
The combined FP/FN score is defined as
\begin{equation}
\text{FPFN} = 0.7\,\text{FP} + 0.3\,\text{FN},
\end{equation}
which penalizes false positives more heavily than false negatives and thus encourages conservative, high-precision behavior. This metric reflects our preference for high precision: selecting an incorrect image is considered more harmful than omitting a borderline valid one.  

\medskip

\noindent\textbf{Quality Control and Disagreement Handling.}
We record that there is a total of 10 submissions that failed to meet the qualifications.
These batches were discarded and fully reassigned. For inherently ambiguous cases, annotations were kept but supplemented with additional annotators to capture multiple valid interpretations.

\noindent\textbf{Consensus Aggregation.}
Each query was annotated by an average of 3.4 annotators. A candidate image was accepted as a valid reference if at least half of the annotators selected it. This majority-vote rule provides a stable consensus while avoiding over-reliance on any single expert.

\subsection{Text Annotation Pipeline}
\label{sec:text-pipeline}

\begin{figure}
    \centering
    \includegraphics[width=1\linewidth]{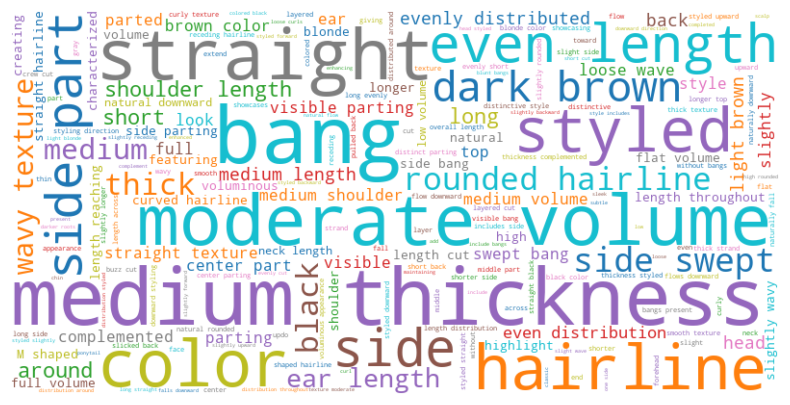}
    \caption{Word cloud of hairstyle caption semantics showing the most frequently used descriptive terms}
    \label{fig:caption-word-cloud}
\end{figure}

\begin{figure}
    \centering
    \includegraphics[width=1\linewidth]{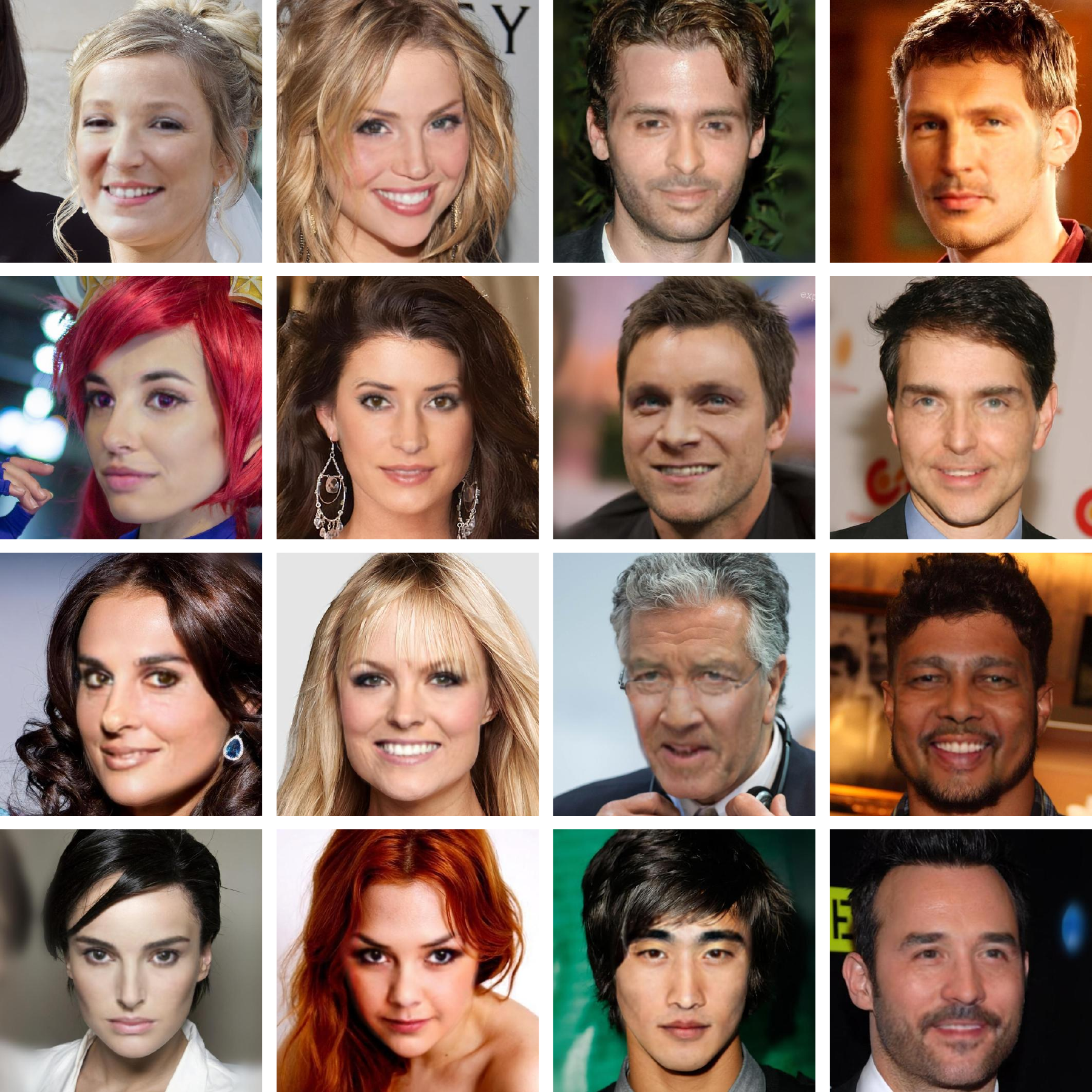}
    \caption{Deepfake images generation}
    \label{fig:deepfakes}
\end{figure}

A \textbf{two-round annotation pipeline} is constructed to produce fine-grained and semantically consistent hairstyle descriptions for an image set with 500 anchor images (Figure~\ref{fig:text-pipeline}). Each round contains an LLM-driven synthesis stage and a human validation stage, ensuring a balance between linguistic diversity and controlled semantic precision. 

\noindent\textbf{Round~1: Attribute-level Annotation.} This round focuses on producing structured attribute-level descriptions using the key attributes defined in \S\ref{sec:shared-guidelines}, LLM receives image-conditioned prompts and generates one textual candidate per attribute for each anchor image, guided by a prompt specified in Table~\ref{tab:prompt1-json} to avoid ambiguity and maintain stylistic uniformity.
The generated attribute descriptions are then reviewed by four expert annotators working in pairs. One annotator refines the LLM output for accuracy and clarity, while the second verifies the result for consistency and discriminability across attributes. Disagreements are resolved collaboratively, and the validated attributes form a stable semantic basis for Round~2. 

\noindent\textbf{Round~2: Hair Description Annotation.} Using the validated attribute set, we leverage LLM to produce up to five complete hairstyle sentences per anchor image. The prompt used to enable the synthesis is provided in Table~\ref{tab:prompt2-text}.
All Round~2 sentences undergo the same two-tier expert validation. Annotators ensure that each description faithfully reflects all validated attributes, maintains grammatical fluency, and remains stylistically consistent with prior accepted examples. Most sentences require only minor edits; sentences that fail are simply discarded, as fivefold generation redundancy provides sufficient diversity without re-prompting.

The text-annotation pipeline relies on unanimous agreement from expert annotators rather than numerical scoring, with minor inconsistencies resolved through manual verification. For each of the 500 anchor identities, this process produces up to five high-quality hairstyle descriptions. When paired with multiple identity-matched visual references, the pipeline yields approximately 68,760 validated text–image triplets $\langle q^{\text{txt}}_{\text{hair}}, I_{\text{target}} \rangle$. Figure~\ref{fig:caption-word-cloud} summarizes the semantic diversity of the captions, highlighting commonly used vocabulary.

\begin{figure*}[t]
    \centering

    \begin{subfigure}[b]{0.5\linewidth}
        \centering
        \includegraphics[width=\linewidth]{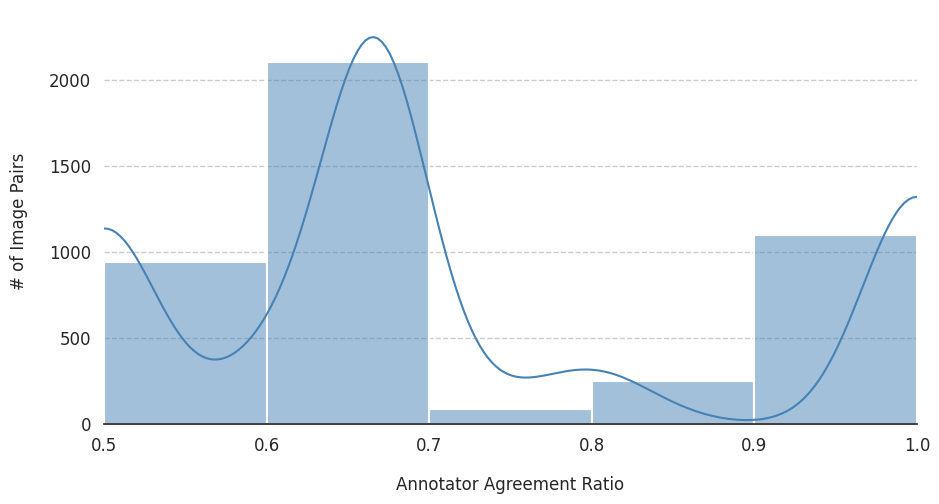}
        \caption{Distribution of annotator agreement ratio across valid image pairs}
        \label{fig:annotator_agreement_ratio}
    \end{subfigure}
    \hfill
    \begin{subfigure}[b]{0.49\linewidth} 
        \centering
        \includegraphics[width=\linewidth]{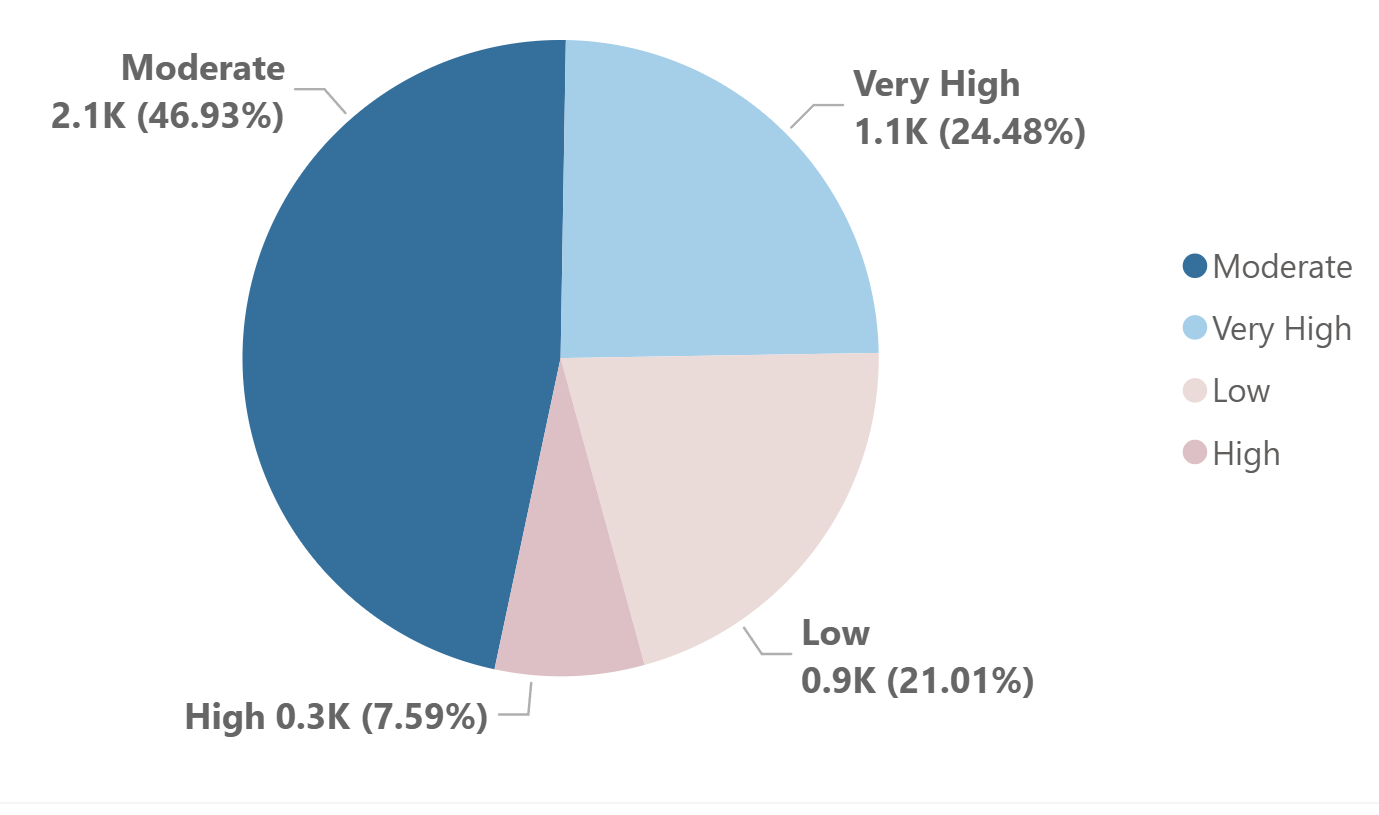}
        \caption{The proportion of annotator agreement buckets across valid image pairs}
        \label{fig:pie_annotator_agreement}
    \end{subfigure}

    \caption{The annotation process yielded 4.4K face–hair pairs that met our agreement threshold. The charts illustrate the distribution of annotator agreement across these validated pairs.}
    \label{fig:annotator_agreement_visualization}
\end{figure*}

\subsection{Image Synthesis Algorithms}
\label{sec:synthesis-algorithm}

To further diversify the retrieval pool, we synthesize additional identity-consistent images using an \textbf{off-the-shelf deepfake pipeline}. In particular, we employ \textbf{Roop}, a robust one-shot face-swapping method known for its high fidelity and stable identity preservation.
During benchmark construction, each query $(I_\text{face}, I_\text{hair})$ corresponds to a single ground-truth image $I_\text{target}$. Annotators are shown the top-50 retrieved candidates and select those that exhibit hairstyles visually similar to $I_\text{target}$. The remaining non-selected candidates often still contain useful hairstyle cues. We exploit these “near-miss” cases by applying controlled identity swapping, enabling us to expand the dataset without additional manual annotation (detailed in Algorithm~\ref{algo:deepfake}).

\begin{algorithm}[H]
\caption{Identity-Preserving Hairstyle Synthesis}
\label{algo:deepfake}
\begin{algorithmic}[1]
\REQUIRE Query $(I_\text{id}, I_\text{hair})$, target image $I_\text{target}$
\STATE Retrieve top-50 candidates $\mathcal{C} = \{C_i\}_{i=1}^{50}$ most similar to $I_\text{target}$
\STATE Let $S^+ \subset \mathcal{C}$ be the annotator-approved matches
\STATE Select the top-5 near-miss candidates $S^- \subset (\mathcal{C} \setminus S^+)$ closest to $I_\text{target}$ in embedding space
\FOR{each $C_j \in S^-$}
    \STATE $\hat{I}_j \leftarrow \text{Deepfake}(I_\text{id}, C_j)$
    \STATE $\mathcal{D}' \leftarrow \mathcal{D}' \cup \{\hat{I}_j\}$
\ENDFOR
\RETURN Augmented retrieval pool $\mathcal{D}'$
\end{algorithmic}
\end{algorithm}

\noindent\textbf{Discussion.}  
As illustrated in Figure~\ref{fig:deepfakes}, modern deepfake frameworks are capable of producing highly realistic face-swapped images with minimal artifacts. By leveraging \textbf{Roop}, we transform near-miss candidates, which contain suitable hairstyles but incorrect identities, into identity-consistent synthetic samples. This augmentation significantly enriches intra-identity hairstyle diversity and proves that off-the-shelf deepfake tools are a viable pipeline for constructing challenging, balanced retrieval benchmarks.

\section{Additional Dataset Analysis}
\label{supp:addiotnal_dataset_analysis}

\subsection{Inter-Annotator Agreement}
\label{supp:bench_extended}

The number of annotators per query in DFHR-Bench ranges from 2 to 6. To assess reliability, we compute Cohen’s and Fleiss’ $\kappa$ \cite{kappa} across groups stratified by annotator count, yielding an overall mean of $\kappa{=}0.30$. Krippendorff’s $\alpha$ \cite{alpha} shows a comparable value of $0.29$, reflecting the inherent difficulty of the task, which requires fine-grained visual comparison and disentangling multiple hairstyle factors. Although our detailed guidelines reduce ambiguity, we intentionally allow limited self-variance to preserve perceptual diversity.
It is important to note that inter-annotator agreement is computed per $\langle I_{\text{result}}, I_{\text{hair}}^{\text{cand}} \rangle$ pair. Because only about 18\% of candidate pairs are considered valid, the label distribution is highly imbalanced, causing conventional $\kappa$ and $\alpha$ to underestimate consensus due to the prevalence paradox \cite{kappa-paradox, ac1}. To correct for this effect, we also report Gwet’s AC1 \cite{ac1}, which provides a more reliable estimate, achieving a coefficient of $0.66$ with an observed pairwise agreement (PA) of $0.78$.
A pair is deemed \textit{officially valid} when selected by more than 50\% of annotators. Among these valid pairs, the mean agreement ratio reaches $0.71$, with roughly 46\% falling between 0.60 and 0.75, indicating moderate consensus. These statistics show that DFHR-Bench achieves a strong balance between annotation precision and controlled perceptual variability, faithfully capturing both the complexity of the task and the natural diversity of human judgments.

\subsection{Qualitative Examples}
\label{sec:qualitative_examples}
\begin{figure*}[t]
  \centering
  \begin{subfigure}{\textwidth}
    \includegraphics[width=0.925\linewidth]{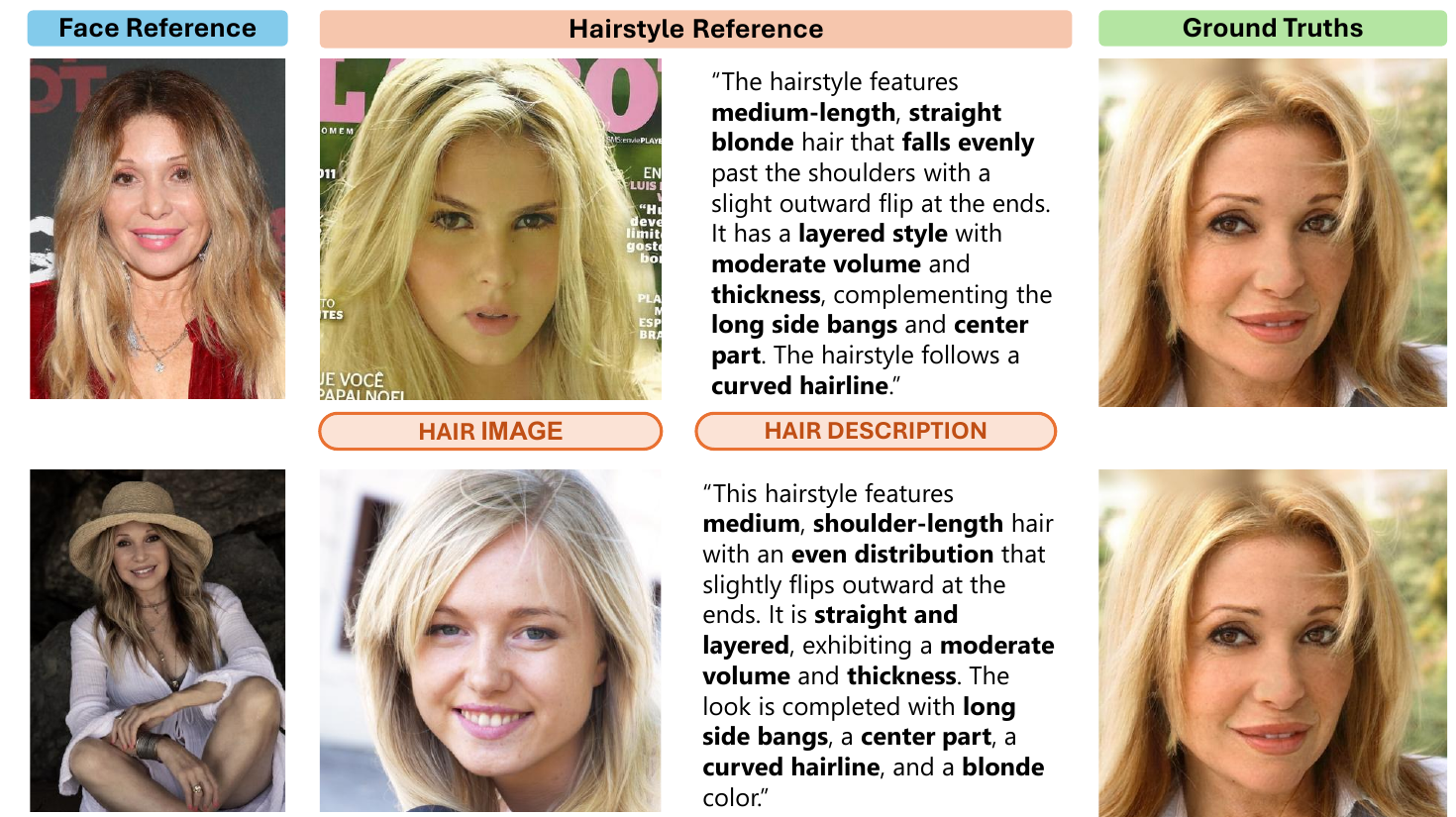}
    \caption{Example 1}
  \end{subfigure}

  \vspace{0.5em}

  \begin{subfigure}{\textwidth}
    \includegraphics[width=0.942\linewidth]{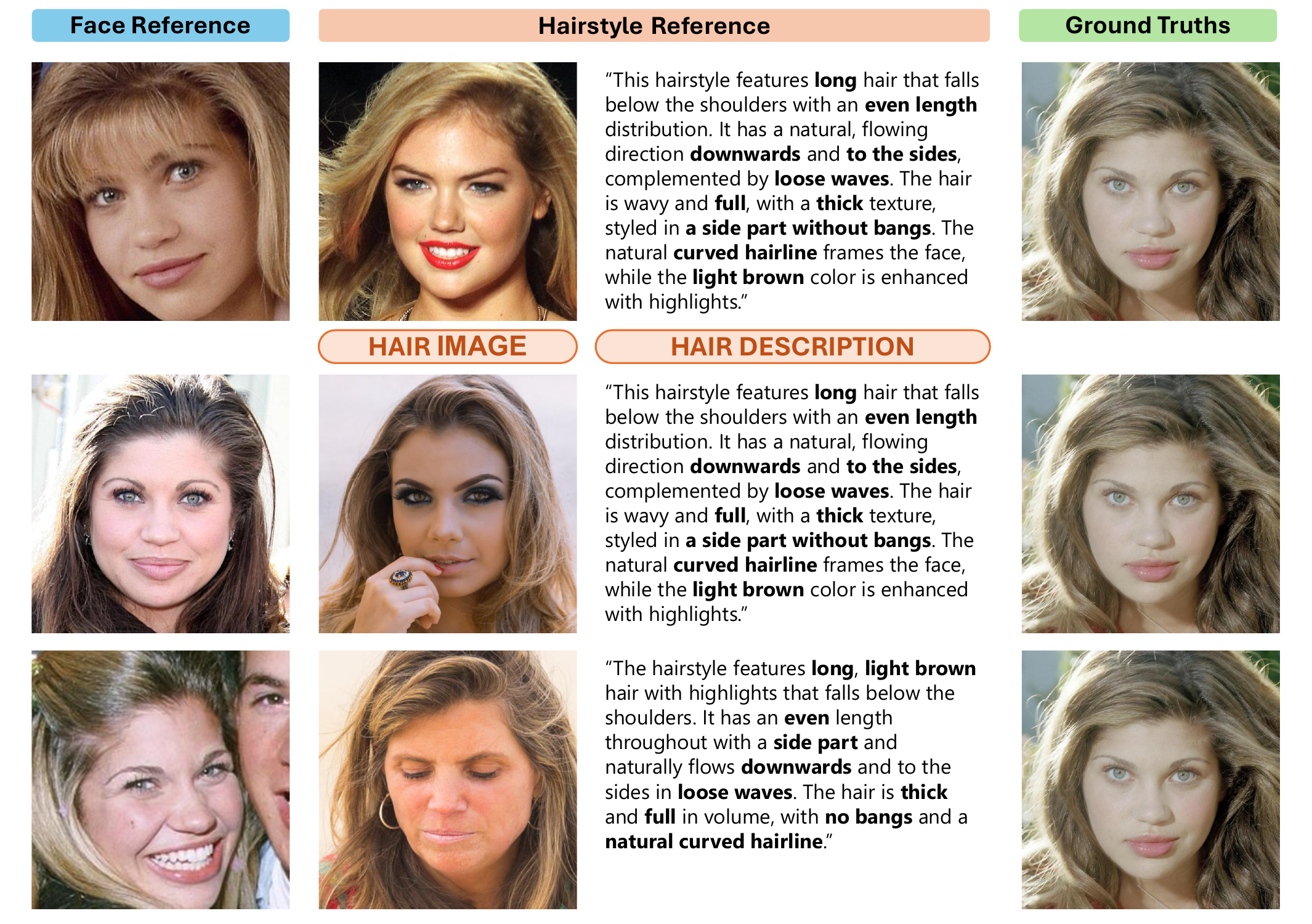}
    \caption{Example 2}
  \end{subfigure}
  \caption{Dataset Qualitative Examples}
  \label{fig:triplets1}
\end{figure*}

\begin{figure*}[t]
  \centering
  \begin{subfigure}{\textwidth}
    \includegraphics[width=1\linewidth]{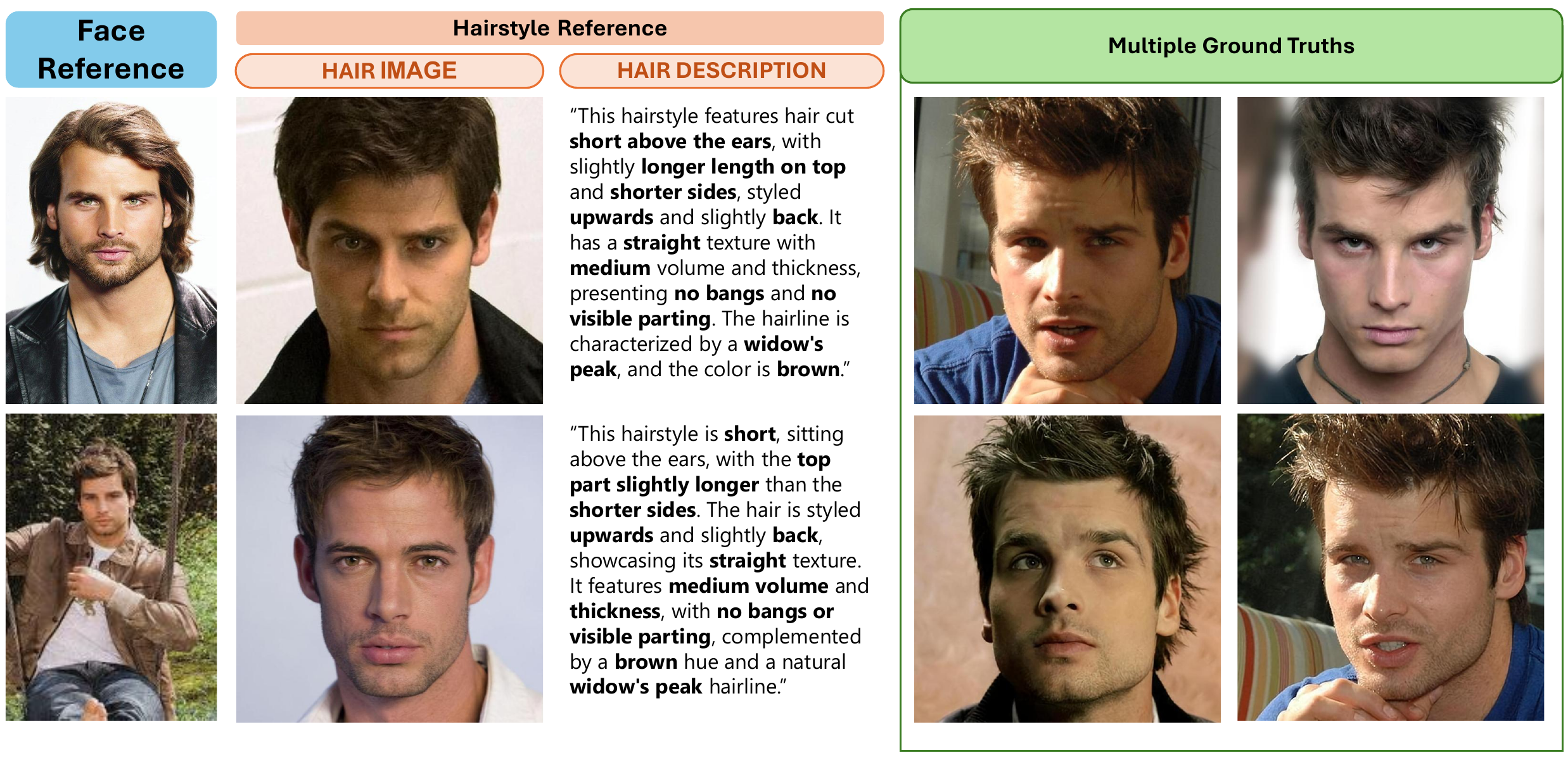}
    \caption{Example 3}
  \end{subfigure}

  \vspace{0.5em}

  \begin{subfigure}{\textwidth}
    \includegraphics[width=1\linewidth]{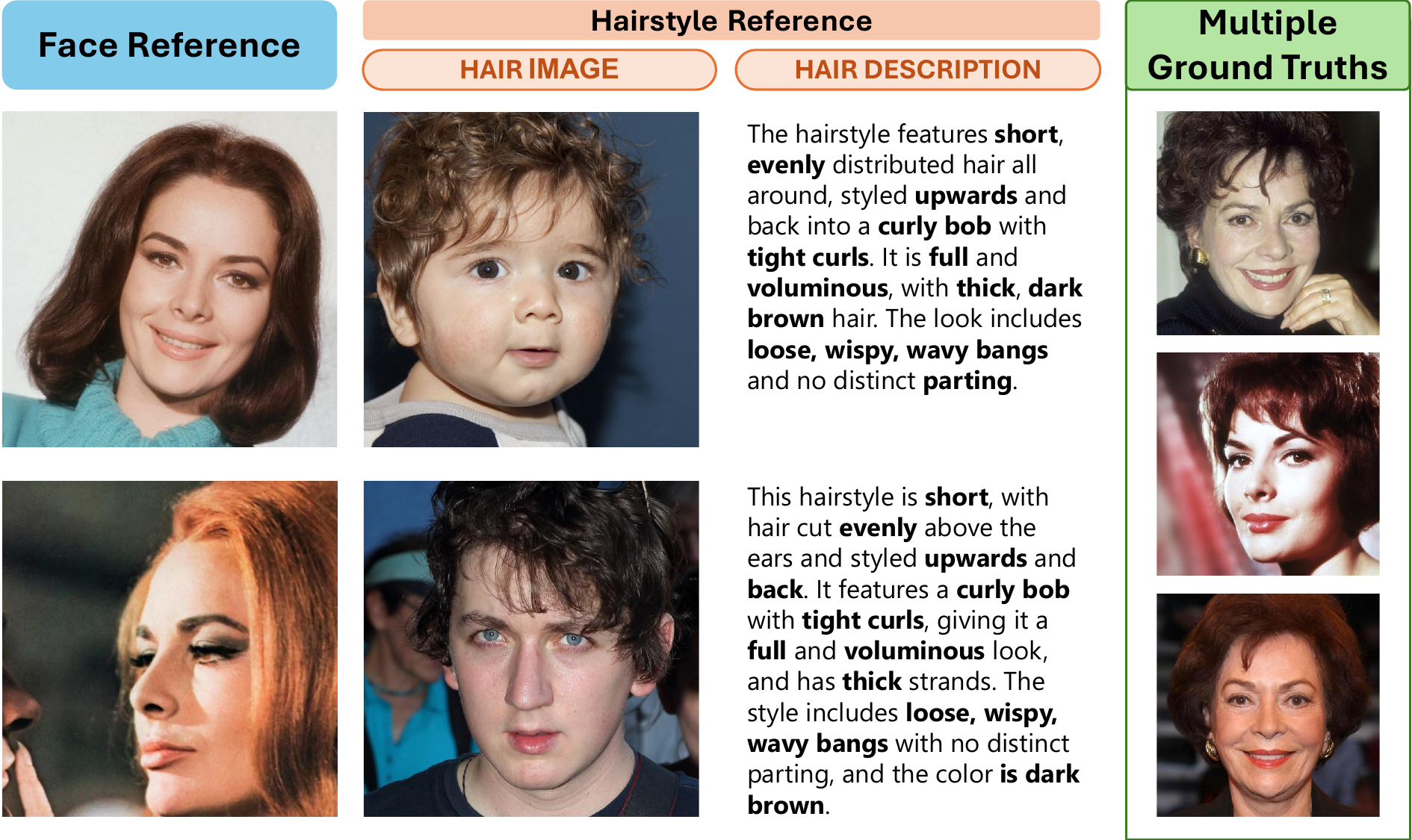}
    \caption{Example 4}
  \end{subfigure}
  \caption{Dataset Qualitative Examples}
  \label{fig:triplets2}
\end{figure*}

\begin{figure*}[t]
  \centering
  \begin{subfigure}{\textwidth}
    \includegraphics[width=1\linewidth]{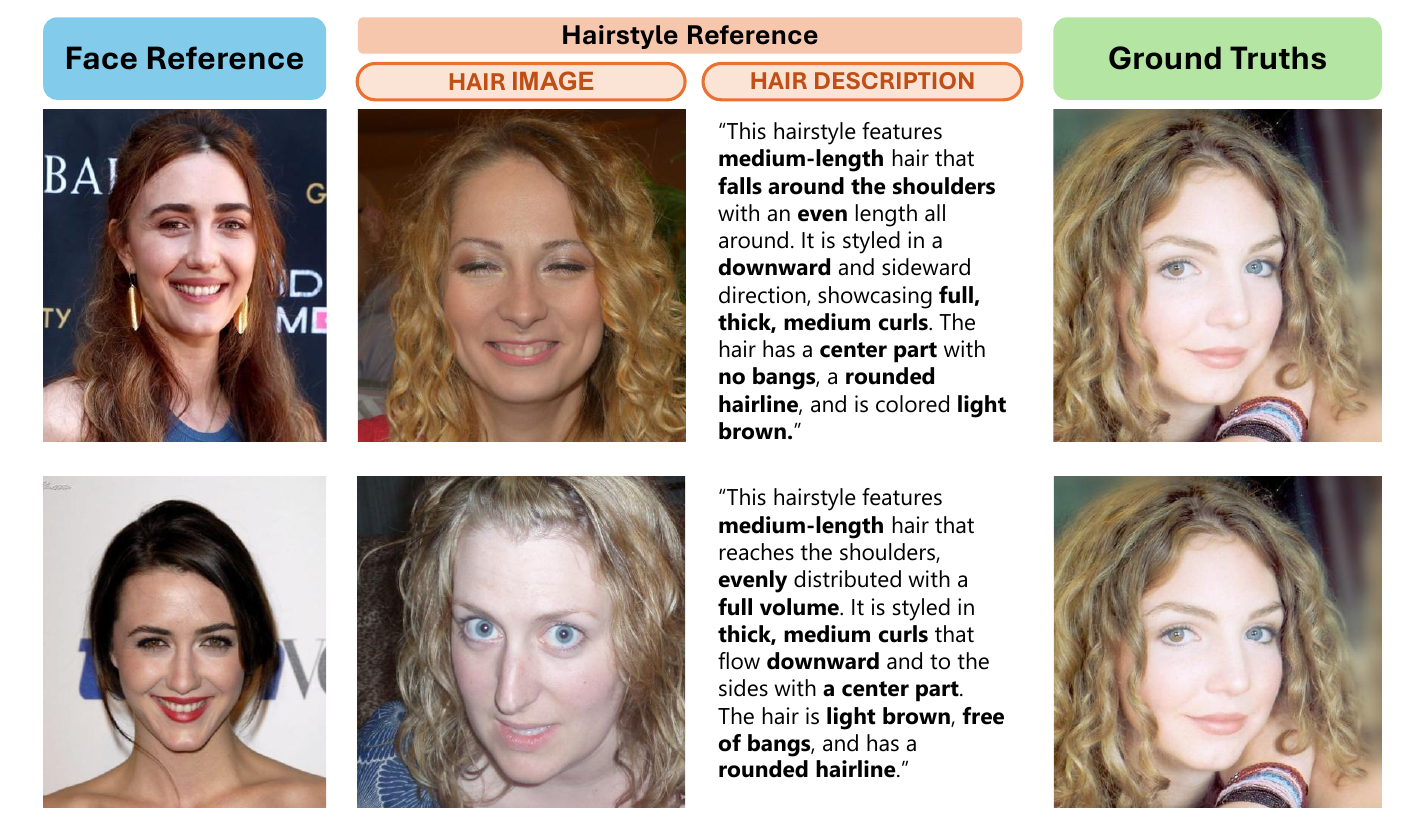}
    \caption{Example 5}
  \end{subfigure}

  \vspace{0.5em}

  \begin{subfigure}{\textwidth}
    \includegraphics[width=1\linewidth]{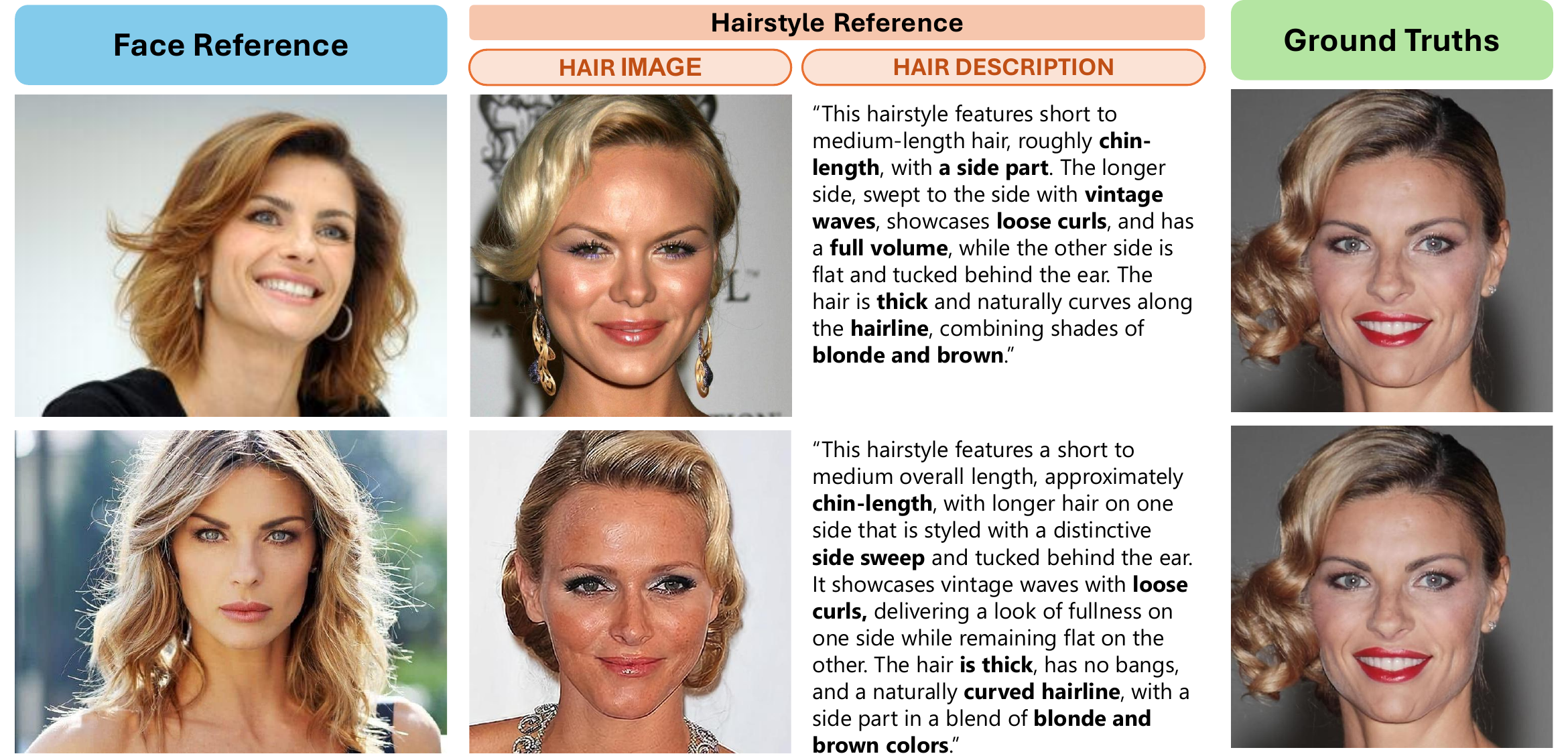}
    \caption{Example 6}
  \end{subfigure}
  \caption{Dataset Qualitative Examples}
  \label{fig:triplets3}
\end{figure*}

\begin{figure*}[t]
  \centering
  \begin{subfigure}{\textwidth}
    \includegraphics[width=1\linewidth]{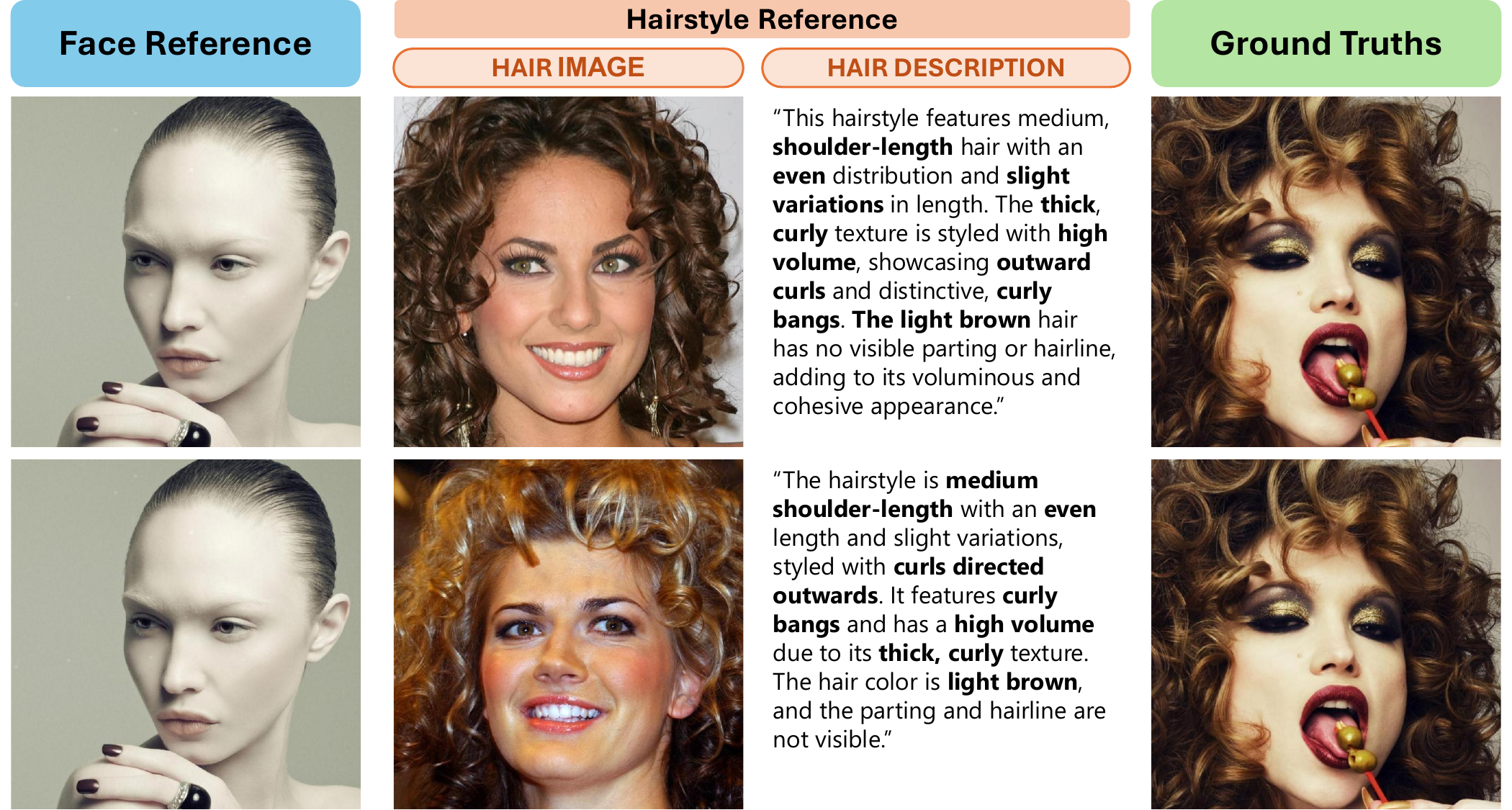}
    \caption{Example 7}
  \end{subfigure}

  \vspace{0.5em}

  \begin{subfigure}{\textwidth}
    \includegraphics[width=1\linewidth]{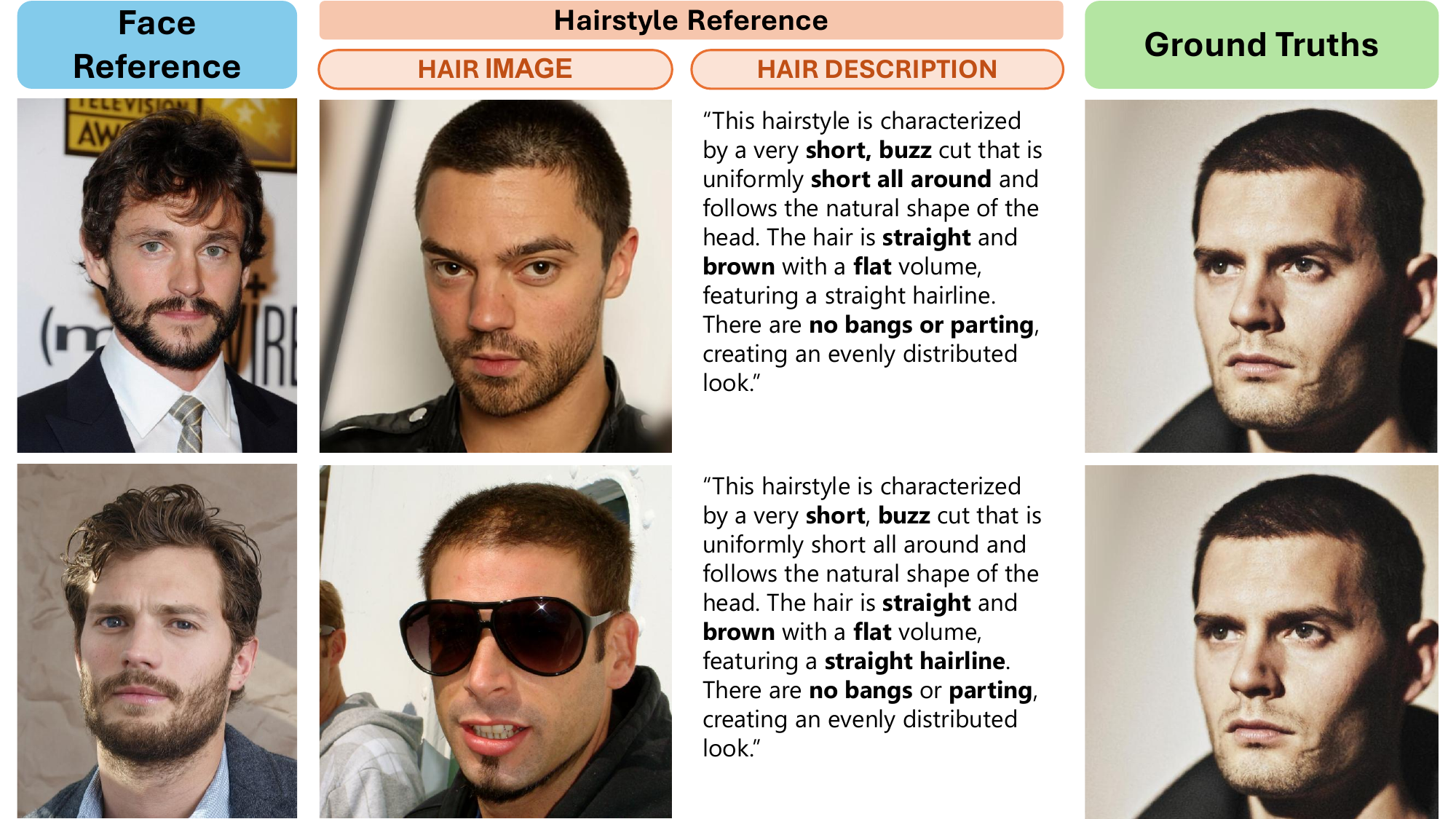}
    \caption{Example 8}
  \end{subfigure}
  \caption{Dataset Qualitative Examples}
  \label{fig:triplets4}
\end{figure*}

\begin{figure*}[t]
  \centering
  \begin{subfigure}{\textwidth}
    \includegraphics[width=1\linewidth]{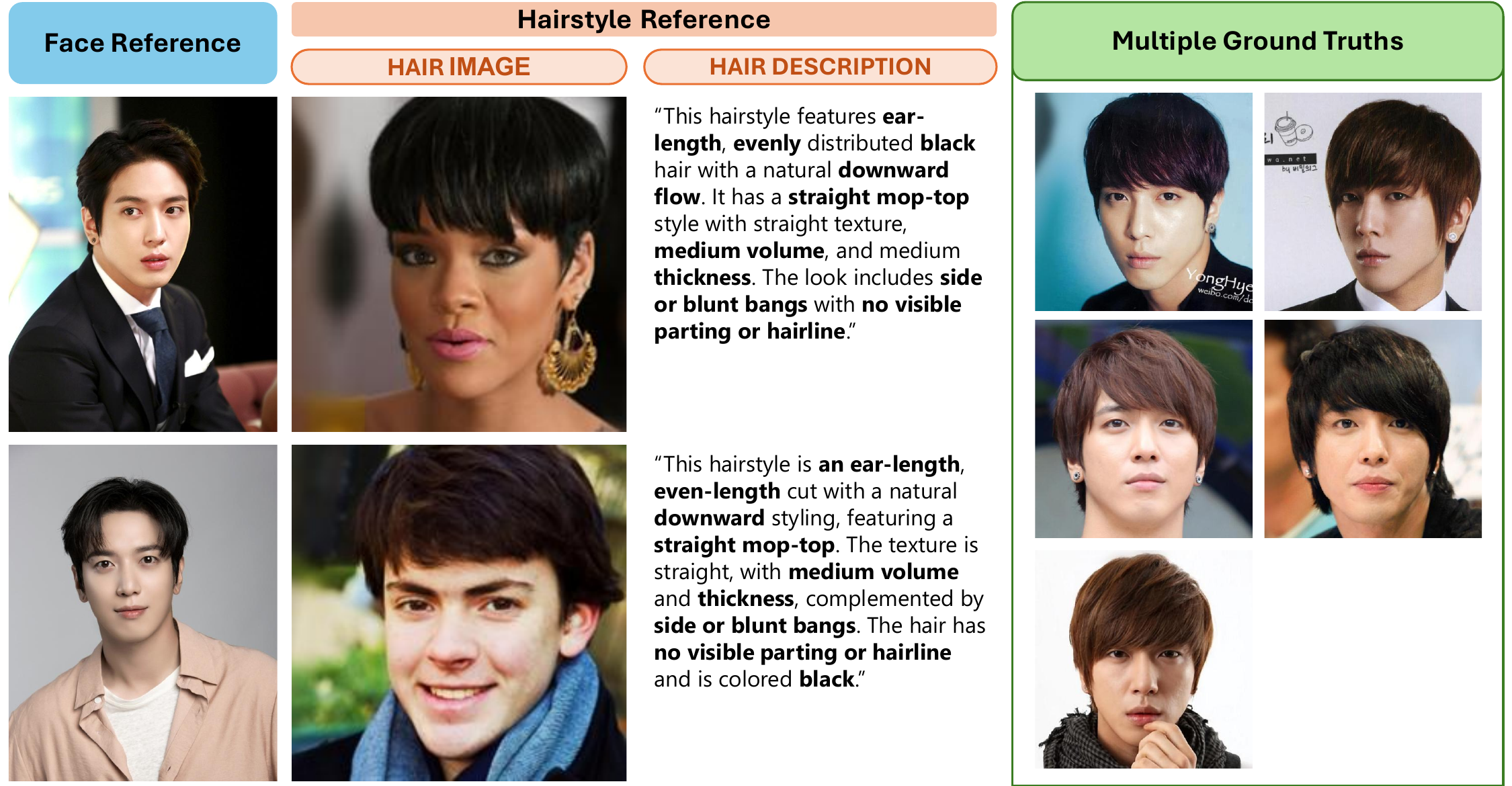}
    \caption{Example 9}
  \end{subfigure}

  \vspace{0.5em}

  \begin{subfigure}{\textwidth}
    \includegraphics[width=1\linewidth]{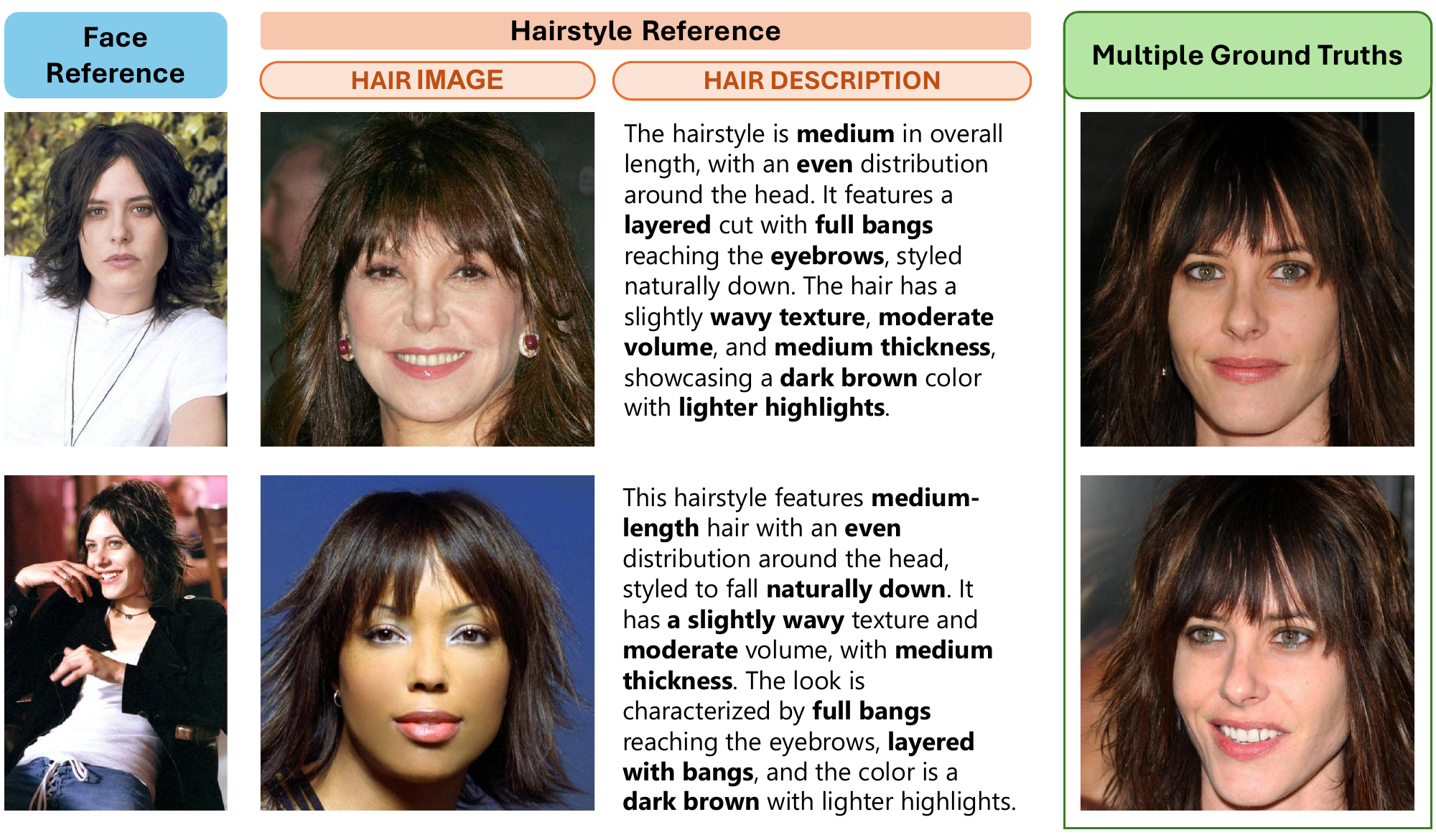}
    \caption{Example 10}
  \end{subfigure}
  \caption{Dataset Qualitative Examples}
  \label{fig:triplets5}
\end{figure*}

\begin{figure*}
    \centering
    \includegraphics[width=\linewidth]{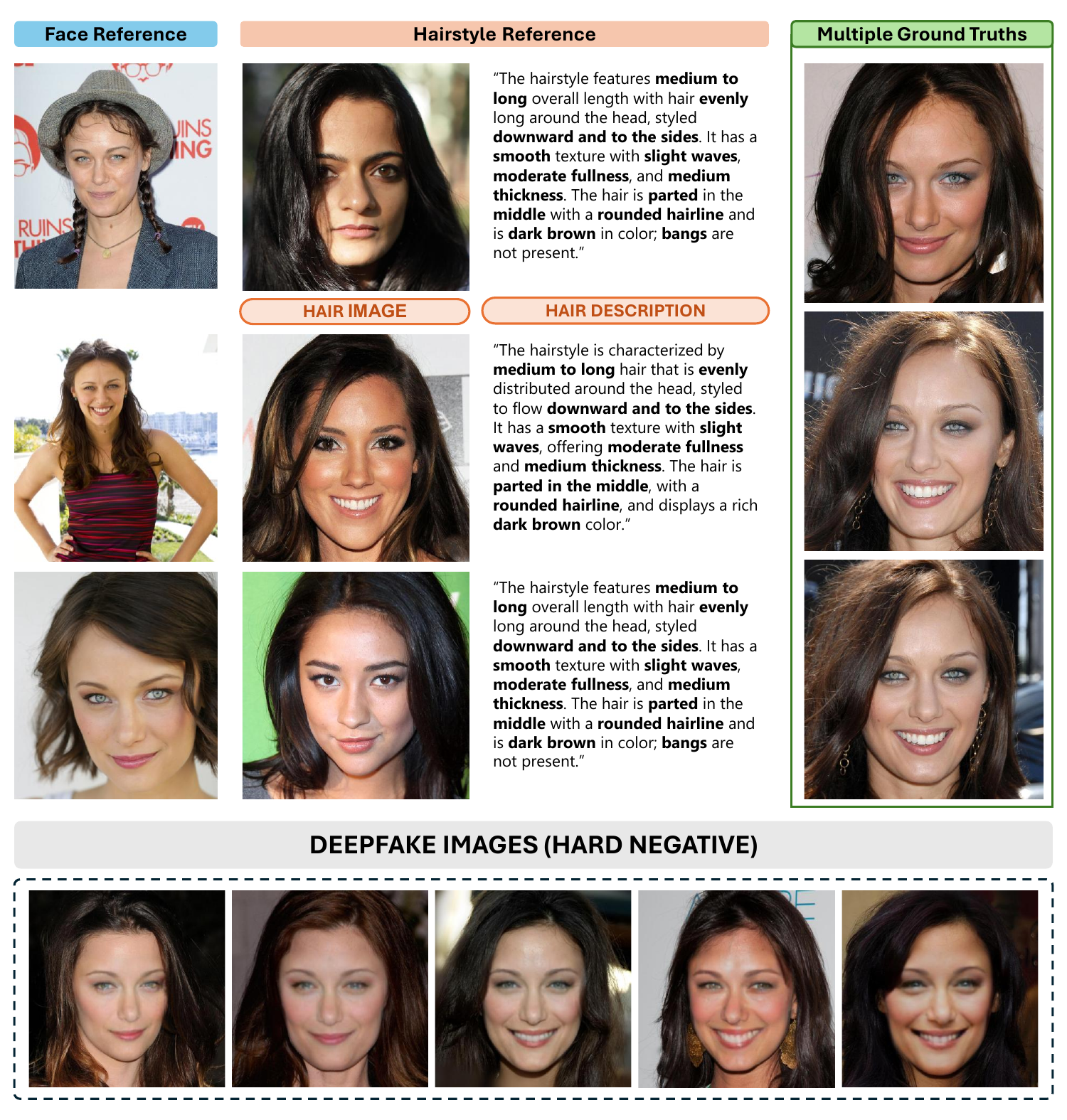}
    \caption{Qualitative examples with deepfake images}
    \label{fig:triplets6}
\end{figure*}

\begin{figure*}
    \centering
    \includegraphics[width=0.83\linewidth]{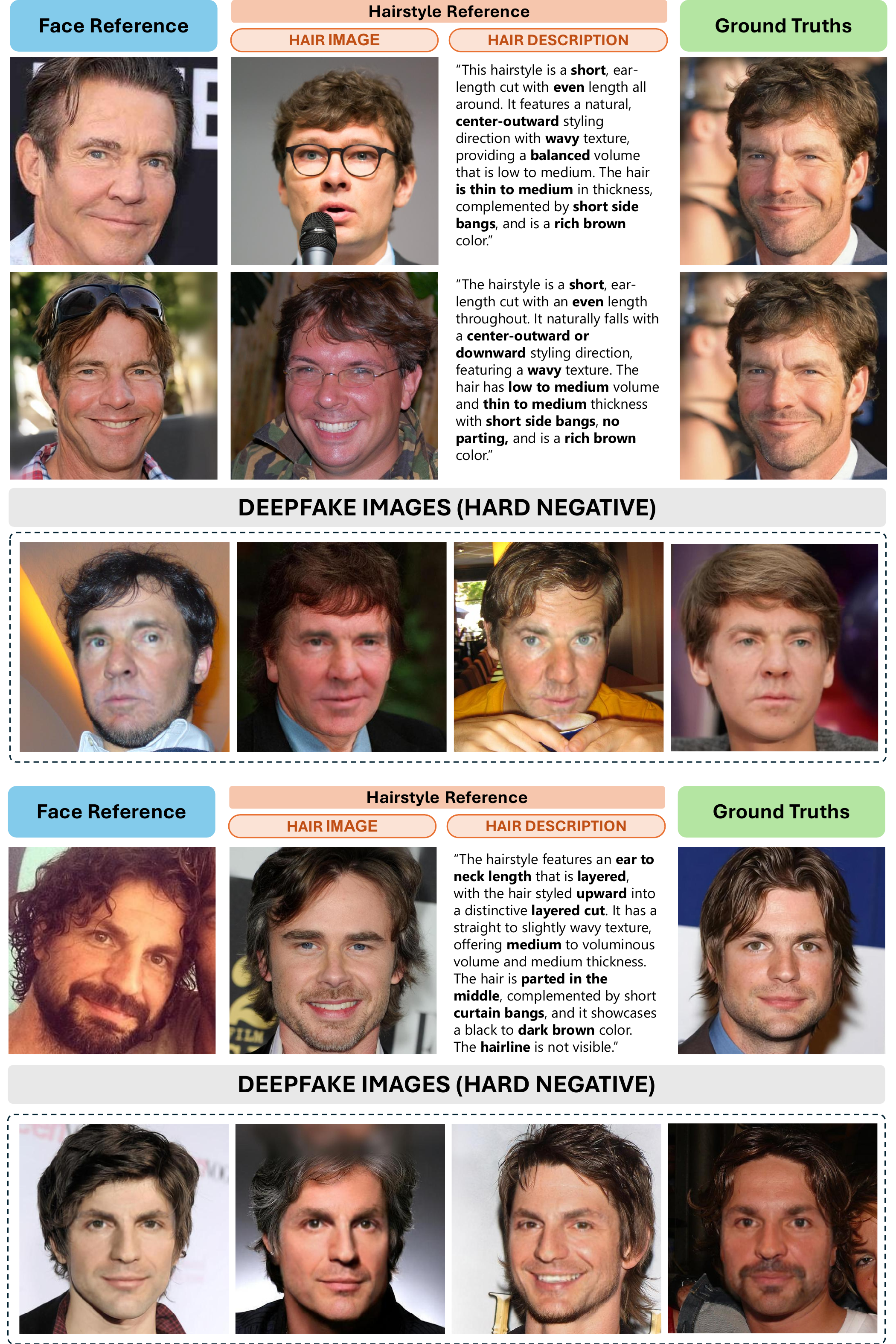}
    \caption{Qualitative examples with deepfake images}
    \label{fig:triplets7}
\end{figure*}

Numerous qualitative triplet examples from DFHR-Bench, covering both Image–-Image and Image-–Text settings with hairstyle cues provided either as images or text descriptions, are shown in Figures~\ref{fig:triplets1}--\ref{fig:triplets7}. Figures~\ref{fig:triplets2}, \ref{fig:triplets5}, and \ref{fig:triplets6} illustrate identities with multiple ground truths, where several visually similar images of the same individual exhibit the same hairstyle. Figures~\ref{fig:triplets6} and \ref{fig:triplets7} further present challenging hard negatives generated through deepfake-based augmentations.

\subsection{Benchmark Comparison and Positioning}
\label{cross-dataset-analysis}
As shown in Table~\ref{tab:comp_benchmarks}, no existing benchmark simultaneously provides identity-awareness and dual-modality query settings, which are essential features for DFHR. Leveraging both Image–-Image and Image–-Text configurations, DFHR-Bench over 10K curated triplets evaluated against a retrieval pool of more than 100K images. This scale is sufficient for meaningful evaluation relative to other contemporaries. To ensure reliable evaluation, we intentionally incorporate hard negatives and curate multiple valid matches per query, preventing the false-negative shortcomings observed in CIR benchmarks. This multi-ground-truth design is aligned with the annotation strategy used by CIRCO \cite{baldrati2023zero}.

\section{Implement Details}
\label{supp:mfhc_implement}


MFHC is trained on \textbf{28{,}906} dual-reference tuples using frozen ArcFace and CLIP ViT-B/32 encoders. Identity and hairstyle features (512-D) are mapped into token spans by two MLP mappers $\phi_{\text{id}}$ and $\phi_{\text{hair}}$, each a two-layer MLP (hidden size 2048) that outputs $n=6$ tokens in the CLIP text-token dimension $d_{\text{text}}$. The CLIP text encoder $\mathrm{CLIP}_{\text{text}}$ is adapted with LoRA (rank~8, $\alpha=32$, dropout~0.1) applied only to its attention and MLP blocks, while the visual encoder remains frozen, making the LoRA adapters and the two token mappers the only trainable components. Training uses the Adam optimizer with a learning rate of $10^{-4}$, batch size 65, cosine annealing over 201 epochs, and a learnable logit scale $\tau$ initialized to 0.07. All experiments use fixed semantic-consistency weights $\lambda_{\text{id}}=1.0$, $\lambda_{\text{hair}}=0.7$, $\lambda_{\text{comp}}=0.7$, $\lambda_{\text{drc}}=0.7$, and hard-negative settings with margin $m=0.1$ and uniform view weights $\alpha_k=1$ for all $k\in\{1,2,3\}$. All MFHC ablations use this same configuration without hyperparameter tuning.

\section{Method Details}
\label{supp:method}

\noindent\textbf{Overall Objectives.}
The Multimodal Face–Hair Combiner (MFHC) is trained end-to-end using a composite objective function that jointly optimizes for cross-modal alignment, semantic factor consistency, and hard-negative discriminability. The total loss $\mathcal{L}_{\text{total}}$ is defined as the weighted sum of the three specific objectives introduced in the main paper:
\begin{equation}
    \mathcal{L}_{\text{total}} = \mathcal{L}_{\text{align}} + \lambda_{\text{sem}}\mathcal{L}_{\text{sem}} + \lambda_{\text{neg}}\mathcal{L}_{\text{hardneg}}.
\end{equation}
Here, $\mathcal{L}_{\text{align}}$ ensures the learned text tokens align with the visual features of the target identity. $\mathcal{L}_{\text{sem}}$ regularizes the token injection process to prevent semantic drift between the disentangled tokens and the full sentence composition. $\mathcal{L}_{\text{hardneg}}$ forces the model to distinguish between valid and invalid identity-hairstyle combinations in the shared embedding space. In our standard experimental setting, we balance these objectives equally, setting $\lambda_{\text{sem}} = 1.0$ and $\lambda_{\text{neg}} = 1.0$.

\section{Additional Analysis}
\label{supp:extended_exp}

\begin{figure*}[t]
  \centering
  \begin{subfigure}[t]{0.49\linewidth}
    \includegraphics[width=\linewidth]{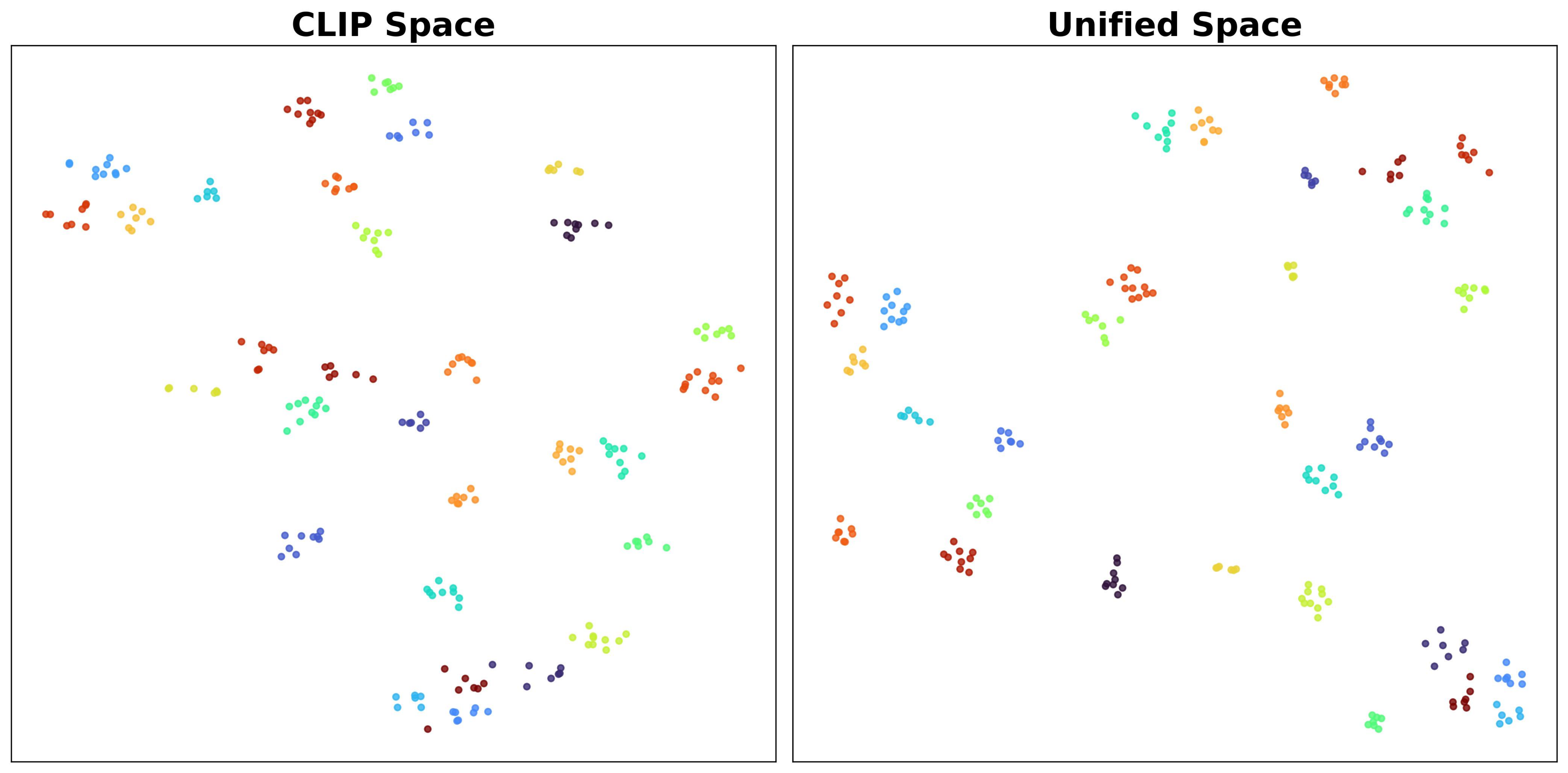}
    \caption{\textit{Identity embedding clusters with no hairstyle variation}. Both CLIP and Unified spaces produce tight, well-separated clusters when each identity appears with a single hairstyle.}
    \label{fig:tsne_gts_only}
  \end{subfigure}\hfill
  \begin{subfigure}[t]{0.49\linewidth}
    \includegraphics[width=\linewidth]{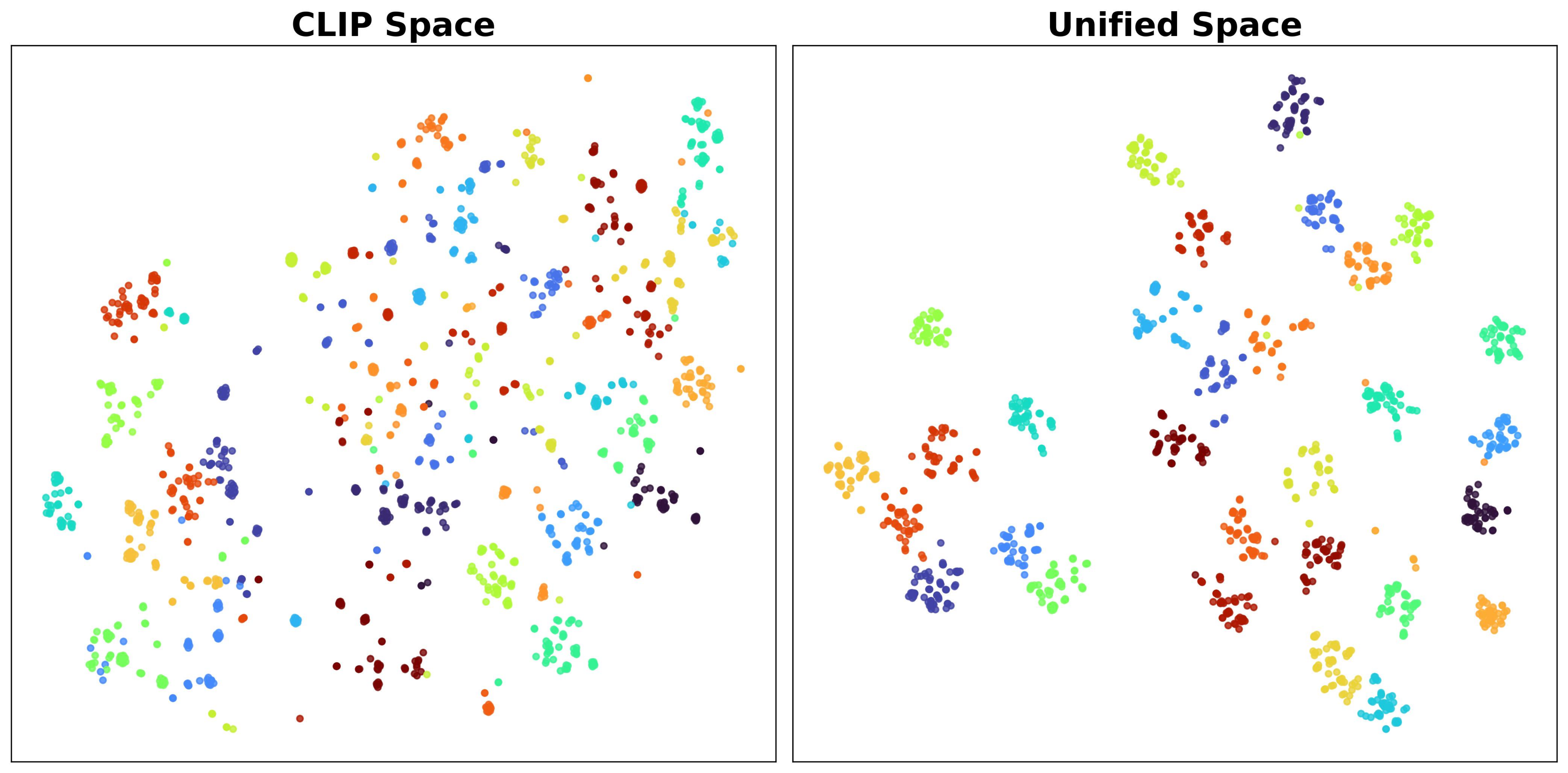}
    \caption{\textit{Identity clusters under hairstyle variation}. CLIP’s image space becomes dispersed and entangled as hairstyles vary, while the Unified space preserves compact, well-separated identity clusters.}
    \label{fig:tsne_under_noise}
  \end{subfigure}
  \caption{}
  \label{fig:tsne-identity-analysis}
\end{figure*}

\subsection{Identity Clustering Under Hairstyle Variation.}
Figure~\ref{fig:tsne-identity-analysis} visualizes identity embeddings in 2D space using t-SNE. We compare CLIP’s original image embedding space (left plots) against our proposed Unified embedding space (right plots) across 30 randomly selected identities.

\textbf{Scenario 1:} No Hairstyle Variation (Figure~\ref{fig:tsne_gts_only}). In this control setting, each identity appears with a single consistent hairstyle. Both embedding spaces yield well-separated, compact clusters. This confirms that when appearance is uniform, CLIP naturally groups identical identities. Crucially, our Unified space maintains this high clustering quality, aligning closely with the identity groupings observed in the image-only CLIP space.

\textbf{Scenario 2:} Hairstyle Variation (Figure~\ref{fig:tsne_under_noise}). A stark divergence appears when identities exhibit multiple different hairstyles. In CLIP’s image space, identity clusters degrade significantly; embeddings for the same person drift apart based on hair changes, and distinct identities with similar hair occasionally merge. This dispersion confirms that CLIP’s image encoder heavily entangles identity with hairstyle cues. In contrast, the Unified space preserves compact, distinct identity clusters despite the hairstyle noise. This demonstrates our method’s ability to effectively disentangle identity from hairstyle, whereas the baseline CLIP space is inherently biased by hair appearance.

\subsection{Analysis of Fusion-Based Retrieval Performance}
Our quantitative results show that simple fusion strategies (both Early and Late) often surpass sophisticated CIR methods like Pic2Word. This phenomenon is explained by the manifold topology visualized in Figure~\ref{fig:tsne_under_noise}. Deep CIR methods are optimized to map composed queries into the \textit{CLIP image space}. However, this target space is inherently entangled—identity clusters scatter significantly when hairstyles change (the dispersed CLIP embeddings). In contrast, fusion baselines bypass this flawed target space entirely. Late Fusion (score aggregation) and Early Fusion (feature concatenation) operate directly on the robust source priors—using ArcFace for identity and CLIP for hair structure. By avoiding the projection into the entangled CLIP image manifold, these methods retain the original discriminative power of the backbones, thereby outperforming CIR methods that "unlearn" identity details in their attempt to align with a noisy visual space.

\subsection{Identity Token Span}
Similar to the hairstyle token analysis, we investigate the sensitivity of the model to the length of the identity token sequence. Table~\ref{tab:id_token_ablation} presents the performance using 4, 6, and 8 identity tokens, while keeping the hairstyle span fixed at the optimal value of 6.

We observe that an identity span of 6 tokens yields the most robust performance across both Image--Image and Image--Text retrieval tasks. Specifically, in the Unified Space, increasing the span from 4 to 6 results in a noticeable improvement, suggesting that 4 tokens may be insufficient to capture the high-frequency details necessary for fine-grained identity preservation. Conversely, extending the span to 8 tokens leads to diminishing returns or slight degradation, indicating that an excessively long sequence may introduce redundancy or optimization difficulties. Furthermore, across all token variations, our Unified Space consistently outperforms the CLIP image space by a significant margin, confirming that the effectiveness of our method stems from the embedding space alignment rather than specific hyperparameter tuning. Based on these results, we adopt 6 identity tokens as the default setting for all main experiments.

\begin{figure*}[t]
    \centering
    \includegraphics[width=0.9\linewidth]{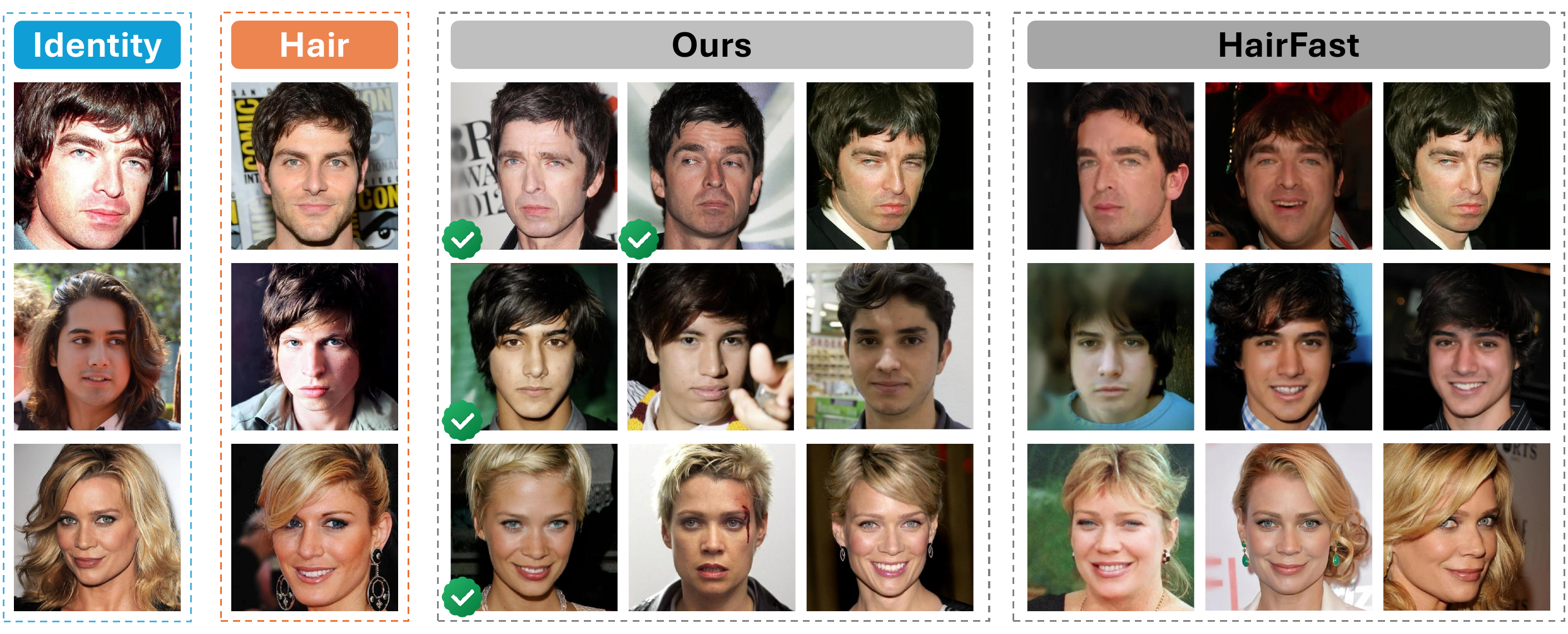}
    \caption{Qualitative comparison between our method (MFHC) and Generation-based method (HairFast) on samples from DFHR-Bench.
    }
    \label{fig:gen-comparison}
\end{figure*}
\begin{figure*}[t]
    \centering
    \includegraphics[width=0.9\linewidth]{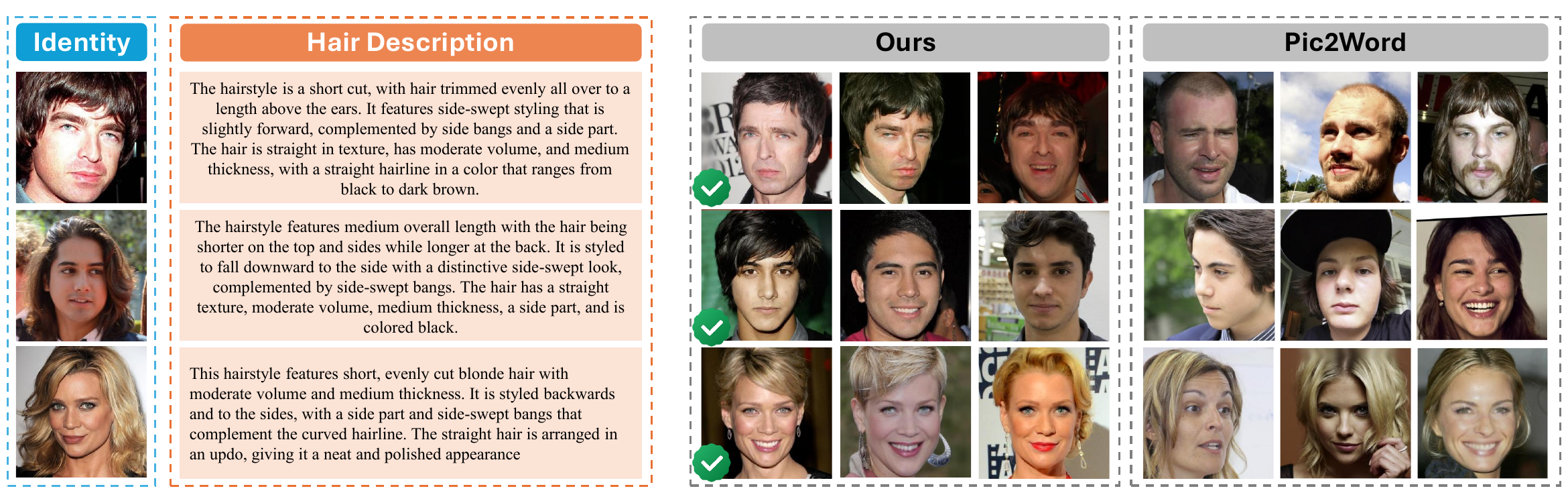}
    \caption{Qualitative comparison between our method (MFHC) and CIR-based method (Pic2Word) on samples from DFHR-Bench.
    }
    \label{fig:cir-comparison}
\end{figure*}

\subsection{Qualitative comparison}
\label{supp:qualitative-comparison}
\begin{table*}[t]
\centering
\caption{Identity-token ablation with fixed hairstyle span of 6. 
We report Recall and Precision at top-$K$ under CLIP image-space indexing 
and unified text-space indexing for both Image--Image and Image--Text DFHR.}
\label{tab:id_token_ablation}
\setlength{\tabcolsep}{4pt}
\renewcommand{\arraystretch}{0.95}
\resizebox{\textwidth}{!}{%

\begin{tabular}{l c
                cc cc
                cc cc
                cc cc
                cc cc}
\toprule
& &
\multicolumn{4}{c}{\textbf{CLIP Space}} &
\multicolumn{4}{c}{\textbf{Unified Space}} \\

\cmidrule(lr){3-6}
\cmidrule(lr){7-10}

\#Identity Tokens & @K &
\multicolumn{2}{c}{Image--Image} &
\multicolumn{2}{c}{Image--Text} &
\multicolumn{2}{c}{Image--Image} &
\multicolumn{2}{c}{Image--Text} \\

& & Recall & Precision & Recall & Precision & Recall & Precision & Recall & Precision \\
\midrule

\multirow{5}{*}{4}
& 1 & 15.45 & 15.45 & 8.76 & 8.76 & 24.74 & 24.74 & 11.88 & 11.88 \\
& 2 & 21.91 & 13.16 & 12.74 & 7.23 & 35.62 & 21.91 & 19.78 & 11.73 \\
& 3 & 26.32 & 11.64 & 16.26 & 6.75 & 44.05 & 19.83 & 25.41 & 11.02 \\
& 4 & 30.81 & 10.80 & 18.76 & 6.29 & 49.41 & 17.95 & 29.71 & 10.40 \\
& 5 & 32.94 & 9.71 & 21.03 & 6.00 & 53.90 & 16.64 & 33.31 & 9.88 \\

\midrule

\multirow{5}{*}{6}
& 1 & 17.49 & 17.49 & 9.23 & 9.23 & 24.82 & 24.82 & 14.70 & 14.70 \\
& 2 & 24.67 & 14.46 & 13.53 & 8.13 & 36.01 & 21.28 & 21.97 & 13.41 \\
& 3 & 29.71 & 12.79 & 17.12 & 7.58 & 42.63 & 18.70 & 26.51 & 12.12 \\
& 4 & 33.10 & 11.41 & 19.23 & 6.90 & 47.12 & 16.75 & 31.12 & 11.30 \\
& 5 & 35.62 & 10.15 & 21.97 & 6.60 & 50.99 & 15.29 & 35.26 & 10.70 \\

\midrule

\multirow{5}{*}{8}
& 1 & 17.73 & 17.73 & 9.93 & 9.93 & 22.70 & 22.70 & 13.21 & 13.21 \\
& 2 & 24.98 & 14.62 & 15.79 & 8.80 & 33.96 & 20.06 & 19.16 & 11.30 \\
& 3 & 29.47 & 12.69 & 18.84 & 7.69 & 41.69 & 18.47 & 25.41 & 11.10 \\
& 4 & 32.94 & 11.25 & 21.97 & 7.27 & 46.26 & 16.00 & 29.32 & 10.32 \\
& 5 & 35.46 & 10.12 & 23.69 & 6.61 & 50.35 & 14.64 & 32.37 & 9.73 \\

\bottomrule
\end{tabular}
}
\end{table*}

\begin{figure*}
    \centering
    \includegraphics[width=0.88\linewidth]{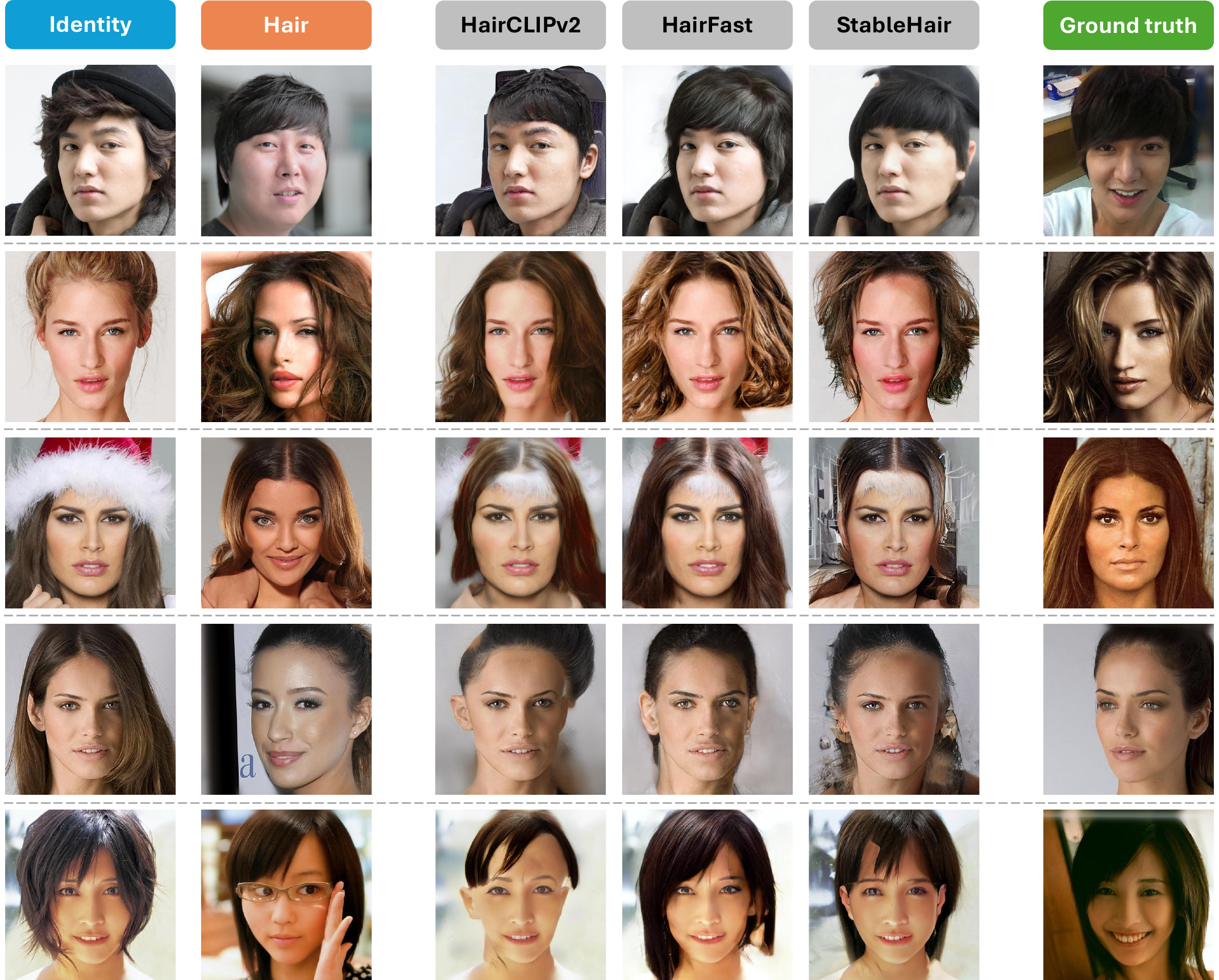}
    \caption{Generative hairstyle-transfer models (HairCLIPv2, HairFast, StableHair) often introduce artifacts such as identity distortion and unrealistic hair textures when combining identity and hairstyle cues. Ground-truth examples preserve clean, natural identity–hairstyle pairs, highlighting the difficulty of synthesis-based baselines for DFHR.}
    \label{fig:hairstyle-transfer}
\end{figure*}

\begin{figure*}
    \centering
    \includegraphics[width=0.9\linewidth]{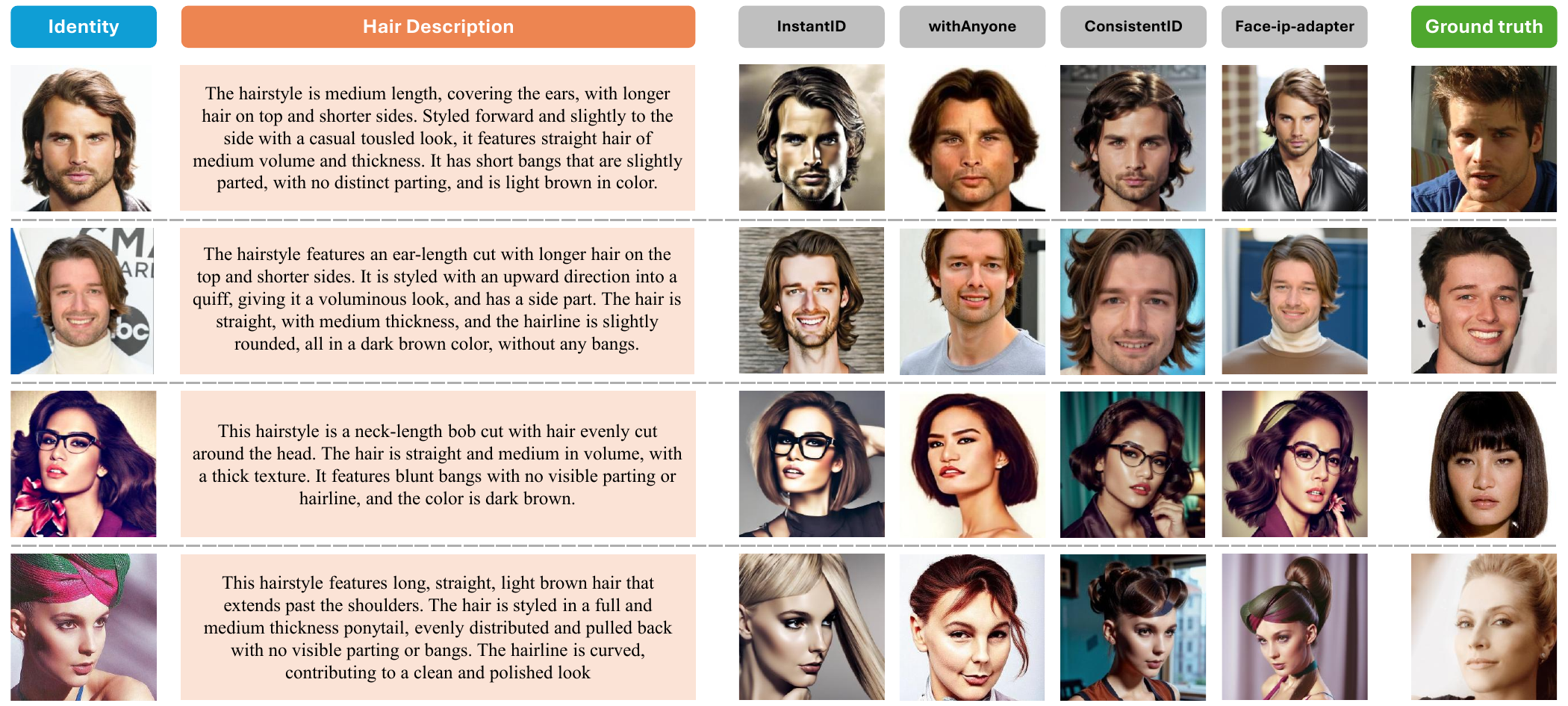}
    \caption{Given an identity image and a text hairstyle description, identity-preserving generators (InstantID, withAnyone, ConsistentID, Face-IP-Adapter) often produce identity drift and hairstyle inaccuracies, whereas ground-truth images retain both factors naturally.}
    \label{fig:id_consistent_gen}
\end{figure*}

\noindent\textbf{Comparison with Generation-based Methods.} Figure~\ref{fig:gen-comparison} demonstrates that our MFHC outperforms the generative-based method (HairFast~\cite{hairfast}), which frequently produces suboptimal results. Figures~\ref{fig:hairstyle-transfer} and \ref{fig:id_consistent_gen} further illustrate that hairstyle transfer and ID-consistent generation methods struggle to synthesize images that faithfully preserve identity while accurately transferring the target hairstyle. While these methods can produce visually well-blended outputs, their results frequently contain distorted facial structures or exaggerated hairstyle attributes. These artifacts propagate into downstream generation-based retrieval, ultimately leading to unstable and unreliable rankings.

\noindent\textbf{Comparison with CIR-based Methods.} Figure~\ref{fig:cir-comparison} presents a qualitative comparison with Pic2Word~\cite{pic2word}, a representative CIR method. As shown in the visual results, standard CIR approaches exhibit severe \textit{identity drift}, frequently failing to retrieve the correct target. This failure stems from the fundamental design of current CIR methods, which map the composed query into the CLIP image space. As detailed in our feature space analysis (Figure~\ref{fig:tsne-identity-analysis}), the CLIP space inherently entangles identity and hairstyle. Consequently, when the baseline integrates the textual modifier, the semantic signals of the hairstyle overpower the visual identity features, treating the face merely as a malleable style attribute rather than a rigid constraint. In contrast, our MFHC framework projects features into a unified text space with explicit disentanglement constraints, successfully retrieving the target hairstyle while retaining high identity fidelity.

\noindent\textbf{Full Quantitative Results.}
Full results regarding the ablation of variants and hairstyle token spans are shown in Table~\ref{tab:full_variants} and Table~\ref{tab:full_hair}. Furthermore, the full results for the main evaluation can be found in Table~\ref{tab:full_imgimg} and Table~\ref{tab:full_imgtxt}.
\begin{table*}[t]
\centering
\caption{Full Results from variant ablation, Recall and Precision at top-$K$ for variants under CLIP image-space indexing and unified text-space indexing.}
\label{tab:full_variants}
\setlength{\tabcolsep}{4pt}
\renewcommand{\arraystretch}{0.92}
\resizebox{\textwidth}{!}{%

\begin{tabular}{l c
                cc cc
                cc cc
                cc cc
                cc cc}
\toprule
& &
\multicolumn{4}{c}{\textbf{CLIP Space}} &
\multicolumn{4}{c}{\textbf{Unified Space}} \\
\cmidrule(lr){3-6}
\cmidrule(lr){7-10}

Variants & @K &
\multicolumn{2}{c}{Image--Image} &
\multicolumn{2}{c}{Image--Text} &
\multicolumn{2}{c}{Image--Image} &
\multicolumn{2}{c}{Image--Text} \\

& & Recall & Precision & Recall & Precision & Recall & Precision & Recall & Precision \\
\midrule

\multirow{5}{*}{A0}
& 1 & 4.57 & 4.57 & 2.27 & 2.27 & 16.71 & 16.71 & 8.99 & 8.99 \\
& 2 & 8.51 & 4.85 & 3.21 & 1.76 & 26.79 & 16.31 & 12.51 & 7.47 \\
& 3 & 9.93 & 4.15 & 3.75 & 1.51 & 33.10 & 14.87 & 15.32 & 6.75 \\
& 4 & 12.14 & 4.02 & 4.53 & 1.41 & 39.56 & 14.42 & 18.06 & 6.33 \\
& 5 & 14.03 & 3.77 & 5.55 & 1.41 & 44.68 & 13.77 & 19.94 & 5.82 \\

\midrule

\multirow{5}{*}{A1}
& 1 & 5.52 & 5.52 & 0.94 & 0.94 & 18.52 & 18.52 & 1.95 & 1.95 \\
& 2 & 9.14 & 5.00 & 1.49 & 0.78 & 26.40 & 15.09 & 3.52 & 2.03 \\
& 3 & 11.98 & 4.70 & 2.66 & 0.96 & 32.55 & 13.29 & 4.61 & 1.90 \\
& 4 & 14.03 & 4.22 & 3.05 & 0.84 & 37.43 & 11.98 & 5.71 & 1.86 \\
& 5 & 15.84 & 3.89 & 3.52 & 0.80 & 40.74 & 10.94 & 6.65 & 1.72 \\

\midrule

\multirow{5}{*}{A2}
& 1 & 11.27 & 11.27 & 5.32 & 5.32 & 13.63 & 13.63 & 6.96 & 6.96 \\
& 2 & 16.55 & 9.54 & 8.68 & 4.93 & 19.23 & 10.17 & 10.40 & 5.75 \\
& 3 & 20.09 & 8.41 & 10.79 & 4.46 & 22.77 & 8.43 & 13.21 & 5.29 \\
& 4 & 22.38 & 7.37 & 13.14 & 4.28 & 25.77 & 7.62 & 16.26 & 5.26 \\
& 5 & 24.35 & 6.67 & 14.46 & 4.07 & 27.82 & 6.79 & 18.84 & 5.00 \\

\midrule

\multirow{5}{*}{A3}
& 1 & 9.22 & 9.22 & 4.85 & 4.85 & 20.41 & 20.41 & 14.93 & 14.93 \\
& 2 & 13.79 & 8.23 & 9.15 & 4.96 & 29.55 & 17.81 & 23.06 & 13.96 \\
& 3 & 16.71 & 7.14 & 11.42 & 4.53 & 35.78 & 15.81 & 28.85 & 12.87 \\
& 4 & 18.99 & 6.32 & 12.82 & 4.03 & 41.06 & 14.07 & 32.92 & 11.67 \\
& 5 & 21.43 & 5.97 & 13.84 & 3.63 & 44.52 & 12.97 & 35.89 & 10.96 \\

\midrule

\multirow{5}{*}{A4}
& 1 & 17.49 & 17.49 & 9.23 & 9.23 & 24.82 & 24.82 & 14.70 & 14.70 \\
& 2 & 24.67 & 14.46 & 13.53 & 8.13 & 36.01 & 21.28 & 21.97 & 13.41 \\
& 3 & 29.71 & 12.79 & 17.12 & 7.58 & 42.63 & 18.70 & 26.51 & 12.12 \\
& 4 & 33.10 & 11.41 & 19.23 & 6.90 & 47.12 & 16.75 & 31.12 & 11.30 \\
& 5 & 35.62 & 10.15 & 21.97 & 6.60 & 50.99 & 15.29 & 35.26 & 10.70 \\

\bottomrule
\end{tabular}
}
\end{table*}

\begin{table*}[t]
\centering
\caption{Full Results of Token-span ablation for MFHC with 6 identity tokens and varying hairstyle token spans. We report Recall and Precision at top-$K$ under CLIP image-space indexing and unified text-space indexing for both Image--Image and Image--Text DFHR.}
\label{tab:full_hair}
\setlength{\tabcolsep}{4pt}
\renewcommand{\arraystretch}{0.92}
\resizebox{\textwidth}{!}{%

\begin{tabular}{l c
                cc cc
                cc cc
                cc cc
                cc cc}
\toprule
& &
\multicolumn{4}{c}{\textbf{CLIP Space}} &
\multicolumn{4}{c}{\textbf{Unified Space}} \\
\cmidrule(lr){3-6}
\cmidrule(lr){7-10}

\#Hairstyle Tokens & @K &
\multicolumn{2}{c}{Image--Image} &
\multicolumn{2}{c}{Image--Text} &
\multicolumn{2}{c}{Image--Image} &
\multicolumn{2}{c}{Image--Text} \\

& & Recall & Precision & Recall & Precision & Recall & Precision & Recall & Precision \\
\midrule

\multirow{5}{*}{2}
& 1 & 17.18 & 17.18 & 7.04 & 7.04 & 24.35 & 24.35 & 11.42 & 11.42 \\
& 2 & 25.06 & 14.46 & 10.56 & 6.14 & 34.59 & 21.16 & 17.12 & 10.13 \\
& 3 & 29.24 & 12.56 & 12.90 & 5.50 & 40.90 & 18.44 & 21.97 & 9.62 \\
& 4 & 33.49 & 11.43 & 15.40 & 5.26 & 46.02 & 16.63 & 25.57 & 9.01 \\
& 5 & 36.88 & 10.48 & 17.59 & 5.07 & 50.04 & 15.35 & 28.54 & 8.48 \\

\midrule

\multirow{5}{*}{4}
& 1 & 17.42 & 17.42 & 7.90 & 7.90 & 23.17 & 23.17 & 10.95 & 10.95 \\
& 2 & 24.03 & 14.11 & 12.28 & 7.00 & 33.65 & 20.61 & 18.53 & 10.83 \\
& 3 & 28.68 & 12.21 & 15.56 & 6.33 & 41.13 & 18.54 & 24.39 & 10.16 \\
& 4 & 32.94 & 11.39 & 19.00 & 6.14 & 45.31 & 16.67 & 27.52 & 9.25 \\
& 5 & 36.96 & 10.64 & 21.66 & 5.90 & 50.28 & 15.48 & 30.10 & 8.48 \\

\midrule

\multirow{5}{*}{6}
& 1 & 17.49 & 17.49 & 9.23 & 9.23 & 24.82 & 24.82 & 14.70 & 14.70 \\
& 2 & 24.67 & 14.46 & 13.53 & 8.13 & 36.01 & 21.28 & 21.97 & 13.41 \\
& 3 & 29.71 & 12.79 & 17.12 & 7.58 & 42.63 & 18.70 & 26.51 & 12.12 \\
& 4 & 33.10 & 11.41 & 19.23 & 6.90 & 47.12 & 16.75 & 31.12 & 11.30 \\
& 5 & 35.62 & 10.15 & 21.97 & 6.60 & 50.99 & 15.29 & 35.26 & 10.70 \\

\midrule

\multirow{5}{*}{8}
& 1 & 17.34 & 17.34 & 9.93 & 9.93 & 22.54 & 22.54 & 12.35 & 12.35 \\
& 2 & 25.37 & 14.46 & 14.15 & 8.33 & 30.81 & 18.40 & 20.17 & 11.81 \\
& 3 & 29.31 & 12.29 & 17.04 & 7.53 & 38.69 & 16.92 & 26.27 & 11.36 \\
& 4 & 33.25 & 11.29 & 19.39 & 6.82 & 43.50 & 15.43 & 30.10 & 10.61 \\
& 5 & 35.93 & 10.23 & 21.50 & 6.40 & 47.91 & 14.12 & 33.62 & 10.01 \\

\midrule

\multirow{5}{*}{10}
& 1 & 16.78 & 16.78 & 10.32 & 10.32 & 25.14 & 25.14 & 12.04 & 12.04 \\
& 2 & 24.27 & 14.22 & 15.01 & 8.52 & 36.88 & 22.22 & 18.61 & 10.95 \\
& 3 & 29.16 & 12.58 & 17.67 & 7.30 & 42.55 & 18.73 & 24.47 & 10.63 \\
& 4 & 32.23 & 11.17 & 20.25 & 6.76 & 47.20 & 17.12 & 27.52 & 9.62 \\
& 5 & 35.07 & 10.07 & 22.20 & 6.08 & 51.14 & 15.62 & 30.96 & 9.01 \\

\bottomrule
\end{tabular}
}
\end{table*}

\section{Adaptation of Existing Methods}
\label{supp:baselines}

To establish a comprehensive comparison for the Dual Face–Hair Retrieval task, we explore three design paradigms: (i) Fusion-based Retrieval, (ii) Generation-based Retrieval, and (iii) Composed Image Retrieval. Each method reflects a different perspective on how to combine identity and hairstyle cues for effective cross-image matching.

\subsection{Fusion-Based Methods}
\label{sec:fusion-baseline}

To establish strong discriminative methods for DFHR, we evaluate fusion strategies that combine identity and hairstyle signals without compositional learning. All baselines share the same pair of frozen pretrained encoders:

\noindent\textbf{Identity encoder} $f_{\text{id}}$: ArcFace~\cite{arcface}, producing $v_{\text{id}}(x)=f_{\text{id}}(x)\in\mathbb{R}^{512}$ from an aligned face crop.

\noindent\textbf{Hairstyle encoder} $f_{\text{hair}}$: CLIP~\cite{clip}, yielding $v_{\text{hair}}(x)=f_{\text{hair}}(x)\in\mathbb{R}^{512}$ from the hair crop. Gallery entries always use the CLIP \emph{image} encoder, while queries use either the image encoder (Image--Image) or the text encoder (Image--Text), depending on the form of $q_{\text{hair}}$.

\subsubsection{Early Fusion Baseline}

Early Fusion constructs a joint representation by concatenating identity and hairstyle features:
\[
g_{\text{EF}}(x_i)
= \mathrm{Norm}\!\left(v_{\text{id}}(x_i)\,\|\,v_{\text{hair}}(x_i)\right)
\in\mathbb{R}^{1024}.
\]

\noindent\textbf{Query Embedding.}
A query $(q_{\text{id}},q_{\text{hair}})$ is mapped to  
$f_{\text{EF}}(q)=\mathrm{Norm}\!\left(v_{\text{id}}(q_{\text{id}})\,\|\,v_{\text{hair}}(q_{\text{hair}})\right)$,  
and gallery candidates $x_i$ are ranked via
$\mathrm{sim}\!\big(f_{\text{EF}}(q), g_{\text{EF}}(x_i)\big)$.

\noindent\textbf{Discussion.}
Early Fusion provides a direct mixed-cue baseline but merely juxtaposes ArcFace and CLIP features.  
It does not align the two embedding spaces, model cross-factor interactions, nor resolve modality gaps, resulting in limited disentanglement.

\subsubsection{Late Fusion Baseline}
Late Fusion maintains fully factorized identity and hairstyle spaces.  
For a query $(q_{\text{id}},q_{\text{hair}})$ and gallery image $x_i$, we compute
$
\text{sim}_{\text{id}}
= \mathrm{sim}\big(v_{\text{id}}(q_{\text{id}}), v_{\text{id}}(x_i)\big),
$ and
$ \text{sim}_{\text{hair}}
= \mathrm{sim}\big(v_{\text{hair}}(q_{\text{hair}}), v_{\text{hair}}(x_i)\big).$ 
The final score is a weighted combination:
\[
\text{sim}_{\text{LF}}(q,x_i)
= \alpha\,\text{sim}_{\text{id}}(q,x_i)
+ (1-\alpha)\,\text{sim}_{\text{hair}}(q,x_i),
\]
where $\alpha\in[0,1]$ controls the identity–hairstyle balance. In our experiments, we set $\alpha=0.5$.
\noindent\textbf{Discussion.}
Late Fusion avoids cross-space interference and enables explicit factor weighting, but it cannot capture identity–hairstyle interactions and remains vulnerable to cross-modal inconsistencies between CLIP image and text spaces.

\subsubsection{Two-stage Pipeline}
We further implement an explicit 2-stage baseline that first retrieves Top-$K$ candidates using a face recognition model and then re-ranks them by hairstyle similarity. As shown in Table~\ref{tab:twostage}, the resulting performance is consistent with Early/Late Fusion trends reported in the main paper and degrades rapidly as $K$ increases. This consistency confirms that treating identity and hairstyle as independent constraints is insufficient under the DFHR evaluation protocol, motivating the need for joint, disentangled optimization rather than factorized retrieval strategies.

\begin{table}[t]
\centering
\caption{\textbf{Comparison with 2-stage baselines.}}
\label{tab:twostage}
\resizebox{1\linewidth}{!}{
\setlength{\tabcolsep}{4.5pt}
\renewcommand{\arraystretch}{0.95}
\begin{tabular}{l|ccc|ccc|cc}
\toprule
& \multicolumn{3}{c|}{\textbf{Img + Img}} 
& \multicolumn{3}{c|}{\textbf{Img + Txt}} 
& \multicolumn{2}{c}{\textbf{Average}} \\
\textbf{Method} 
& R@1 & R@5 & mAP 
& R@1 & R@5 & mAP 
& @1 & $\Delta$ \\
\midrule
\textbf{MFHC} 
& \textbf{24.8} & \textbf{51.0} & \textbf{22.6}
& \textbf{14.7} & 35.3 & \textbf{13.3}
& \textbf{19.8} & -- \\

Late Fusion
& 15.1 & 35.2 & 12.6
& 11.4 & 27.7 & 9.8
& 13.2 & \textbf{\textcolor{red!60!black}{-6.6}} \\

2-stage (K=30)
& 18.2 & 45.6 & 16.2
& 12.6 & \textbf{37.4} & 11.4
& 15.4 & \textbf{\textcolor{red!60!black}{-4.4}} \\

2-stage (K=50)
& 15.5 & 39.6 & 13.4
& 8.1 & 28.2 & 7.8
& 11.8 & \textbf{\textcolor{red!60!black}{-8.0}} \\

2-stage (K=100)
& 11.7 & 33.7 & 10.5
& 4.8 & 17.8 & 4.8
& 8.3 & \textbf{\textcolor{red!60!black}{-11.5}} \\
\bottomrule
\end{tabular}
}
\end{table}

\subsection{Generation-Based Baselines}
\label{sec:generation-baseline}

We additionally evaluate DFHR under a generation-based paradigm, where identity-conditioned, hairstyle-guided generative models synthesize a composite image for each query.  
Given $(q_{\text{id}}, q_{\text{hair}})$, a generator produces an image $\tilde{x}$ that is expected to preserve the identity of $q_{\text{id}}$ while adopting the target hairstyle from $q_{\text{hair}}$.  
Retrieval is then performed directly in the image space by embedding $\tilde{x}$ and ranking gallery images by similarity.

\noindent\textbf{Motivation.}
Standard ArcFace is optimized to be invariant to hairstyle, background, and other appearance factors, focusing solely on identity discrimination.  
As a result, cosine similarity between vanilla ArcFace embeddings cannot reliably reflect whether a generated image $\tilde{x}$ correctly matches the desired hairstyle, especially for \emph{intra-identity} hairstyle changes.  
To obtain a fair metric for generation-based DFHR, we make ArcFace \textit{hair-aware within each identity}.

\noindent\textbf{Hair-Aware ArcFace Adapter.}
Rather than fine-tuning the full ArcFace backbone, we freeze the pretrained identity encoder $f_{\text{id}}$ and learn a lightweight adapter on top of its embeddings.  
Let $e(x) = f_{\text{id}}(x) \in \mathbb{R}^{512}$ denote the frozen ArcFace feature.
We introduce a two-layer MLP $h_{\theta}:\mathbb{R}^{512}\rightarrow\mathbb{R}^{512}$ with a nonlinearity:
\[
z(x) = f_{\text{HA}}(x)
= \mathrm{Norm}\big(h_{\theta}(e(x))\big),
\]
where $\mathrm{Norm}(\cdot)$ denotes $\ell_2$ normalization.  
This adapter is trained on precomputed ArcFace embeddings using hairstyle-sensitive triplets.

\noindent\textbf{Intra-Identity Hair Triplets.}
We construct triplets $(x^{\text{anc}}, x^{\text{pos}}, x^{\text{neg}})$ within each identity from Hair Encoder from \S\ref{sec:hair-encoder-implementation}:
the anchor $x^{\text{anc}}$ and positive $x^{\text{pos}}$ share both identity and hairstyle, while the negative $x^{\text{neg}}$ shares the same identity but exhibits a different hairstyle.  
Let $e_{\text{anc}}, e_{\text{pos}}, e_{\text{neg}}$ denote the frozen ArcFace embeddings, and
$z_{\text{anc}}, z_{\text{pos}}, z_{\text{neg}}$ the adapted features.
We employ a cosine triplet loss with a small margin $m$:
\[
\mathcal{L}_{\text{triplet}}
= \max\Big(0,\,
\cos(z_{\text{anc}}, z_{\text{neg}})
-
\cos(z_{\text{anc}}, z_{\text{pos}})
+ m \Big),
\]
which encourages the adapter to pull together same-hairstyle pairs and push away different-hairstyle variants of the \emph{same} identity.

\noindent\textbf{Regularization toward ArcFace.}
To preserve the strong identity structure of ArcFace, we additionally constrain the adapted embedding to stay close to the original one via cosine regularization:
\begin{align}
\mathcal{L}_{\text{reg}}
= \tfrac{1}{3}\Big[
    &\big(1 - \cos(z_{\text{anc}}, e_{\text{anc}})\big)
    + \big(1 - \cos(z_{\text{pos}}, e_{\text{pos}})\big)
    \nonumber \\
    &\quad + \big(1 - \cos(z_{\text{neg}}, e_{\text{neg}})\big)
\Big].
\end{align}

The total training objective is

\[
\mathcal{L}_{\text{HA}}
= \lambda_{\text{triplet}}\mathcal{L}_{\text{triplet}}
+ \lambda_{\text{reg}}\mathcal{L}_{\text{reg}},
\]
where $\lambda_{\text{triplet}}$ and $\lambda_{\text{reg}}$ balance hairstyle sensitivity and identity preservation.

\noindent\textbf{Retrieval with Synthetic Images.}
After training, the hair-aware embedding function $f_{\text{HA}}$ is used for both gallery and synthetic images.  
Given a generated image $\tilde{x}$ for query $(q_{\text{id}}, q_{\text{hair}})$ and a gallery image $x_i$, we compute
$v_{\tilde{x}} = f_{\text{HA}}(\tilde{x})$ and $v_{x_i} = f_{\text{HA}}(x_i)$, and rank candidates according to cosine similarity $\mathrm{sim}(v_{\tilde{x}}, v_{x_i})$.  
This setup allows generation-based baselines to be evaluated in a space that remains identity-discriminative but is explicitly sensitive to intra-identity hairstyle variation.

Because DFHR-Bench contains multiple ground-truth images per query (same identity and hairstyle), we treat each image as a query and require it to retrieve another image with the same identity–hairstyle pair. Under this multi–ground-truth setting, we evaluate the Hair-Aware ArcFace adapter on 1,725 intra-identity hairstyle queries from DFHR-Bench. As shown in Table~\ref{tab:haireval}, the adapter consistently improves over standard ArcFace, which is largely hairstyle-invariant. At Rank-1, the success rate increases from 84.00\% to 87.54\% (+3.54\%), indicating better discrimination of hairstyle variations within the same identity. Smaller but steady gains are observed at higher ranks, and precision metrics follow the same pattern, with the largest improvement again at Rank-1. Overall, these results show that the adapter successfully injects hairstyle sensitivity into ArcFace while preserving its identity structure, providing a more suitable retrieval space for generation-based DFHR evaluation.

\subsection{Composed Image Retrieval Baselines}
\label{sec:composed-baseline}

Composed Image Retrieval (CIR) is a well-established paradigm that learns to modify a visual reference according to a textual description, and has shown strong performance on benchmarks such as CIRR~\cite{liu2021_cirr} and FashionIQ~\cite{wu2020_fashioniq}.
However, these settings do not constrain identity, since the goal is to retrieve images that match a semantic modification rather than preserve the appearance of a specific person. In contrast, DFHR explicitly focuses on \emph{identity-aware} retrieval, where the model must preserve the person’s identity while modifying only the hairstyle.
To fairly assess the applicability and limitations of CIR-style methods in this stricter setting, we reproduce their core idea of \emph{image–text composition} and adapt it to our mixed-modality DFHR protocol.

We therefore build two adapted CIR baselines—ZS-CIR~\cite{ZS-CIR} and Pic2Word~\cite{pic2word}—that keep their original compositional mechanisms as intact as possible while incorporating an explicit identity encoder. Although numerous other CIR architectures exist, we limit our evaluation to these representative methods because, as detailed in \S\ref{supp:extended_exp}, the CLIP image space fundamentally fails to support the strict identity preservation required for DFHR.

\begin{table}[t]
\centering
\small
\setlength{\tabcolsep}{6pt}
\caption{Evaluation of Hair-Aware ArcFace on 1,725 intra-identity hairstyle queries. Each query and target share the same identity and hairstyle.}
\label{tab:haireval}
\begin{tabular}{lcc}
\toprule
\textbf{Metric} & \textbf{ArcFace} & \textbf{Fine-Tuned} \\
\midrule
R@1 & 84.00 & 87.54 \\
R@2 & 94.61 & 95.13 \\
R@3 & 96.23 & 96.52 \\
R@4 & 97.22 & 97.62 \\
R@5 & 97.86 & 98.38 \\
\midrule
P@1 & 84.00 & 87.54 \\
P@2 & 72.72 & 73.51 \\
P@3 & 63.36 & 64.23 \\
P@4 & 56.70 & 56.91 \\
P@5 & 51.15 & 51.52 \\
\bottomrule
\end{tabular}
\end{table}

\begin{table*}[t]
\centering
\renewcommand{\arraystretch}{0.95}
\caption{
\textbf{Full Results of Image--Image Evaluation.} Comparison on the \textbf{(a) Bimodal Alignment Subset} and \textbf{(b) Official Set}. Best and second-best in \textbf{bold} and \underline{underlined}.
}
\label{tab:full_imgimg}
\resizebox{0.95\textwidth}{!}{
\begin{tabular}{@{} l *{5}{c} | *{5}{c} | c @{}}
\toprule
\multirow{2}{*}{Method} & \multicolumn{5}{c|}{Recall@k (\%)} & \multicolumn{5}{c|}{Precision@k (\%)} & \multirow{2}{*}{mAP} \\
\cmidrule(lr){2-6} \cmidrule(lr){7-11}
 & 1 & 2 & 3 & 4 & 5 & 1 & 2 & 3 & 4 & 5 & \\
\midrule
\multicolumn{12}{c}{\cellcolor{gray!10}\textbf{(a) Bimodal Alignment Subset}} \\
\midrule
\multicolumn{12}{@{}l}{\textit{Fusion-Based}} \\
Face-only & 7.51 & 11.81 & 14.07 & 16.50 & 18.76 & 7.51 & 7.31 & 6.59 & 5.96 & 5.69 & 7.13 \\
Hair-only & 0.24 & 0.79 & 0.79 & 1.42 & 2.05 & 0.24 & 0.39 & 0.34 & 0.41 & 0.46 & 0.39 \\
Early Fusion & 9.14 & 15.13 & 20.09 & 27.03 & 34.99 & 9.14 & 8.67 & 8.43 & 8.77 & \underline{9.49} & 10.37 \\
Late Fusion & \underline{15.05} & \underline{23.48} & \underline{29.55} & \underline{33.25} & \underline{35.15} & \underline{15.05} & \underline{12.65} & \underline{11.24} & \underline{10.01} & 8.86 & \underline{12.57} \\
\multicolumn{12}{@{}l}{\textit{Generation-Based}} \\
HairCLIPv2 \cite{hairclipv2} & 4.38 & 7.58 & 10.56 & 14.46 & 20.02 & 4.38 & 4.22 & 4.25 & 4.61 & 5.22 & 5.48 \\
HairFast \cite{hairfast} & 6.33 & 10.09 & 14.00 & 18.22 & 26.19 & 6.33 & 5.79 & 5.79 & 6.06 & 6.94 & 7.49 \\
StableHair \cite{stablehair} & 5.71 & 9.85 & 13.29 & 17.90 & 22.99 & 5.71 & 5.67 & 5.58 & 5.75 & 6.10 & 6.67 \\
\rowcolor{hilite}
\textbf{Ours (MFHC)} & \textbf{24.82} & \textbf{36.01} & \textbf{42.63} & \textbf{47.12} & \textbf{50.99} & \textbf{24.82} & \textbf{21.28} & \textbf{18.70} & \textbf{16.75} & \textbf{15.29} & \textbf{22.63} \\
\midrule
\multicolumn{12}{c}{\cellcolor{gray!10}\textbf{(b) Official Set}} \\
\midrule
\multicolumn{12}{@{}l}{\textit{Fusion-Based}} \\
Face-only & 6.52 & 10.15 & 12.38 & 15.03 & 16.87 & 6.52 & 6.14 & 5.71 & 5.20 & 4.91 & 5.89 \\
Hair-only & 0.43 & 0.72 & 0.98 & 1.31 & 1.42 & 0.43 & 0.36 & 0.35 & 0.35 & 0.30 & 0.32 \\
Early Fusion & 7.91 & 13.72 & 20.03 & 26.96 & \underline{35.51} & 7.91 & 7.69 & 8.07 & 8.50 & \underline{9.39} & 9.33 \\
Late Fusion & \underline{13.46} & \underline{22.17} & \underline{28.96} & \underline{32.21} & 34.52 & \underline{13.46} & \underline{12.32} & \underline{11.48} & \underline{10.14} & 9.05 & \underline{11.90} \\
\multicolumn{12}{@{}l}{\textit{Generation-Based}} \\
HairCLIPv2 \cite{hairclipv2} & 4.02 & 7.22 & 10.26 & 13.53 & 19.26 & 4.02 & 3.97 & 3.98 & 4.10 & 4.85 & 4.88 \\
HairFast \cite{hairfast} & 5.28 & 9.41 & 13.35 & 17.76 & 24.72 & 5.28 & 5.15 & 5.19 & 5.60 & 6.46 & 6.59 \\
StableHair \cite{stablehair} & 4.77 & 8.23 & 11.56 & 15.67 & 21.20 & 4.77 & 4.65 & 4.70 & 5.12 & 5.74 & 5.98 \\
\rowcolor{hilite}
\textbf{Ours} & \textbf{24.72} & \textbf{35.77} & \textbf{42.75} & \textbf{48.25} & \textbf{52.43} & \textbf{24.72} & \textbf{21.05} & \textbf{18.50} & \textbf{16.61} & \textbf{15.17} & \textbf{22.77} \\
\bottomrule
\end{tabular}%
}
\end{table*}

\subsubsection{Adapted ZS-CIR Method.}
To evaluate how zero-shot composed image retrieval paradigm transfers to the identity-aware DFHR setting, we adapt the ZS-CIR formulation to incorporate an explicit identity encoder.  
Following the original design, the method performs \emph{image–text composition} by fusing the visual reference with a hairstyle description in CLIP space.\footnote{We implement ZS-CIR using the official released codebase and default CLIP ViT-B/32 configuration.}  
However, the original CLIP image encoder does not preserve person identity reliably, as it is optimized for semantic rather than identity discrimination.  
To address this mismatch, we replace the CLIP visual encoder with an ArcFace encoder, whose identity-discriminative features ensure that the composed representation retains subject identity while relying on the CLIP text branch to inject hairstyle-modification cues.  
The remainder of the composition mechanism is kept as close as possible to the official implementation to maintain fidelity to the ZS-CIR formulation.

\begin{table*}[t]
\centering
\renewcommand{\arraystretch}{0.95}
\caption{
\textbf{Full Results of Image--Text Evaluation.} Comparison on the \textbf{(a) Bimodal Alignment Subset} and \textbf{(b) Official Set}. Best and second-best in \textbf{bold} and \underline{underlined}.
}
\label{tab:full_imgtxt}
\resizebox{0.95\textwidth}{!}{%
\begin{tabular}{@{} l *{5}{c} | *{5}{c} | c @{}}
\toprule
\multirow{2}{*}{Method} & \multicolumn{5}{c|}{Recall@k (\%)} & \multicolumn{5}{c|}{Precision@k (\%)} & \multirow{2}{*}{mAP} \\
\cmidrule(lr){2-6} \cmidrule(lr){7-11}
 & 1 & 2 & 3 & 4 & 5 & 1 & 2 & 3 & 4 & 5 & \\
\midrule
\multicolumn{12}{c}{\cellcolor{gray!10}\textbf{(a) Bimodal Alignment Subset}} \\
\midrule
\multicolumn{12}{@{}l}{\textit{Fusion-Based}} \\
Face-only & 7.51 & 11.81 & 14.07 & 16.50 & 18.76 & 7.51 & 7.31 & 6.59 & 5.96 & 5.69 & 7.13 \\
Hair-only & 0.00 & 0.00 & 0.00 & 0.00 & 0.00 & 0.00 & 0.00 & 0.00 & 0.00 & 0.00 & 0.00 \\
Early Fusion & 7.97 & 13.37 & 16.89 & 20.72 & 27.29 & 7.97 & 7.90 & 7.22 & 7.02 & 7.63 & 8.42 \\
Late Fusion & \underline{11.42} & \underline{17.04} & \underline{21.42} & \underline{24.94} & 27.68 & \underline{11.42} & 9.93 & 8.99 & 8.21 & 7.66 & 9.82 \\
\multicolumn{12}{@{}l}{\textit{Generation-Based}} \\
InstantID \cite{instantid} & 7.58 & 11.73 & 15.64 & 19.70 & 24.86 & 7.58 & 7.04 & 6.93 & 6.86 & 7.04 & 8.43 \\
withAnyone \cite{withanyone} & 3.44 & 6.41 & 9.30 & 12.43 & 15.09 & 3.44 & 3.60 & 3.83 & 4.09 & 4.19 & 4.49 \\
ConsistentID \cite{consistentid} & 9.23 & 15.64 & 19.94 & 23.85 & \underline{28.07} & 9.23 & 9.11 & 8.68 & 8.37 & 8.19 & 10.36 \\
Face-IP-Adapter \cite{han2024face} & 10.63 & 16.34 & 20.17 & 24.00 & 26.82 & 10.63 & \underline{10.32} & \underline{9.51} & \underline{9.23} & \underline{8.71} & \underline{10.76} \\
\multicolumn{12}{@{}l}{\textit{Composed Image Retrieval}} \\
ZS-CIR \cite{ZS-CIR} & 2.42 & 3.91 & 6.18 & 7.11 & 8.37 & 2.42 & 2.19 & 2.32 & 2.17 & 2.14 & 2.19 \\
Pic2Word \cite{pic2word} & 2.81 & 3.99 & 4.30 & 5.08 & 5.79 & 2.81 & 2.27 & 1.85 & 1.76 & 1.64 & 1.87 \\
\rowcolor{hilite}
\textbf{Ours (MFHC)} & \textbf{14.70} & \textbf{21.97} & \textbf{26.51} & \textbf{31.12} & \textbf{35.26} & \textbf{14.70} & \textbf{13.41} & \textbf{12.09} & \textbf{11.30} & \textbf{10.70} & \textbf{13.31} \\
\midrule
\multicolumn{12}{c}{\cellcolor{gray!10}\textbf{(b) Official Set}} \\
\midrule
\multicolumn{12}{@{}l}{\textit{Fusion-Based}} \\
Face-only & 7.76 & 12.23 & 14.73 & 17.49 & 20.01 & 7.76 & 7.39 & 6.87 & 6.32 & 6.14 & 7.23 \\
Hair-only & 0.00 & 0.02 & 0.07 & 0.07 & 0.07 & 0.00 & 0.01 & 0.02 & 0.02 & 0.02 & 0.01 \\
Early Fusion & 8.58 & 13.98 & 18.40 & 22.92 & 29.26 & 8.58 & 8.18 & 7.86 & 7.81 & 8.28 & 8.74 \\
Late Fusion & \underline{12.37} & \underline{18.40} & \underline{22.56} & 26.05 & 28.59 & \underline{12.37} & \underline{10.56} & 9.57 & 8.71 & 8.09 & 10.26 \\
\multicolumn{12}{@{}l}{\textit{Generation-Based}} \\
InstantID \cite{instantid} & 6.17 & 11.44 & 15.02 & 18.81 & 23.71 & 6.17 & 6.75 & 6.46 & 6.49 & 6.69 & 7.51 \\
withAnyone \cite{withanyone} & 3.72 & 6.80 & 9.15 & 12.21 & 16.24 & 3.72 & 4.00 & 3.91 & 4.07 & 4.50 & 4.64 \\
ConsistentID \cite{consistentid} & 10.19 & 17.23 & 22.03 & \underline{26.12} & \underline{30.54} & 10.19 & 9.98 & 9.51 & 9.08 & 8.89 & 10.99 \\
Face-IP-Adapter \cite{han2024face} & 11.46 & 17.13 & 20.95 & 24.99 & 28.28 & 11.46 & 10.48 & \underline{9.79} & \underline{9.53} & \underline{9.14} & \underline{11.34} \\
\multicolumn{12}{@{}l}{\textit{Composed Image Retrieval}} \\
ZS-CIR \cite{ZS-CIR} & 2.79 & 4.95 & 6.63 & 7.90 & 8.82 & 2.79 & 2.68 & 2.61 & 2.46 & 2.29 & 2.55 \\
Pic2Word \cite{pic2word} & 2.96 & 4.23 & 5.24 & 6.03 & 6.61 & 2.96 & 2.37 & 2.07 & 1.88 & 1.73 & 1.93 \\
\rowcolor{hilite}
\textbf{Ours} & \textbf{14.70} & \textbf{21.89} & \textbf{26.58} & \textbf{31.20} & \textbf{35.18} & \textbf{14.70} & \textbf{13.29} & \textbf{12.09} & \textbf{11.32} & \textbf{10.70} & \textbf{13.29} \\
\bottomrule
\end{tabular}%
}
\end{table*}

\subsubsection{Adapted Pic2Word Method.}
Pic2Word~\cite{pic2word} learns a \emph{pseudo-word} that translates a reference image into a CLIP text-token embedding, enabling compositional reasoning by inserting this learned token into a natural-language prompt.  
To align this mechanism with DFHR, we reinterpret the pseudo-word as an \emph{identity token} rather than an object-level appearance descriptor.

\noindent\textbf{Identity-as-Token.}
In its original design, Pic2Word maps a CLIP image feature into a token that is appended to a template prompt such as \texttt{a photo of [*]}; a contrastive loss then encourages the resulting text embedding to reconstruct the visual representation.  
Under DFHR’s identity-sensitive constraints, we retain this formulation but \emph{substitute the CLIP visual encoder with ArcFace}, ensuring that the pseudo-token originates from an identity-discriminative embedding.  
Specifically, the ArcFace identity feature $e_{\text{id}}$ is projected into a text-token embedding via a lightweight MLP:
\[
t_{\text{id}} = \phi_{\text{id}}(e_{\text{id}}) \in \mathbb{R}^{d_{\text{text}}}.
\]
This identity token replaces the \texttt{[id]} placeholder in prompts such as  
\texttt{a photo of [id] with \{textual hair description\}},  
after which the complete token sequence is processed by the frozen CLIP text encoder.

This adaptation preserves the core Pic2Word mechanism—learning a pseudo-word through text-space composition—while minimally modifying the architecture to respect DFHR’s identity requirement.  
Nevertheless, similar to ZS-CIR, the resultant representation remains anchored in CLIP’s native space, where identity and hairstyle are entangled, leading retrieval to favor visual resemblance over strict identity fidelity.

\section{Limitations}
\label{sec:limitation}
\noindent\textbf{Inter-Identity Coverage.} 
DFHR-Bench currently fixes the number of identity–hairstyle \emph{target scenarios} to roughly 500, which limits inter-identity coverage. We mitigate this by maximizing \emph{factor coverage}: query faces span varied poses, crops, and incidental hairstyles to stress identity disentanglement, while hair queries include look-alike styles from different identities to stress hairstyle fidelity. As a benchmark for evaluation (not pretraining), this design yields rich combinatorial difficulty despite a bounded number of targets. The resource is structured for incremental \emph{scalability} (adding identities and edge-case hairstyles), and our reliance on publicly available identities preserves \emph{ethical and practical} guarantees (clear licensing, privacy compliance, and reproducibility).

\noindent\textbf{\noindent\textbf{Challenges in Hairstyle Representation}}
Our framework inherits the representational biases of CLIP and related vision and language encoders. While common global attributes are captured well, very specialized or syntactically intricate hairstyles, such as multi braid protective styles, hybrid undercut variations, micro textured perms, or extreme fashion shape designs, are often underrepresented in the CLIP pretraining distribution. As a result, both text based and image based hair queries in these categories may produce degraded similarity estimates or inconsistent emphasis on key attributes. This limitation reflects a broader challenge related to the long tail distribution of hairstyle representation in large foundation models.

\noindent\textbf{Higher Inference Cost for Unified Embedding Space.}
MFHC’s unified text–space embedding provides stronger identity–hairstyle disentanglement but introduces additional computation compared to operating directly in CLIP’s visual space. On a batch of 1000 gallery images, our unified-space indexing requires approximately 41.0 seconds, whereas CLIP-space indexing requires about 19.6 seconds under identical hardware settings. The overhead arises from the extra projection, token-mapping, and fusion steps needed to construct disentangled identity and hairstyle representations. Although this cost is acceptable for offline gallery indexing and yields substantially better retrieval quality.

\begin{table*}[t]
\centering
\caption{Prompt 1 — Generate hairstyle attributes as JSON.}
\label{tab:prompt1-json}
\small
\setlength{\tabcolsep}{8pt}
\renewcommand{\arraystretch}{1.25}
\begin{tabularx}{\textwidth}{@{}p{0.16\textwidth} X@{}}
\hline
\rowcolor{gray!10}
\textbf{System} &
You are a language assistant that generates concise hairstyle analyses as JSON from a single hair image. \\
\hline
\textbf{Prompt} &
\begin{minipage}[t]{\linewidth}\raggedright
Generate the hairstyle description for the following image: \\
\hspace*{1.5em}\textbf{[HAIR IMAGE]} \\
\textless instruction\textgreater \\
You must analyze only what is visible in the image and output \emph{one} JSON object that strictly follows the schema below. Use simple, standard terms. If a field cannot be determined, write ``\texttt{Not Available}''. Do not add any extra text outside the JSON. Keep phrases short and unambiguous. \\
\hspace*{1.5em}Field guidelines: \\
\hspace*{2.5em}1.\ \texttt{overall\_length} — total length (e.g., \textit{Short (above ears)}, \textit{Medium (at jawline)}, \textit{Long (past shoulders)}). \\
\hspace*{2.5em}2.\ \texttt{length\_distribution} — how length varies (top/sides/back; layered vs even). \\
\hspace*{2.5em}3.\ \texttt{styling.direction} — forward / backward / side-swept left or right / downward / upward / mixed. \\
\hspace*{2.5em}4.\ \texttt{styling.distinctive\_styles} — named style (e.g., Bob, Pixie, Crew cut, Undercut, Ponytail, Bun, Afro) or None. \\
\hspace*{2.5em}5.\ \texttt{texture} — straight / wavy / curly / coily. \\
\hspace*{2.5em}6.\ \texttt{volume} — flat / light / medium / high. \\
\hspace*{2.5em}7.\ \texttt{thickness} — fine / medium / coarse. \\
\hspace*{2.5em}8.\ \texttt{bangs} — none / blunt / side-swept / wispy / curtain / micro. \\
\hspace*{2.5em}9.\ \texttt{parting} — center / left / right / zigzag / no visible part. \\
\hspace*{2.5em}10.\ \texttt{hairline} — straight / M-shaped / receding / rounded / widow’s peak / not visible. \\
\hspace*{2.5em}11.\ \texttt{color} — apparent color (e.g., black, dark brown, light brown, blonde, platinum, red, gray, salt-and-pepper, dyed [color], highlights [color]). \\
\hspace*{1.5em}Required output format: \\
\hspace*{2.5em}\{ \\
\hspace*{3.8em}"description": \{ \\
\hspace*{5.1em}"overall\_length": "...", \\
\hspace*{5.1em}"length\_distribution": "...", \\
\hspace*{5.1em}"styling.direction": "...", \\
\hspace*{5.1em}"styling.distinctive\_styles": "...", \\
\hspace*{5.1em}"texture": "...", \\
\hspace*{5.1em}"volume": "...", \\
\hspace*{5.1em}"thickness": "...", \\
\hspace*{5.1em}"bangs": "...", \\
\hspace*{5.1em}"parting": "...", \\
\hspace*{5.1em}"hairline": "...", \\
\hspace*{5.1em}"color": "..." \\
\hspace*{3.8em}\} \\
\hspace*{2.5em}\} \\
\textless/instruction\textgreater
\vspace{0.5em}
\end{minipage}
\\
\hline
\end{tabularx}
\end{table*}
\begin{table*}[t]
\centering
\caption{Prompt 2 — Write a concise natural description from attributes JSON.}
\label{tab:prompt2-text}
\small
\setlength{\tabcolsep}{8pt}
\renewcommand{\arraystretch}{1.25}
\begin{tabularx}{\textwidth}{@{}p{0.16\textwidth} X@{}}
\hline
\rowcolor{gray!10}
\textbf{System} & You are a professional hairstylist. \\
\hline
\textbf{Prompt} &
\begin{minipage}[t]{\linewidth}\raggedright
You are given hairstyle attribute data in JSON format: \{attributes\} \\
Write a concise and natural description of the hairstyle. \\
\hspace*{1.5em}- Cover each available attribute (overall, length, styling.direction, styling.distinctive, texture, volume, thickness, bangs, parting, hairline, color). \\
\hspace*{1.5em}- Keep it clear and straightforward, avoiding excessive adjectives or elaborate phrasing. \\
\hspace*{1.5em}- If an attribute is "Not Available", skip it. \\
\hspace*{1.5em}- The description should be 2--3 sentences, factual but still natural.
\vspace{0.5em}
\end{minipage}
\\
\hline
\end{tabularx}
\end{table*}



%
%

\end{document}